\documentclass[acmtog]{acmart}
\setcopyright{none}          
\renewcommand{\footnotetextcopyrightpermission}[1]{}

\AtBeginDocument{%
  }

\usepackage{colortbl}
\definecolor{highlightyellow}{RGB}{255,248,172} 
\definecolor{highlightorange}{RGB}{255,203,152}
\definecolor{highlightred}{RGB}{255,152,152} 

\newcommand{\added}[1]{#1}  

\newcommand{\name}{NSL}
\newcommand{\namem}{\text{NSL}\text{-Mono}}
\newcommand{\nameb}{\text{NSL}\text{-Bino}}
\newcommand{\names}{\text{NSL}\text{-Stereo}}

\newcommand{\image}{I}

\newcommand{\leftIR}{I_l}
\newcommand{\rightIR}{I_r}
\newcommand{\pattern}{I_p}

\newcommand{\leftIRFeat}{F_l}
\newcommand{\rightIRFeat}{F_r}
\newcommand{\patternFeat}{F_p}
\newcommand{\leftIRContextFeat}{F_{l}^{c}}

\newcommand{\leftFeat}{F_L}
\newcommand{\rightFeat}{F_R}

\newcommand{\depthInit}{D_{init}}
\newcommand{\depthFinal}{D}
\newcommand{\depthGT}{D_{gt}}

\providecommand{\formattedgraphics}[1]{}
\providecommand{\colorbar}[1]{}
\newlength{\imgwidth}
\newlength{\colorbarwidth}
\usepackage{pdfpages}

\begin{document}

\title[Robust Single-shot Structured Light 3D Imaging via Neural Feature Decoding]{Robust Single-shot Structured Light\\ 3D Imaging via Neural Feature Decoding}


\author{Jiaheng Li}
\authornote{Equal contribution}
\email{jiahengli@buaa.edu.cn}
\affiliation{
    \institution{Wangxuan Institute of Computer Technology, Peking University}
    \city{Beijing}
    \country{China}
}

\author{Qiyu Dai}
\authornotemark[1]
\email{qiyudai@pku.edu.cn}
\affiliation{
    \institution{School of Intelligence Science and Technology, Peking University}
    \city{Beijing}
    \country{China}
}

\author{Lihan Li}
\email{lh_li@stu.pku.edu.cn}
\affiliation{
    \institution{Yuanpei College, Peking University}
    \city{Beijing}
    \country{China}
}

\author{Praneeth Chakravarthula}
\email{cpk@cs.unc.edu}
\affiliation{
    \institution{University of North Carolina at Chapel Hill}
    \city{Chapel Hill}
    \state{North Carolina}
    \country{The United States}
}

\author{He Sun}
\email{hesun@pku.edu.cn}
\affiliation{
    \institution{College of Future Technology, Peking University}
    \city{Beijing}
    \country{China}
}

\author{Baoquan Chen}
\authornote{Co-corresponding authors}
\email{baoquan@pku.edu.cn}
\affiliation{
    \institution{School of Intelligence Science and Technology, Peking University}
    \city{Beijing}
    \country{China}
}
\affiliation{
    \institution{State Key Laboratory of General Artificial Intelligence, Peking University}
    \city{Beijing}
    \country{China}
}

\author{Wenzheng Chen}
\authornotemark[2]
\email{wenzhengchen@pku.edu.cn}
\affiliation{
    \institution{Wangxuan Institute of Computer Technology, Peking University}
    \city{Beijing}
    \country{China}
}
\affiliation{
    \institution{State Key Laboratory of General Artificial Intelligence, Peking University}
    \city{Beijing}
    \country{China}
}

\renewcommand{\shortauthors}{Li et al.}

\begin{abstract}

We consider the problem of active 3D imaging using single-shot structured light systems, which are widely employed in commercial 3D sensing devices such as Apple Face ID and Intel RealSense.
Traditional structured light methods typically decode depth correspondences through pixel-domain matching algorithms, resulting in limited robustness under challenging scenarios like occlusions, fine-structured details, and non-Lambertian surfaces.
Inspired by recent advances in neural feature matching, we propose a learning-based structured light decoding framework that performs robust correspondence matching within feature space rather than the fragile pixel domain.
Our method extracts neural features from the projected patterns and captured infrared (IR) images, explicitly incorporating their geometric priors by building cost volumes in feature space, achieving substantial performance improvements over pixel-domain decoding approaches.
To further enhance depth quality, we introduce a depth refinement module that leverages strong priors from large-scale monocular depth estimation models, improving fine detail recovery and global structural coherence.
To facilitate effective learning, we develop a physically-based structured light rendering pipeline, generating nearly one million synthetic pattern-image pairs with diverse objects and materials for indoor settings.
Experiments demonstrate that our method, trained exclusively on synthetic data \added{with multiple structured light patterns}, generalizes well to real-world indoor environments\added{, effectively processes various pattern types without retraining}, and consistently outperforms both commercial structured light systems and passive stereo RGB-based depth estimation methods.
\added{Project page: 
\url{https://namisntimpot.github.io/NSLweb/}
}
\end{abstract}


\begin{teaserfigure}
  \includegraphics[width=\textwidth]{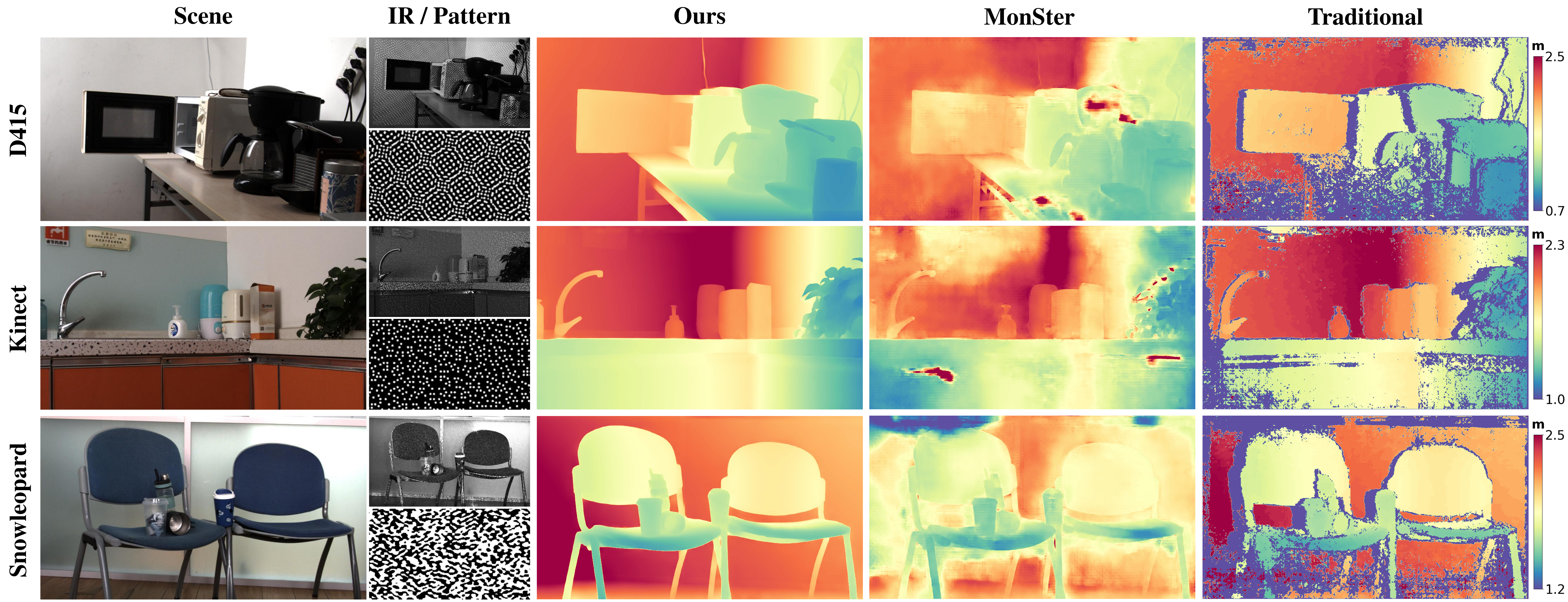}
  \caption{
  Trained entirely on synthetic data, our structured light 3D imaging method generalizes well to real-world indoor scenes. It supports various pattern types (rows) and effectively handles challenging cases, including low-texture regions, fine structural details, and reflective or transparent surfaces. 
  Our method significantly outperforms both RGB-based stereo methods (e.g., MonSter) and traditional pixel-matching-based structured light decoding approaches.
} 
  \Description{Qualitative comparison of real scene-level data.}
  \label{fig:teaser}
\end{teaserfigure}


\maketitle

\section{Introduction}
\label{sec:introduction}

Active structured light (SL) 3D imaging has made significant progress over the past two decades, enabling accurate and efficient 3D reconstruction for numerous applications, including augmented reality, robotics, and industrial automation.
Single-shot structured light systems, which are widely adopted in commercial devices such as Apple Face ID~\cite{FaceID}, Microsoft Kinect~\cite{AzureKinectDK}, and Intel RealSense~\cite{IntelRealSense}, operate by projecting spatially-coded infrared patterns onto the scene and decoding depth information by finding correspondence between the projected pattern and the captured image.
These systems have attracted growing attention due to their simplicity, high efficiency, and suitability for real-time 3D sensing.


Traditional single-shot structured light decoding methods predominantly rely on pixel-domain correspondence matching, i.e., decoding depth from local patch intensity cues.
Commercial systems such as Intel RealSense~\cite{IntelRealSense} and Kinect V1~\cite{tolgyessy2021evaluation} project dot patterns and use patch-based template matching.
However, these approaches are fundamentally limited: image patches contain only low-level, local intensity information, which is often insufficient for robust correspondence, especially under challenging conditions such as occlusions, 
low-texture surfaces, or non-Lambertian reflectance.
As a result, pixel-level decoding becomes highly unstable, compromising 3D acquisition quality in complex scenes.


On the other side, although deep learning has significantly advanced passive RGB-based  stereo tasks, its application to structured light remains limited and ineffective, even though the core of both tasks is stereo matching.
This is mainly due to constraints in both data and method design. On the data side, SL lacks large-scale datasets comparable to those for RGB-based stereo settings. Existing approaches often train on small synthetic datasets or use semi-supervised learning on limited real SL data without ground-truth depth, resulting in poor generalization\cite{zhang2018activestereonet,riegler2019connecting,baek2021polka,xu2022monobino}.
On the methodological side, most prior works ignore the known projected patterns, treating SL as merely RGB stereo with added texture, rather than leveraging the spatial priors encoded in the projections.


In this work, we present \textbf{\name}: a novel \textbf{N}eural \textbf{S}tructured \textbf{L}ight decoding framework that 
simultaneously addresses above challenges. 
At the core of {\name} is to shift stereo matching from the fragile pixel domain to a robust neural feature space.
Inspired by advances in learning-based passive stereo~\cite{chang2018pyramid,guo2019group,xu2020aanet,zhang2019ga,lipson2021raft,jing2023uncertainty,li2024local,chen2024mocha}, we argue that structured light can similarly benefit from learned features.
Our key idea is to match projected patterns and IR images via deep features rather than raw intensities, which moves beyond traditional structured light decoding methods that operate purely in the pixel intensity domain, and substantially improves robustness and accuracy. \added{In contrast to passive stereo methods, which rely solely on images of natural scenes without projected patterns, {\name} extracts matching features mainly from the actively projected, spatially encoded patterns.}


Following setups in commercial SL devices, our method supports monocular (pattern + single IR) and binocular (pattern + stereo IR) inputs.
It consists of two neural stages: a learned feature matching module and a monocular depth refinement module.
The first stage, inspired by RAFT~\cite{lipson2021raft}, uses a stereo network to extract features from both pattern and IR inputs, build hierarchical cost volumes, and iteratively estimate an initial dense depth map.
Importantly, leveraging the pattern features allows the network to exploit spatial priors encoded in the projection in an end-to-end manner, outperforming conventional stereo in textureless regions even using only one camera and one projector (monocular structured light).


To enhance depth quality, we introduce a monocular refinement module as a second stage, aimed at recovering structural detail and correcting mismatches from triangulation.
We adopt a fine-tuned monocular depth estimation (MDE) backbone~\cite{lin2024promptda}, using the initial depth as a prompt.
By combining visual context from the IR image with geometric cues from the coarse depth, the module outputs sharper, more coherent depth maps, particularly around boundaries and fine structures.


To support the training of \name, we develop a physically-based structured light simulation platform in Blender and generate a large-scale, high-fidelity synthetic indoor dataset containing 953K pattern-image pairs.
The dataset covers diverse indoor layouts, materials, textures, lighting conditions, projected patterns and hardware settings, and includes synchronized RGB, IR, and pattern with ground-truth depth labels.
\added{In training, we use a mixture of projected patterns, enabling {\name} to be trained once and then process multiple patterns at inference without retraining.}
Importantly, although our method is trained entirely on synthetic data, it generalizes well to real-world scenes without requiring any fine-tuning. 
This strong generalization capability comes from the fact that the network learns correspondence matching from the projected pattern cues, which exhibit minimal domain gaps between simulation and reality.
%

Extensive experiments show that our method consistently outperforms traditional structured light systems, existing neural structured light approaches, and passive stereo-based learning methods, particularly in challenging regions such as occlusions, fine structural details, reflective surfaces, and low-texture areas.

\begin{figure*}[!t]
\centering
\includegraphics[width=1.0\textwidth]{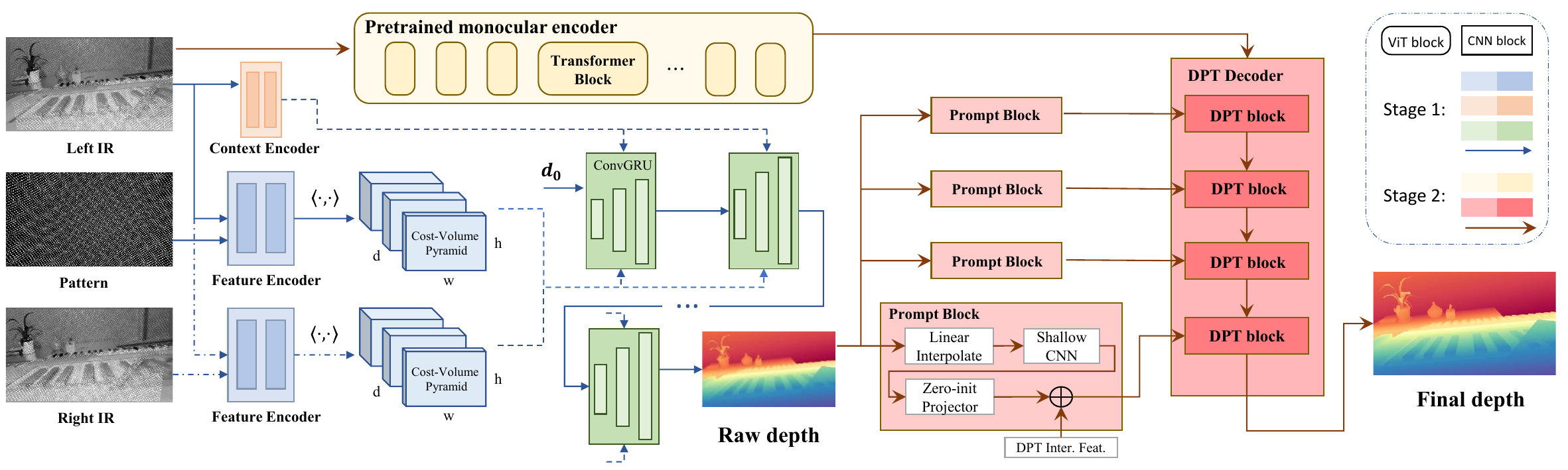}
\caption{\textbf{The pipeline of \name.} Given a single or stereo pair of IR images and a projected pattern, \name\ first estimates an initial raw depth map via the Neural Feature Matching module, which extracts deep features from both IR and pattern inputs, followed by cost volume construction and GRU-based iterative refinement (left path).
Next, the Monocular Depth Refinement module incorporates priors from a monocular depth estimation model, using the initial depth as a prompt (right path), and generates a final depth map with enhanced structural detail.
}
\label{fig:pipeline}
\end{figure*}

\section{Related Work}

In this section, we review three related areas. First, we discuss the evolution of depth imaging techniques, highlighting the contrast between passive and active methods. Next, we introduce traditional structured light systems, focusing on their core decoding principles. Lastly, we examine recent deep learning-based approaches for single-shot structured light decoding.

\subsection{Depth Imaging}
Depth imaging methods can be broadly categorized into \emph{passive} and \emph{active} approaches, depending on the use of external illumination. Passive methods infer depth from naturally available image cues such as shading~\cite{tao2015depth}, parallax~\cite{godard2017unsupervised}, defocus~\cite{hazirbas2019deep}, polarization~\cite{kadambi2015polarized}, and silhouettes~\cite{laurentini1994visual}. These cues are typically exploited in multiview stereo or structure-from-motion pipelines. Although advances in dense camera arrays~\cite{levoy2023light}, neural scene representations~\cite{mildenhall2021nerf} \added{and deep feature–based matching strategies~\cite{zhan2018unsupervised}} have improved reconstruction fidelity, passive approaches still struggle with low-texture surfaces and lighting variations due to their reliance on natural image features and dense viewpoint coverage.

To overcome these challenges, active systems augment the scene with controlled illumination to enhance correspondence. Time-of-flight sensors~\cite{sun2023consistent}, LiDAR~\cite{huang2023neurallidar}, and structured light (SL)~\cite{geng2011structured} are commonly used. SL systems, in particular, offer high-resolution depth sensing at low cost, ideal for near-field real-time applications. Nevertheless, conventional SL techniques often depend on pixel-wise intensity matching, which is sensitive to ambient illumination, specular reflections, and textureless regions.

\subsection{Traditional Active Structured Light Depth Imaging}

Structured Light (SL) has been a cornerstone of active depth sensing for several decades~\cite{will1971grid,posdamer1982surface}, with methods broadly classified into temporal and spatial encoding \cite{salvi2010state}. Temporal SL projects time-varying patterns (e.g., binary \cite{scharstein2003high}, gray \cite{aliaga2008photogeometric,posdamer1982surface}, or fringe patterns \cite{kawasaki2008dynamic,koninckx2006real,sagawa2011dense,taguchi2012motion}) and decodes per-pixel signals for fast depth recovery, but suffers from motion artifacts due to multi-frame requirements.

Spatial SL, on the other hand, uses a single static 1D or 2D pattern \cite{maruyama1993range,zhang2002rapid,le1988structured,vuylsteke1990range}, enabling one-shot capture and better motion robustness. This shifts the challenge to correspondence estimation via image matching. For example, Kinect V1 \cite{martinez2013kinect} employs local block matching against a reference, but such correlation-based approaches can struggle in non-ideal regions due to assumptions like local photo-consistency.

\subsection{Deep Learning for Single-Shot Structured Light Decoding}
Despite the success of deep learning in passive stereo, its application to single-shot structured light (SL) decoding remains underexplored. Early methods lacked end-to-end learning frameworks and achieved limited performance~\cite{fanello2016hyperdepth,fanello2017ultrastereo}. ActiveStereoNet~\cite{zhang2018activestereonet} was among the first to apply CNN-based training for SL stereo matching, but it excluded the projected pattern from inference and relied on semi-supervised learning due to limited data that lacks ground-truth depth, resulting in unstable training and poor generalization.
Connecting~\cite{riegler2019connecting}, though categorized under neural SL, is not a decoding method. It formulates the problem as monocular depth estimation without using the projected pattern during inference, leading to poor cross-device generalization. 

Polka~\cite{baek2021polka} introduced a parametric diffractive optical element (DOE) model for jointly optimizing pattern design and depth decoding, while MonoStereoFusion~\cite{xu2022monobino} proposed a two-stage pipeline where the coarse depth is generated through traditional decoding between the projected pattern and the left IR image, and subsequently used to guide stereo matching between binocular IR images. Both methods highlight the benefit of incorporating pattern information, but Polka is restricted to DOE-generated dot patterns, and MonoStereoFusion excludes the pattern from end-to-end learning, underutilizing SL priors.
Furthermore, both approaches synthesize training data by overlaying patterns on passive stereo datasets—an approach that lacks physical realism and fails to capture challenges such as reflectance and ambient light, thus limiting generalization.

In contrast, our method extracts neural features from both the projected pattern and IR image in an end-to-end framework, using a physically based simulator to generate large-scale synthetic data with diverse materials and pattern types. This results in improved accuracy, robustness, and sim-to-real generalization across occlusions, reflective surfaces, and low-texture regions.

\section{Methods} 

We now describe the pipeline of \name, a single-shot structured light decoding framework that replaces traditional pixel-domain template matching with robust neural feature decoding. 
Given either a single or stereo pair of IR images captured by a structured light system and the corresponding projected pattern, \name\ produces a dense, high-quality depth map of the scene. 
As illustrated in Figure~\ref{fig:pipeline}, \name\ consists of two main components: (1) a Neural Feature Matching module that extracts deep features from the IR image(s) and pattern to estimate an initial depth map, and (2) a Monocular Depth Refinement module that leverages priors from a monocular depth estimation model to improve depth quality.
Below, we describe each module in detail.

\subsection{Neural Feature Matching} 
\label{sec:method_matching}

To replace fragile pixel-domain decoding in structured light (SL), we propose a neural feature matching module that estimates dense correspondences in learned feature space. Inspired by RAFT-Stereo~\cite{lipson2021raft}, our model extracts deep features from the projected pattern and IR image(s), builds multi-level cost volumes, and iteratively refines depth predictions using a GRU-based updater.

\paragraph{Feature Extraction.} 
Let $\leftIR \in \mathbb{R}^{H \times W}$ denote the left IR image (used in both monocular and binocular SL modes), $\rightIR \in \mathbb{R}^{H \times W}$ denote the right IR image (used only in binocular SL mode), and $\pattern \in \mathbb{R}^{H \times W}$ denote the projected pattern.
A CNN-based \textit{Feature Encoder} $\text{Enc}^{lp}$ extracts 1/4-resolution feature maps from the Left IR image and the projected pattern. For stereo IR input, another encoder $\text{Enc}^{lr}$ with the same architecture is also used to process the Left and Right IR images.
\begin{equation}
\begin{aligned}
\leftIRFeat^{lp} = \text{Enc}^{lp}(\leftIR), \quad \patternFeat = \text{Enc}^{lp}(\pattern), \\
\leftIRFeat^{lr} = \text{Enc}^{lr}(\leftIR), \quad \rightIRFeat = \text{Enc}^{lr}(\rightIR).
\end{aligned}
\end{equation}

To utilize global image priors in depth reasoning, we further use a separate \textit{Context Encoder} to extract multi-scale context features from $\leftIR$:
\begin{equation}
\leftIRContextFeat = \text{ContextEnc}(\leftIR),
\end{equation}
where $\leftIRContextFeat$ provides multi-resolution cues used in the GRU-based refinement stage.

\paragraph{Cost Volume Construction.} 
In traditional RGB-based stereo matching, a 3D cost volume $C \in \mathbb{R}^{H \times W \times W}$ is constructed by computing the dot product between feature maps $\leftFeat$ and $\rightFeat$:
\begin{equation}
C(i, j, k) = \sum_{h} {\leftFeat}_{i,j,h} \cdot {\rightFeat}_{i,k,h},
\end{equation}
where $(i, j)$ denotes the pixel location in the left image and $k$ enumerates correspondence candidates in the right image. This gives a matching score between pixel $(i,j)$ in the left image and pixel $(i,k)$ in the right image.

In our \name\ framework, $\leftFeat$ and $\rightFeat$ correspond to the left IR image feature $\leftIRFeat^{lp}$ and the pattern feature $\patternFeat$.
This formulation enables pattern-to-image matching, allowing the network to leverage the spatial information encoded in the pattern. 
For stereo IR input, a second cost volume is additionally constructed from $\leftIRFeat^{lr}$ and $\rightIRFeat$.

To capture multi-scale cues, we build a 4-level cost volume pyramid by applying average pooling along the last dimension of $C$:
\begin{equation}
C^{(l)} = \text{AvgPool}^{(l)}(C), \quad \text{for } l = 0, 1, 2, 3.
\end{equation}

In the binocular SL setup, 
features from the two cost volume pyramids, one for left-pattern matching and one for stereo ir matching, are concatenated together and used jointly in the refinement stage.

\paragraph{GRU-based Depth Prediction.}
Following RAFT-Stereo, we apply a GRU-based iterative update to predict disparity. Starting from an initial disparity estimate $d_0$, the module generates a sequence of disparity maps $\{d_1, \dots, d_N\}$ over $N$ iterations. The final predicted disparity map $d_N$ is then converted into the initial raw depth map $\depthInit$.

At each iteration $t$, we sample correlation features from the multi-scale cost volume at the current disparity $d_t$, denoted as $\phi(C, d_t)$. These features, along with the context features $\leftIRContextFeat$, are fed into a convolutional GRU:
\begin{equation}
h_{t+1} = \text{GRU}(h_t, \phi(C, d_t), \leftIRContextFeat),
\end{equation}
\begin{equation}
\Delta d_t = \text{Regress}(h_{t+1}), \quad d_{t+1} = d_t + \Delta d_t,
\end{equation}

where $h$ is the hidden feature in each iteration. After $N$ iterations, the final disparity prediction $d_N$ is converted to an initial depth $\depthInit$, which is used as the initial depth estimate for the refinement module.

\setlength{\tabcolsep}{1.2pt}

\setlength{\imgwidth}{0.15\textwidth}
\setlength{\colorbarwidth}{0.0182\textwidth} 

\renewcommand{\formattedgraphics}[1]{%
      \includegraphics[width=\imgwidth, keepaspectratio]{#1}%
}

\renewcommand{\colorbar}[1]{%
      \includegraphics[width=\colorbarwidth, keepaspectratio]{#1}%
}

\begin{figure}[htb!]
  \centering
  \resizebox{0.5\textwidth}{!}{%
      \begin{tabular}{cccc}
        \textbf{Scene} & \textbf{Initial Depth} & \textbf{Final Depth} \\           
            \includegraphics[width=0.15\textwidth,keepaspectratio]{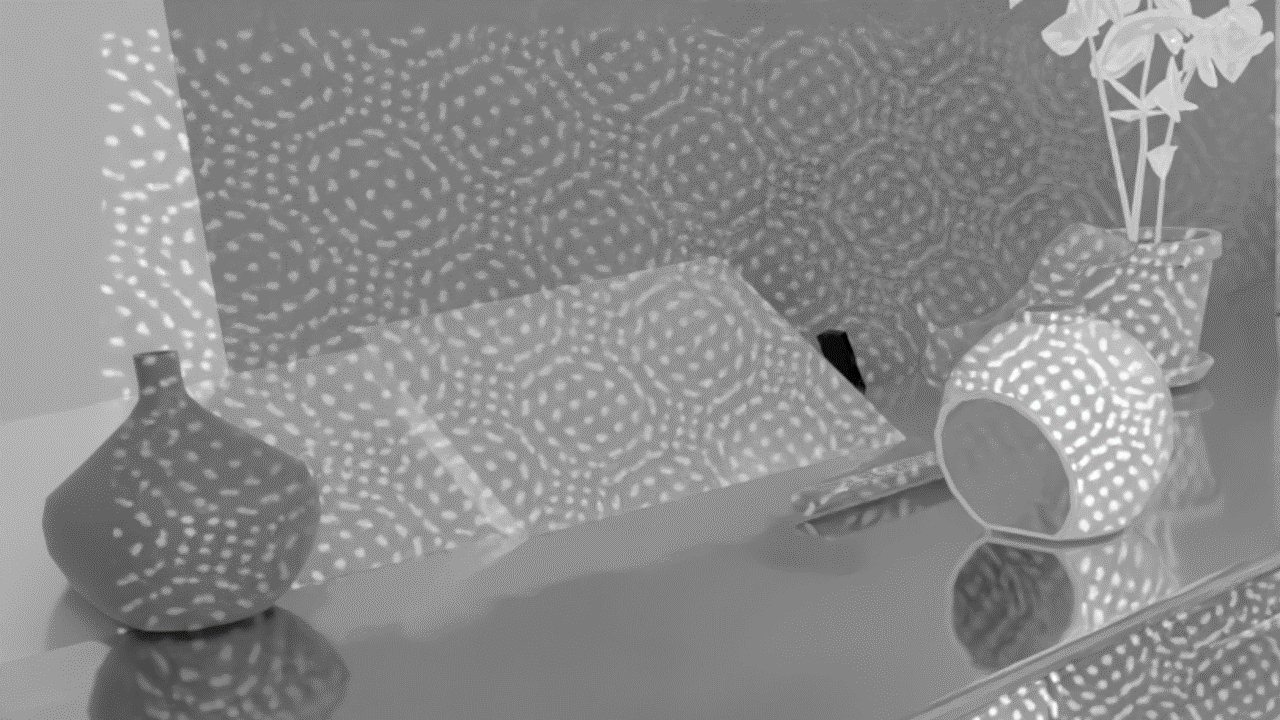} &
            \includegraphics[width=0.15\textwidth,keepaspectratio]{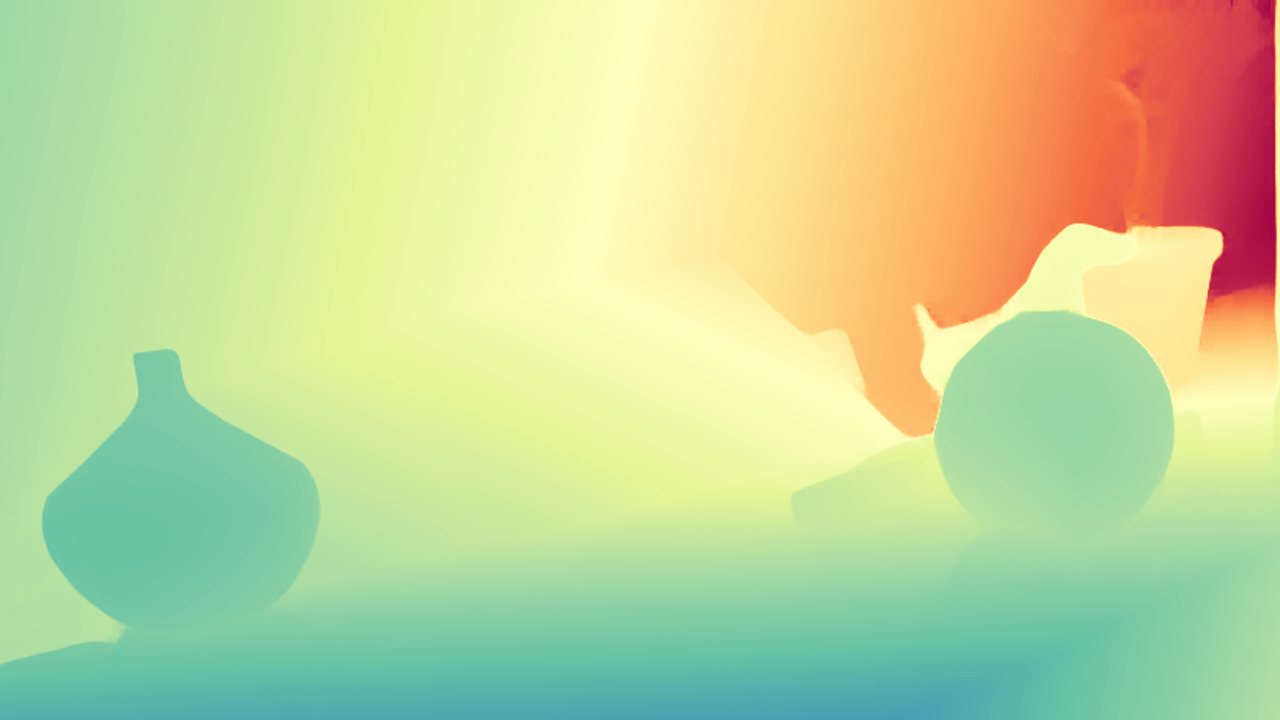} &
            \includegraphics[width=0.15\textwidth,keepaspectratio]{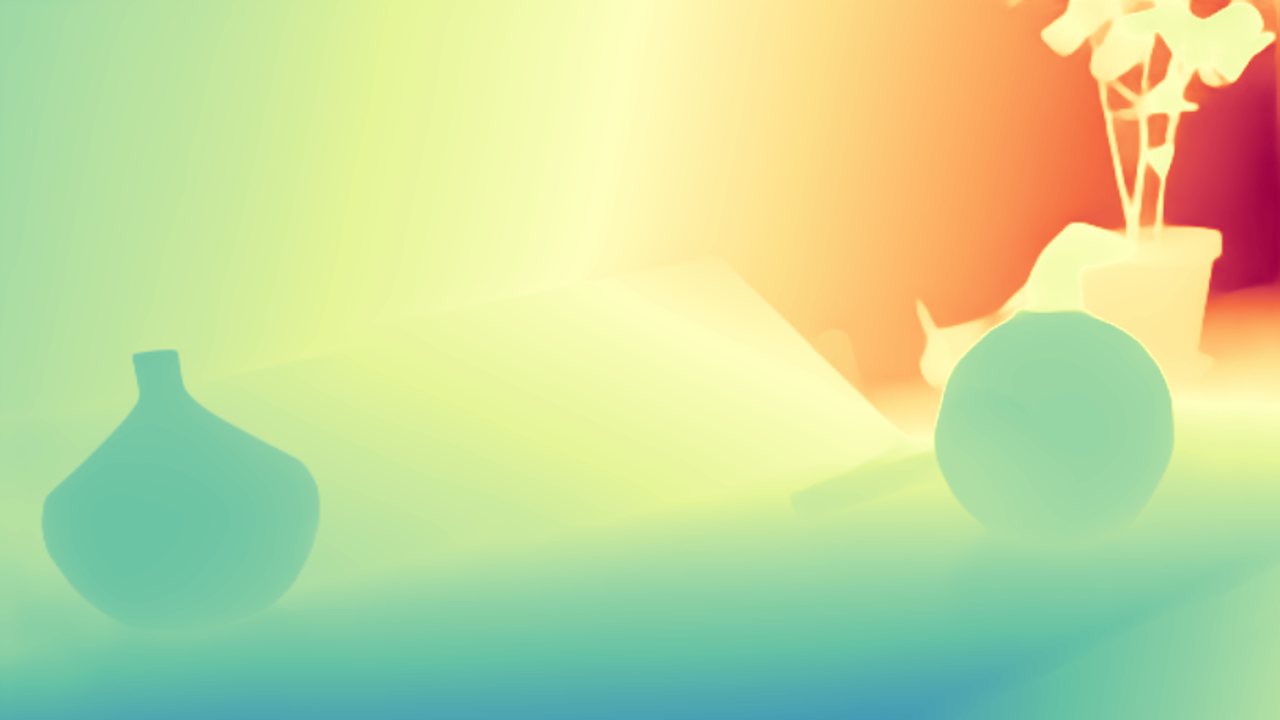} &
            \includegraphics[width=0.0182\textwidth,keepaspectratio]{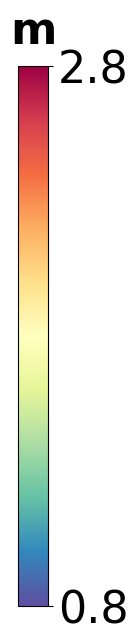} \\[0.5pt]
            \includegraphics[width=0.15\textwidth,keepaspectratio]{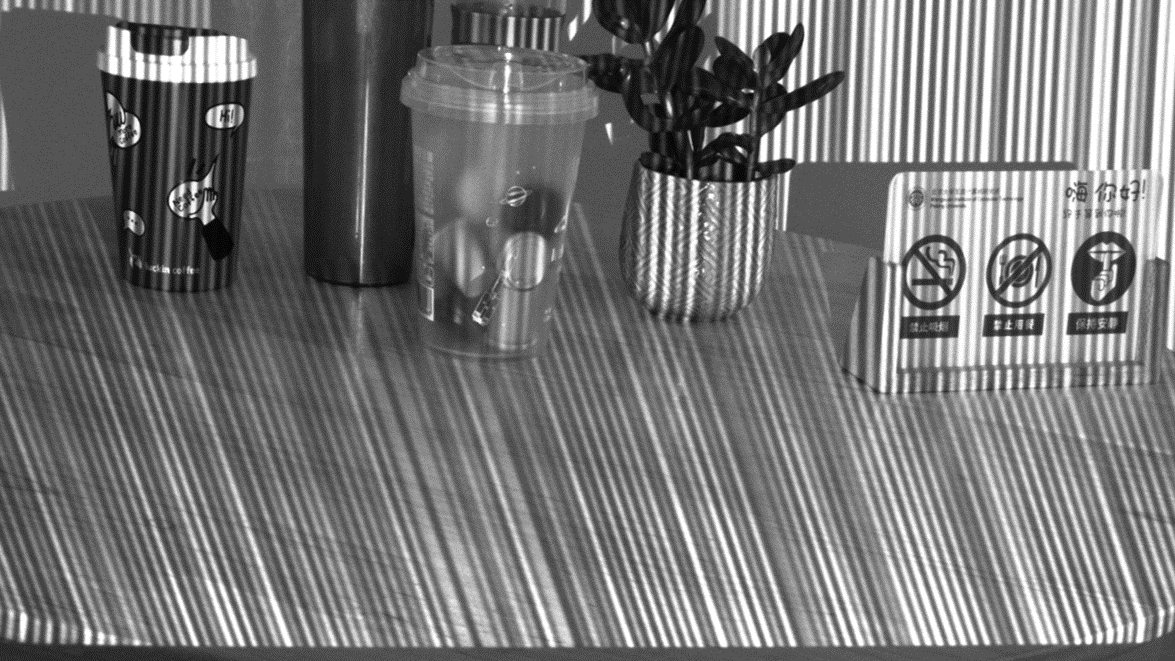} &
            \includegraphics[width=0.15\textwidth,keepaspectratio]{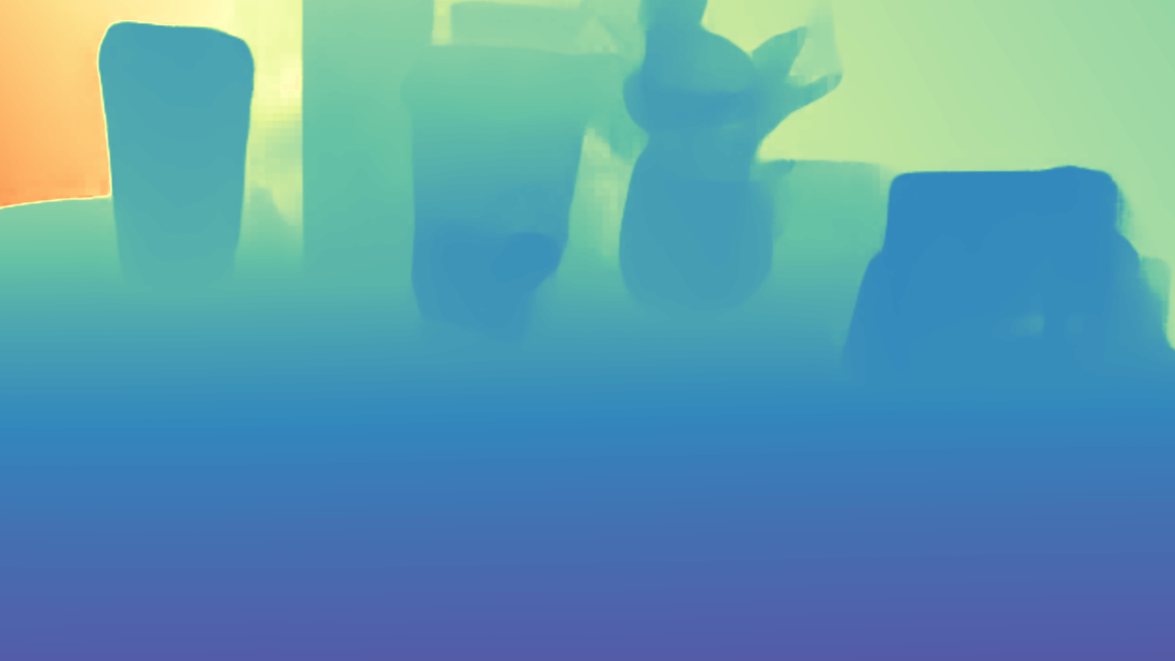} &
            \includegraphics[width=0.15\textwidth,keepaspectratio]{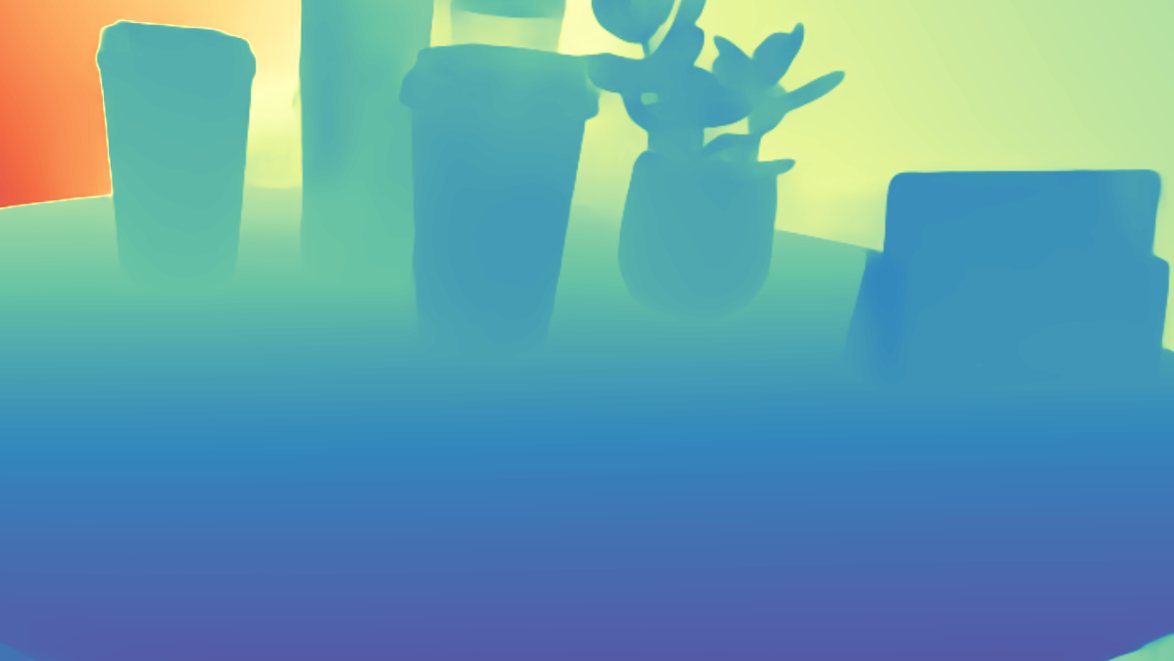} &
            \includegraphics[width=0.0182\textwidth,keepaspectratio]{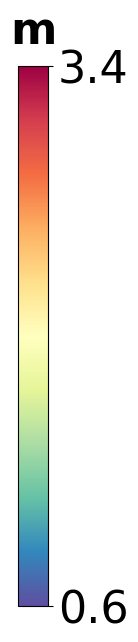} \\
      \end{tabular}
    }
  \caption{
  \textbf{Effectiveness of Monocular Depth Refinement}. The first row shows results on synthetic data, and the second row shows results on a real-captured scene. The refinement module uses monocular depth priors to recover challenging regions, such as thin structures and sharp edges. 
  }
  \label{fig:abla_mdr_module}
\end{figure}

\subsection{Monocular Depth Refinement} \label{sec:method_refine}

While neural feature matching significantly improves structured light decoding, the resulting depth can remain suboptimal, particularly in challenging regions such as thin structures, occlusions, or indirect illumination.
To address this, we introduce a Monocular Depth Refinement module that leverages global image priors from monocular depth estimation (MDE) foundation models~\cite{lin2024promptda}.
These models, trained on millions of images, can infer plausible depth from a single RGB input, but typically suffer from scale ambiguity. 
We resolve this issue by using the structured light initial depth $\depthInit$ as a geometric prior to guide the model and produce metrically accurate depth maps. Fig.~\ref{fig:abla_mdr_module} presents a visual comparison with and without this module.

\paragraph{Depth Foundation Model.}
We adopt Depth Anything v2~\cite{yang2024dav2} as our base model. It first extracts multi-scale image features using a Vision Transformer (ViT)~\cite{oquabdinov2}, then aggregates them using a DPT-style decoder~\cite{ranftl2021vision}. The final depth is predicted via a convolutional head. 

Previous works~\cite{lin2024promptda} demonstrate that foundation models like Depth Anything v2 supports the injection of auxiliary geometric signals, such as LiDAR, by incorporating them into the DPT decoder. We then modify its input and prompt settings to accommodate to our structured light setting.

\paragraph{Structured Light Prompt Injection.}
In our case, we replace the RGB image $\image$ with the IR input $\leftIR$. Additionally, we use the initial depth $\depthInit$ as a depth prompt. The prompt network adopts the architecture proposed in PromptDA (see Fig.~\ref{fig:pipeline}). For more efficient training and inference, our approach employs a backbone model at the ViT-Base scale.

\subsection{Structured Light Simulation and Data Generation} \label{sec:method_data} 

A key bottleneck preventing the application of neural methods to SL decoding is the lack of large-scale labeled data. 
Robust neural decoders typically require training on millions of diverse image pairs to generalize well across real-world variations. 
However, collecting such real-world SL data at scale is time-consuming and costly. 

To address this, we train our model entirely on synthetic data. We develop a physically-based SL simulation platform using Blender’s Cycles ray-tracing engine for rendering a large amount of synthetic SL data for training.

\paragraph{Dataset Composition.} We construct a large-scale indoor dataset tailored for structured light decoding. The dataset includes: 
\begin{itemize} 
\item \textbf{Scenes:} 2860 indoor scenes generated from Procthor\cite{procthor}, illuminated with realistic HDR environment maps (100 maps). 
\item \textbf{Objects:} 5000+ 3D objects from sources like ShapeNet or Objaverse, covering a wide range of categories with diffuse, specular, and transparent materials. 
\item \textbf{Patterns:} eight structured light patterns (e.g., random binary textures, dot patterns), of which six are used for training and two for testing. 
\end{itemize} 
During rendering, we randomize object placement, materials, scene layout, camera poses, projected patterns, camera parameters, baseline distance, and lighting conditions. This diversity helps simulate the wide range of challenging scenarios encountered in real-world structured light capture, such as occlusions, low-texture surfaces, non-Lambertian materials, and complex illumination. We generate a total of 953K samples. Each sample includes binocular RGB images, binocular IR images, a pattern image, and a ground-truth depth map. We show the dataset in the Supp.

\paragraph{Sim2Real Generalization.} Although our model is trained entirely on synthetic data, it generalizes effectively to real-world structured light systems. We attribute this to the physically accurate simulation process and the fact that our decoder learns correspondences directly from pattern-guided feature matching, which minimizes the domain gap between synthetic and real data. And the monocular depth refinement module also maintains this generalization, leveraging its robust foundation model backbone and focusing on localized corrections guided by the initial depth, as shown in 
Fig.~\ref{fig:teaser},~\ref{fig:abla_mdr_module} and ~\ref{fig:real_more_scene}.

\subsection{Network Training} \label{sec:method_training}

We train the Neural Feature Matching and Monocular Depth Refinement modules in two stages: 200k iterations for the former, followed by 100k iterations fine-tuning the latter using $\depthInit$ from stage one.

\paragraph{Neural Feature Matching.}
The feature matching module predicts a sequence of disparity maps $\{d_1, \dots, d_N\}$ through iterative GRU refinement. We supervise each stage with a weighted L1 loss against the ground truth disparity $d_{gt}$: \begin{equation} \mathcal{L}_{\text{stage 1}} = \sum_{t=1}^{N} \lambda_t \cdot |d_t - d_{gt}|_1, \end{equation} where $\lambda_t$ denotes the weight for iteration $t$, with larger weights assigned to later stages for better supervision on refined outputs.

\paragraph{Monocular Depth Refinement.}
The refinement module is initialized from the pre-trained Depth Anything v2~\cite{yang2024dav2} and fine-tuned on our dataset. It takes the IR image $\leftIR$ and the initial depth $\depthInit$ as input and is supervised by two terms: \begin{equation} \mathcal{L}_{\text{stage 2}} = |\depthFinal - \depthGT|_1 + \alpha \cdot |\nabla \depthFinal - \nabla \depthGT|_1, \end{equation} where the first term ensures metric accuracy, and the second enforces edge consistency. $\alpha$ is a fixed weight (set to 0.5), and $\nabla$ denotes the spatial gradient.
\section{Experimental Results}
We conduct comprehensive experiments on both synthetic and real-world datasets to evaluate the effectiveness and generalizability of the proposed \name\ framework. Section~\ref{sec:exp_setup} details the evaluation metrics and baseline methods. Section~\ref{sec:synthetic_results} presents results on the synthetic dataset, followed by ablation studies in Section~\ref{sec:ablation}, which analyze the impact of various design choices. Finally, Section~\ref{sec:real_results} demonstrates the framework's practical applicability using data captured from both a custom structured light setup and the commercially available RealSense D435 RGB-D camera.

\subsection{Comparison Details} \label{sec:exp_setup}

\paragraph{Configuration.}
As our method is very flexible, it supports various input configuration including monocular (single IR + Pattern) and binocular (dual IR + Pattern) SL settings. 
We train two models under both settings, noted as \textbf{{$\namem$} (monocular, single IR + Pattern)} and \textbf{{$\nameb$} (binocular, dual IR + Pattern)}.

\paragraph{Baselines.}
\added{We compare against both \textbf{intensity-based SL} and more recent \textbf{learning-based SL}. For the intensity-based baseline, we adopt a \textbf{traditional template-matching SL method (TM)}\\ \cite{libSGM}, which is a pixel intensity based approach widely used in commercial products. For the learning-based baseline, we compare with \textbf{ActiveStereoNet} \cite{zhang2018activestereonet}, a CNN-based SL approach.} 
\added{We note that other learning-based SL works, MonoStereoFusion~\cite{xu2022monobino} (using a non-standard RGB+IR+pattern setup) and Polka~\cite{baek2021polka} (requiring pattern-decoder co-optimization), are not included since they provide no public-available code and are incompatible with our evaluation settings.}.

ActiveStereoNet is retrained and tested on our dataset using only image pairs rendered with binary patterns—where pixel values are strictly 0 or 1—since it crashes when pairs with non-binary patterns are used. Even under this restricted setup, training remains highly unstable, and the results reported are the best among multiple runs.

\added{Additionally, to contextualize {\name} in the broader depth estimation landscape and provide broader comparison, we include a state-of-the-art passive RGB stereo network MonSter~\cite{cheng2025monster}. We fine-tune it using the passive RGB pairs in our dataset.}

\paragraph{Evaluation Metrics.}
We use standard depth estimation metrics: RMSE(Root Mean Squared Error), MAE(Mean Absolute Error), and REL(Mean Relative Error) to measure overall, average, and relative errors. We also report $\delta$-accuracy (thresholds $\delta \in \{1.05, 1.10, 1.25\}$), which measures the percentage of pixels where $\max(\frac{D_i}{D_i^*}, \frac{D_i^*}{D_i}) < \delta$. \added{Though disparity based metrics like EPE (Endpoint error) are also commonly used in stereo systems, we choose depth based metrics because depth is the ultimate target and is insensitive to irrelevant conditions like image resolution.}

\subsection{Synthetic data evaluation}
\label{sec:synthetic_results}

\paragraph{Experimental Setup.}
We validate our method on the proposed synthetic dataset. \added{We use a resolution of 1280$\times$720 for evaluation.}  
\added{During training, we only use a mixture of 6 patterns, while we evaluate the performance under all 8 patterns (including 2 unseen patterns).} See Supp. for full details. Below, we compare baselines under both monocular setting and binocular setting.

\begin{table}[htbp]
\caption{Quantitative Comparison of Our Method with Traditional Template Matching Methods (TM) and RAFT.}
\centering
\resizebox{0.48\textwidth}{!}{%
    \renewcommand{\arraystretch}{1.2}
    \setlength{\tabcolsep}{2.5pt}
    \begin{tabular}{p{3cm}|ccc|ccc}
    \toprule[0.9pt]
    \textbf{Methods} & \textbf{MAE}(m) ↓ & \textbf{RMSE} ↓ & \textbf{REL} ↓ & $\boldsymbol{\delta_{1.25}}$ ↑ & $\boldsymbol{\delta_{1.10}}$ ↑ & $\boldsymbol{\delta_{1.05}}$ ↑ \\
    \hline
    \multicolumn{7}{c}{\textbf{Monocular Setting}} \\
    \hline
    TM(IR+Pattern)$^{\dagger}$ & 0.5250 & 1.1984 & 0.2886 & 0.7467 & 0.6955 &  0.5716\\
    \textbf{Ours($\namem$)} & \cellcolor{highlightorange} 0.0408 & \cellcolor{highlightorange} 0.0907 & \cellcolor{highlightorange} 0.0357 & \cellcolor{highlightorange} 0.9704 & \cellcolor{highlightorange} 0.9437 & \cellcolor{highlightorange} 0.9142  \\
    \hline
    \multicolumn{7}{c}{\textbf{Binocular Setting}} \\
    \hline
    TM(Dual IR)$^{\dagger}$ & 0.3568 & 1.9008 & 0.2988 & 0.8001 & 0.7790 & 0.7532 \\
    ActiveStereoNet$^{\ddagger}$ & 0.5693 & 75.5485 & 0.4984 & 0.8455 & 0.7912 & 0.7287 \\
    MonSter(passive) & 0.0862 & 0.1457 & 0.1839 & 0.9059 & 0.8611 & 0.8087  \\
    \textbf{Ours($\nameb$)} & \cellcolor{highlightred} 0.0237 & \cellcolor{highlightred} 0.0634 & \cellcolor{highlightred} 0.0210 & \cellcolor{highlightred} 0.9817 & \cellcolor{highlightred} 0.9666 & \cellcolor{highlightred} 0.9502 \\
    \bottomrule[0.9pt]
    \end{tabular}
}

\vspace{3pt} 
\begin{minipage}{0.48\textwidth}
\footnotesize
$^{\dagger}$ For the pixel-matching based traditional method (TM)~\cite{libSGM}, large depth outliers are excluded before metric calculation; other methods do not undergo outlier removal. \\
$^{\ddagger}$ ActiveStereoNet is evaluated only on the subset rendered with binary patterns. Others are evaluated on all patterns.
\end{minipage}

\label{tab:baselinecomp}
\end{table}

\paragraph{Performance Discussion.}
Quantitative results are summarized in Table~\ref{tab:baselinecomp}. Our method significantly outperforms the classical single-shot pixel-based matching decoding approach, the representative SL neural decoding method ActiveStereoNet and the state-of-the-art passive RGB-based stereo method, across all metrics.

Specifically,  even monocular SL configuration ({\namem}, IR + pattern) already surpasses MonSter and ActiveStereoNet, which takes dual RGB or dual IR as input. Extending to the binocular SL setting (dual IR + pattern) further enhances our method's performance, demonstrating the advantage of utilizing SL pattern priors when combined with feature-space correspondence learning.

\begin{figure*}[htb]
  \centering
  \resizebox{0.95\textwidth}{!}{%
      \begin{tabular}{cccccccc}
        \textbf{Scene} & \textbf{IR} & \textbf{GT Depth} & 
        \textbf{Ours} & \textbf{MonSter} &  \textbf{Traditional} & \textbf{ActiveStereoNet}
        \\[0.5pt]
        \includegraphics[width=0.138\textwidth,keepaspectratio]{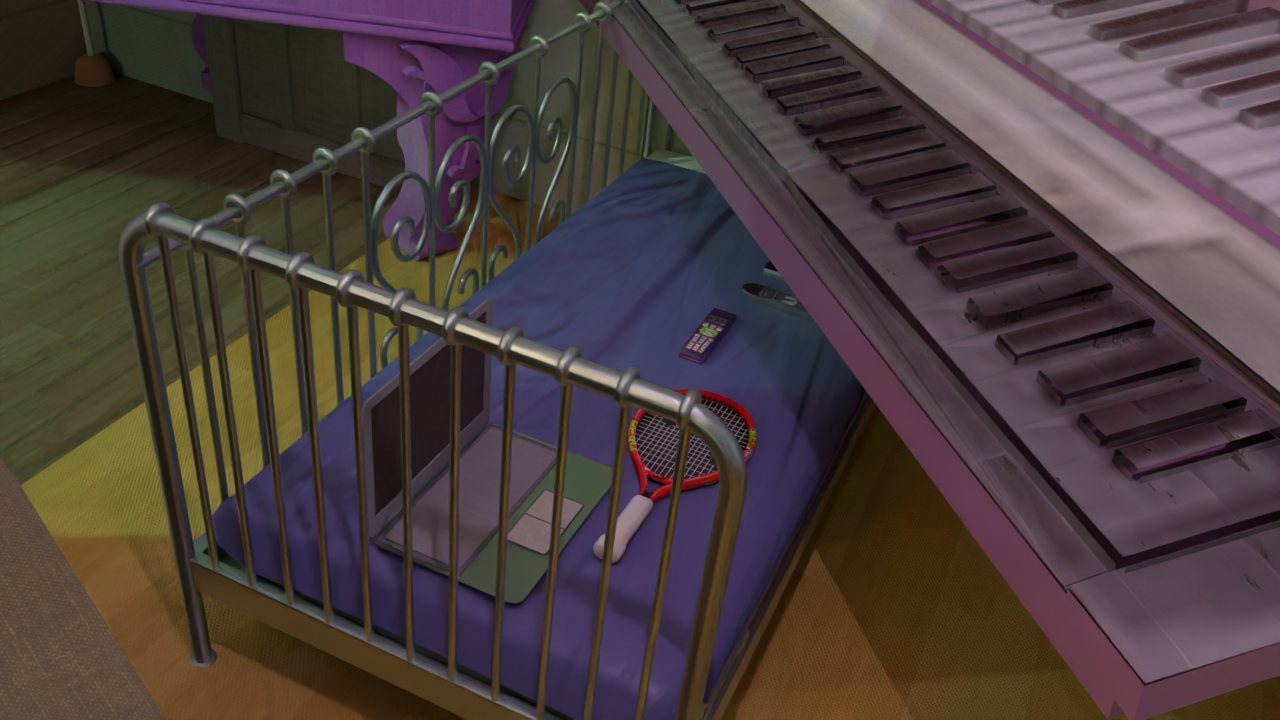} &
        \includegraphics[width=0.138\textwidth,keepaspectratio]{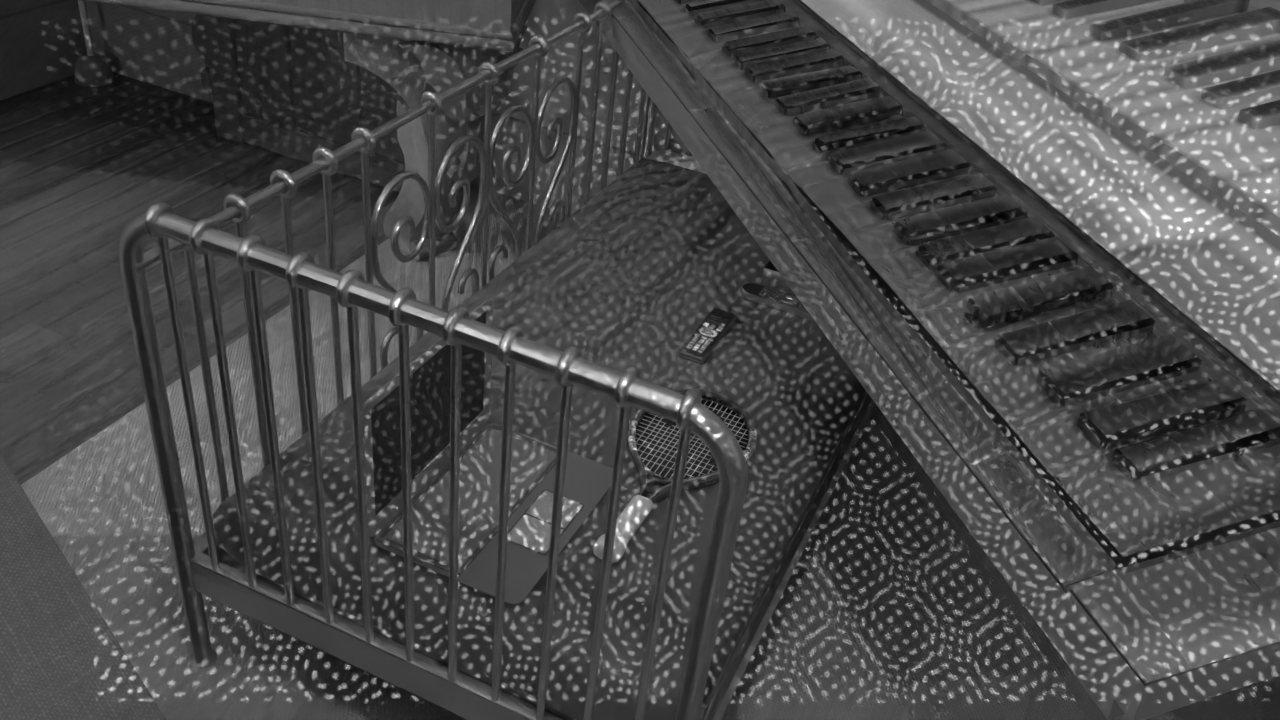} &
        \includegraphics[width=0.138\textwidth,keepaspectratio]{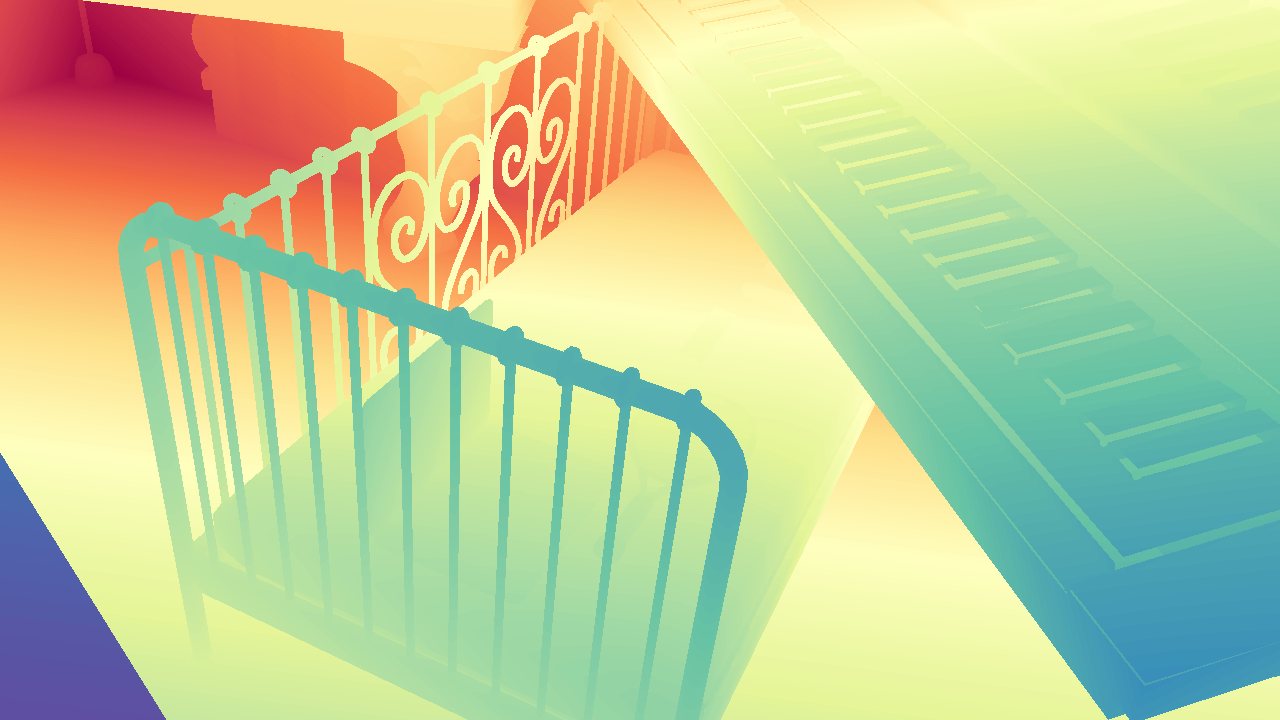} &
        \includegraphics[width=0.138\textwidth,keepaspectratio]{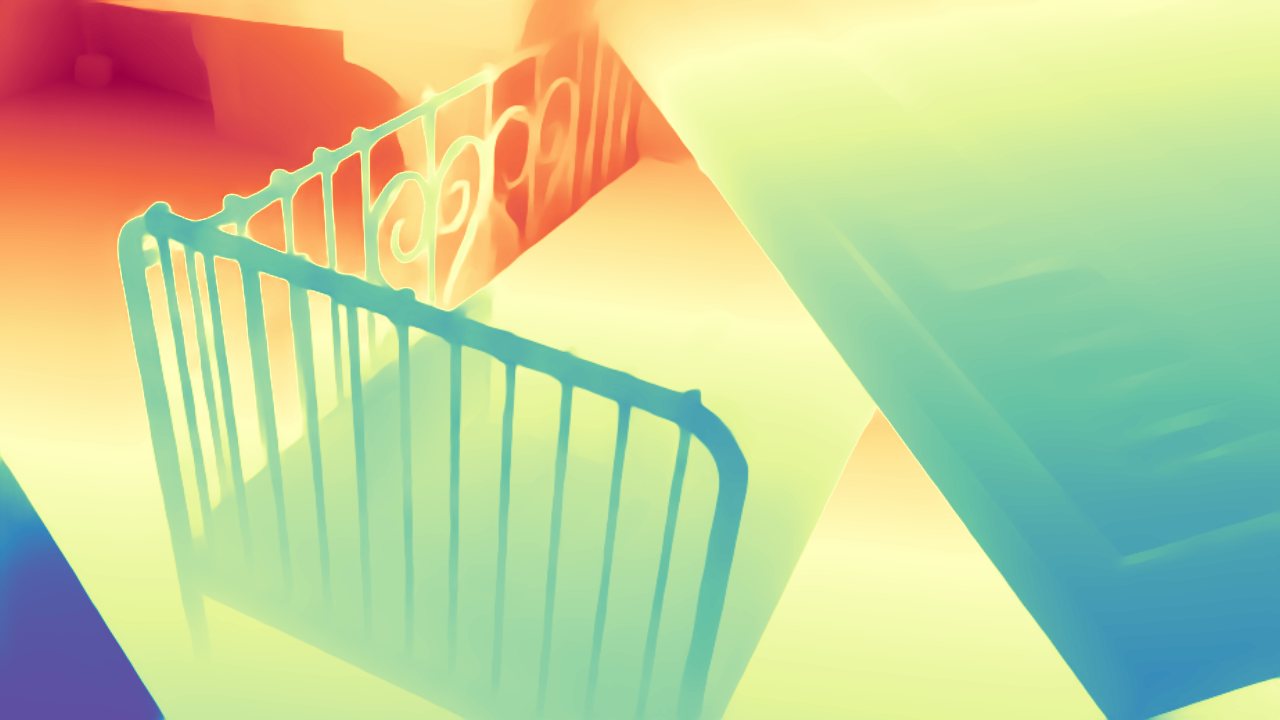} &
        \includegraphics[width=0.138\textwidth,keepaspectratio]{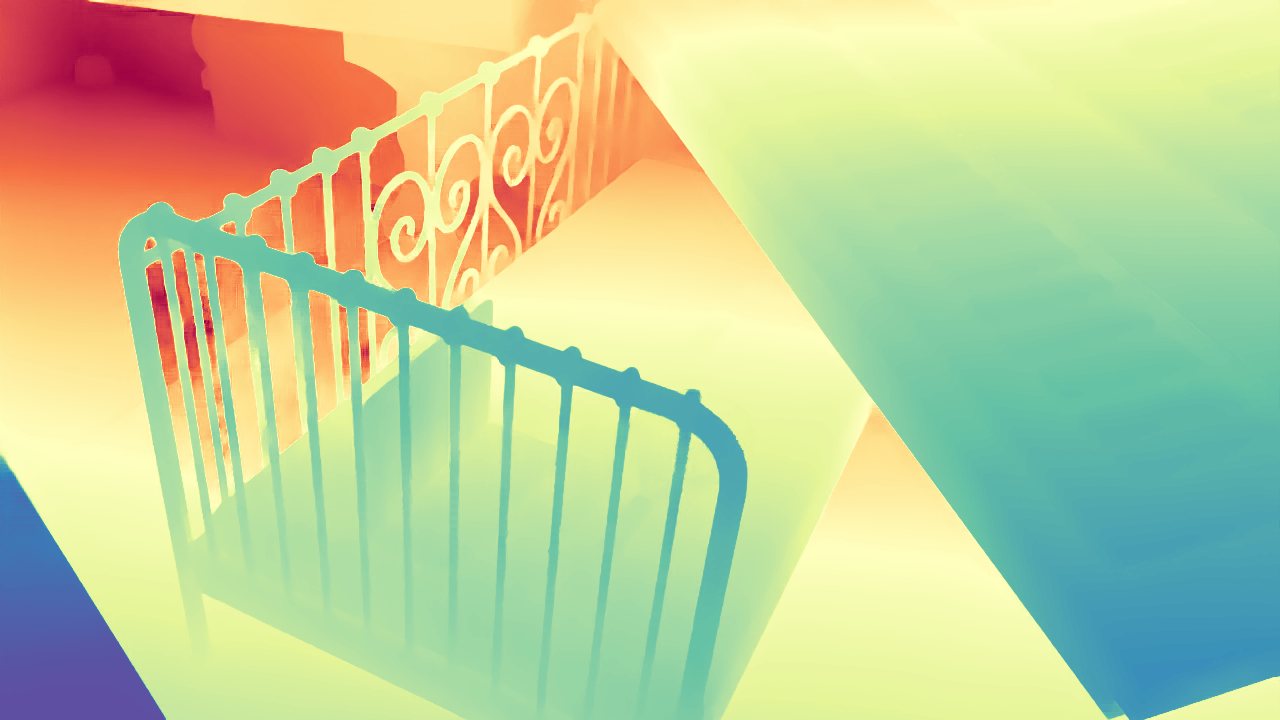} &
        \includegraphics[width=0.138\textwidth,keepaspectratio]{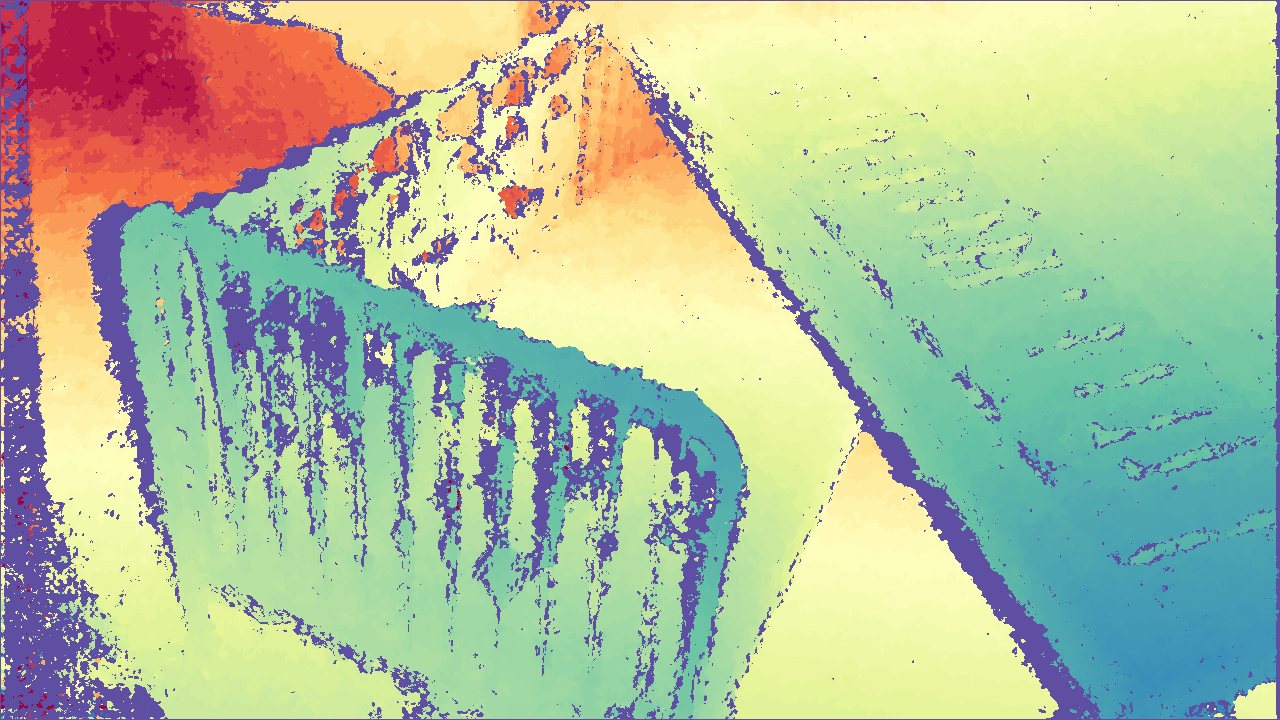} &
        \includegraphics[width=0.138\textwidth,keepaspectratio]{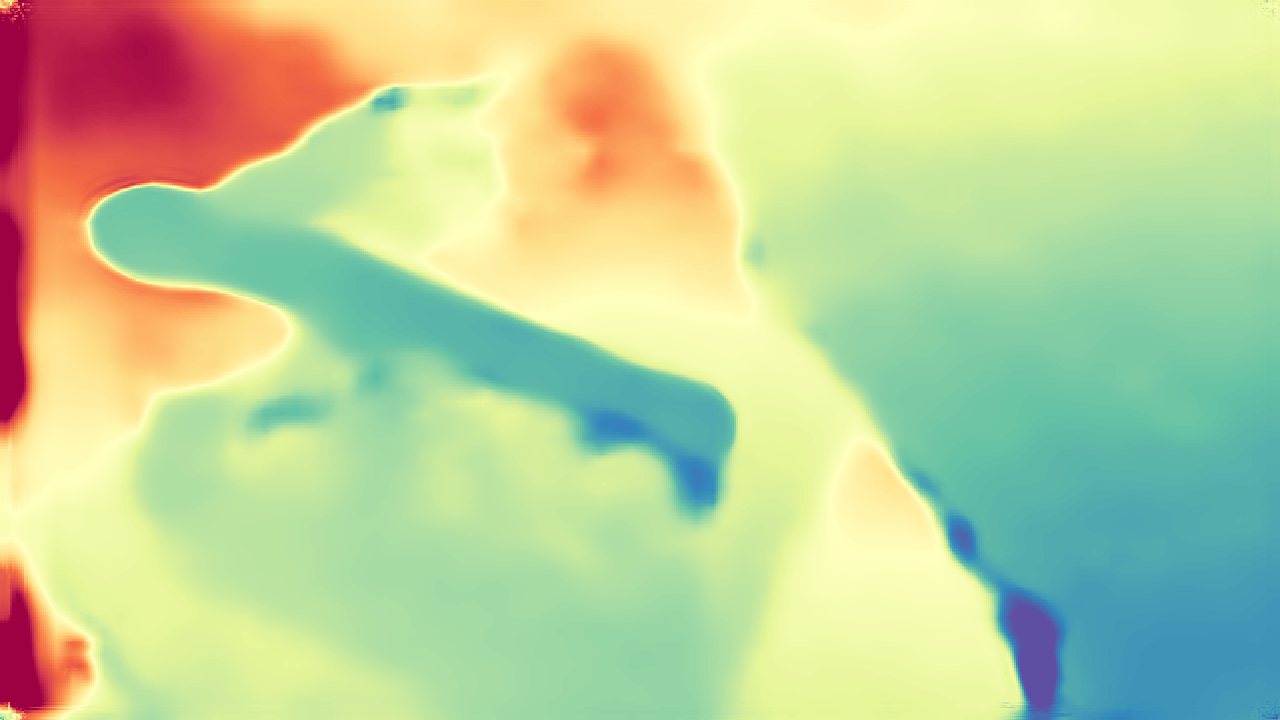} & 
        \includegraphics[width=0.0182\textwidth,keepaspectratio]{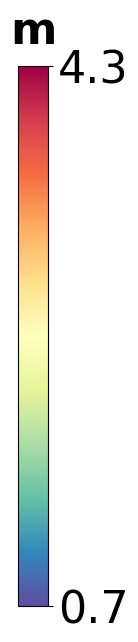}\\

        \includegraphics[width=0.138\textwidth,keepaspectratio]{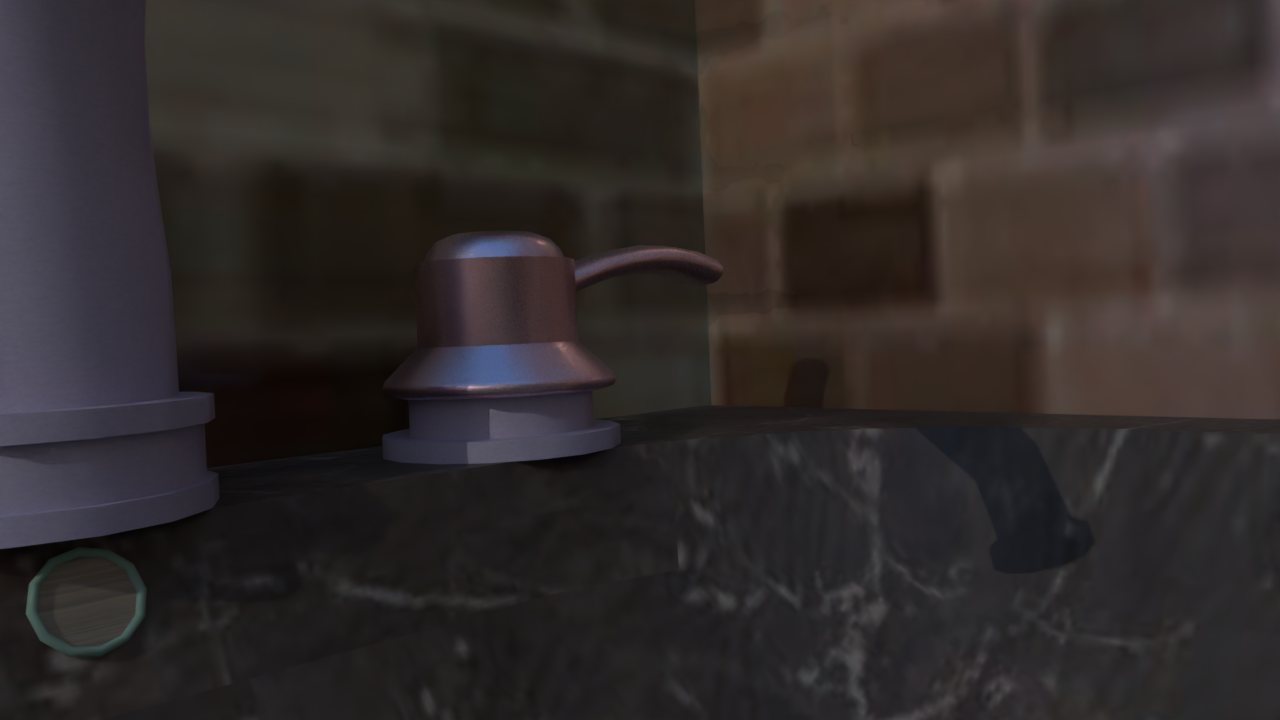} &
        \includegraphics[width=0.138\textwidth,keepaspectratio]{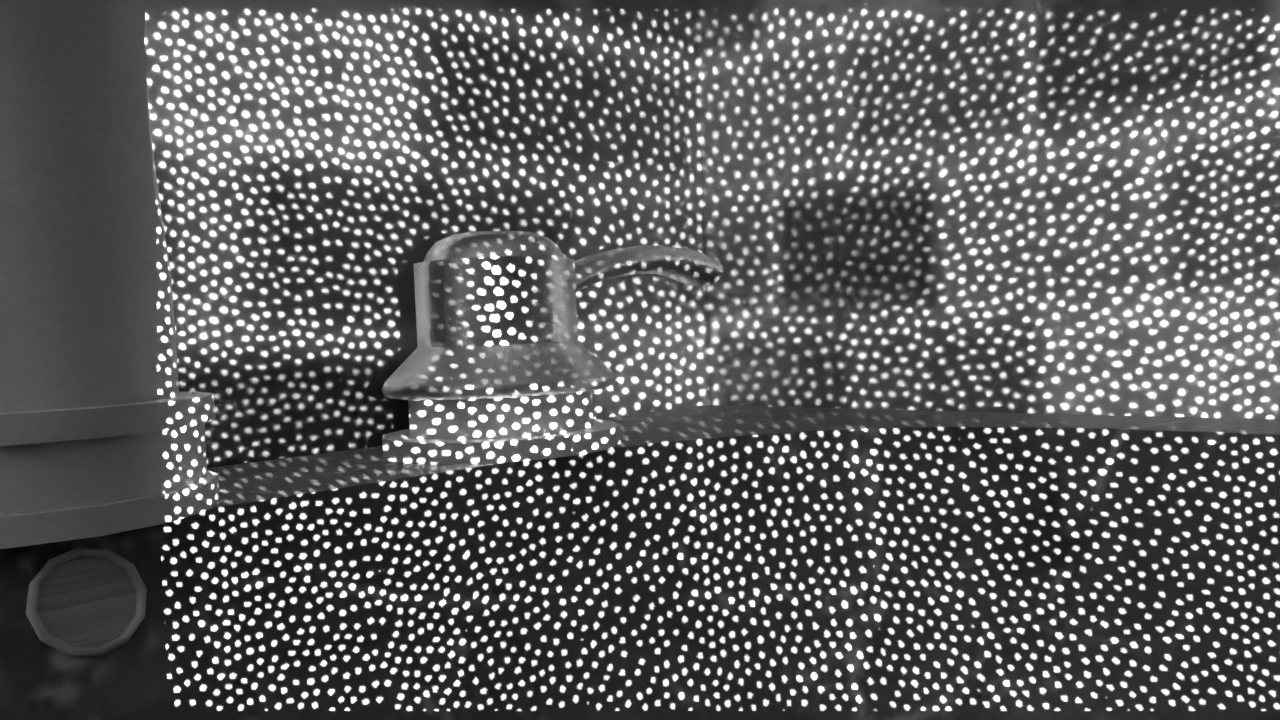} &
        \includegraphics[width=0.138\textwidth,keepaspectratio]{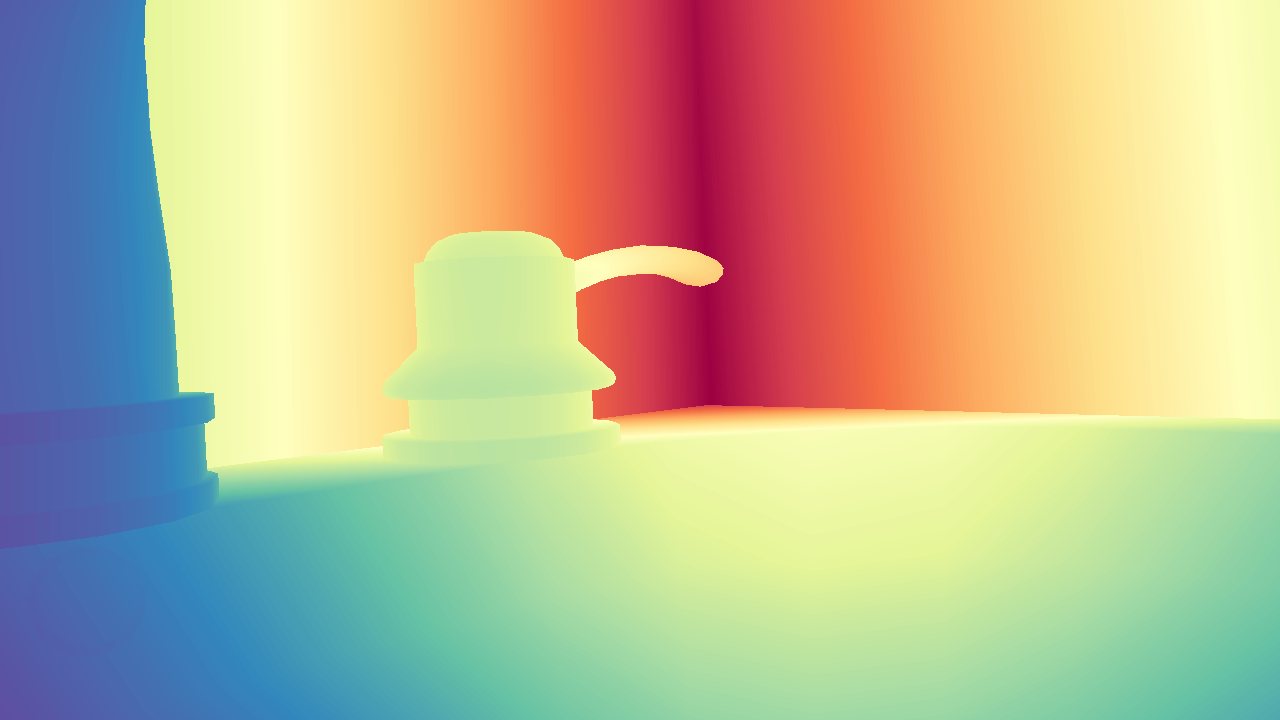} &
        \includegraphics[width=0.138\textwidth,keepaspectratio]{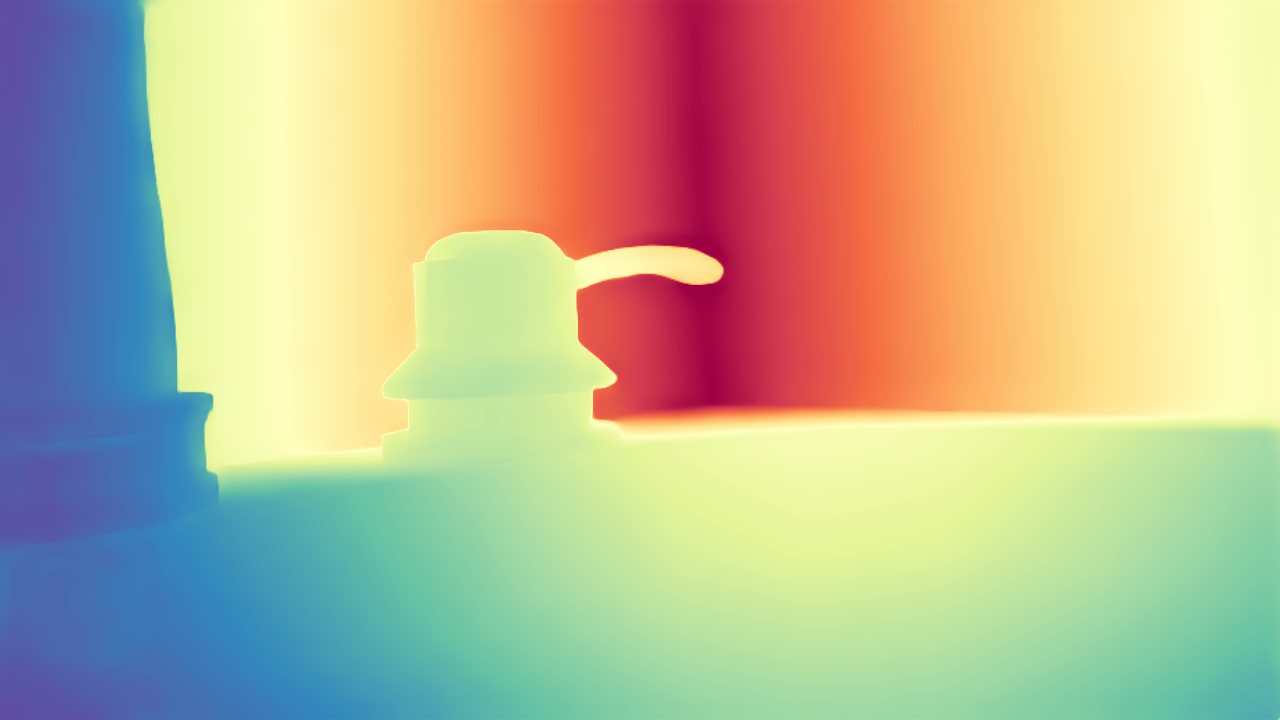} &
        \includegraphics[width=0.138\textwidth,keepaspectratio]{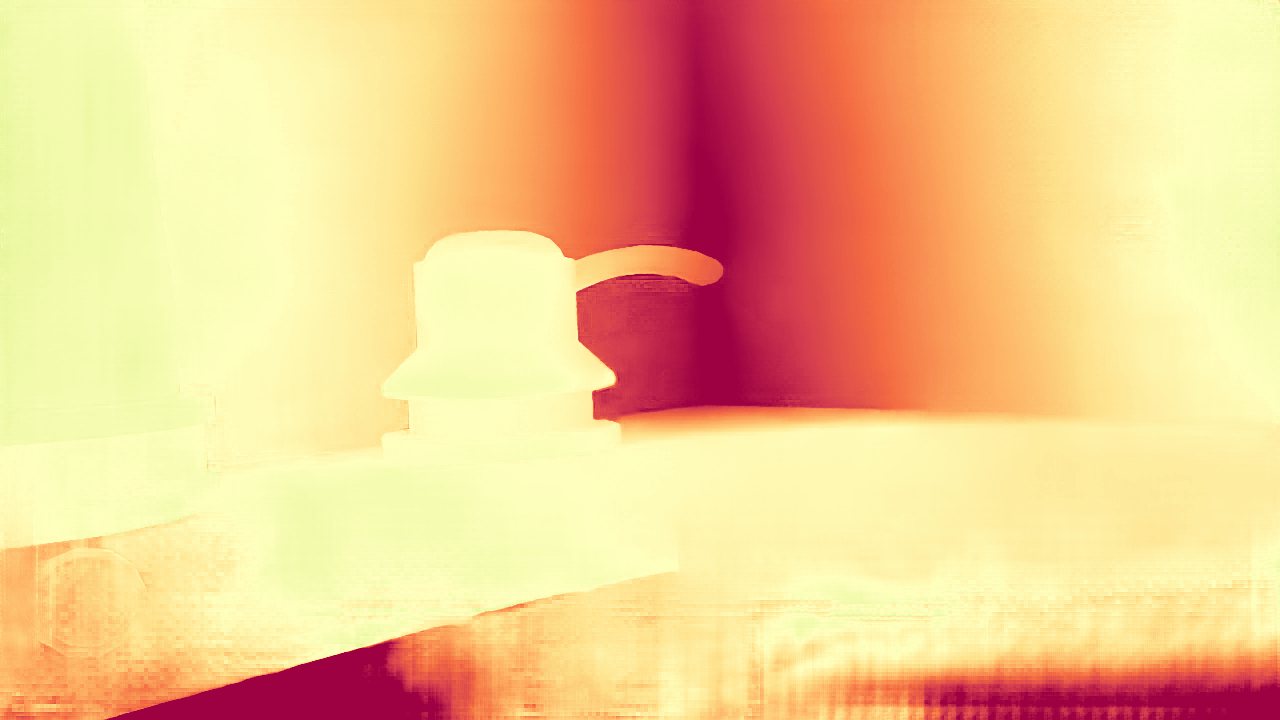} &
        \includegraphics[width=0.138\textwidth,keepaspectratio]{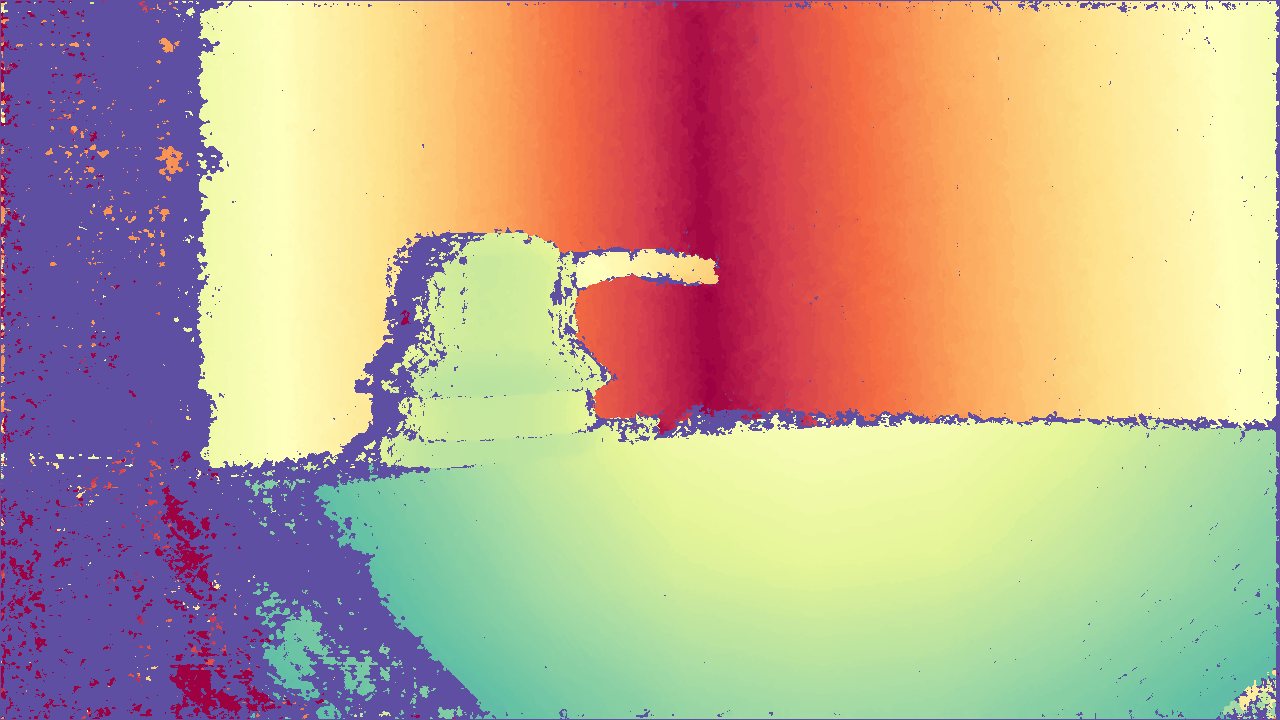} &
        \includegraphics[width=0.138\textwidth,keepaspectratio]{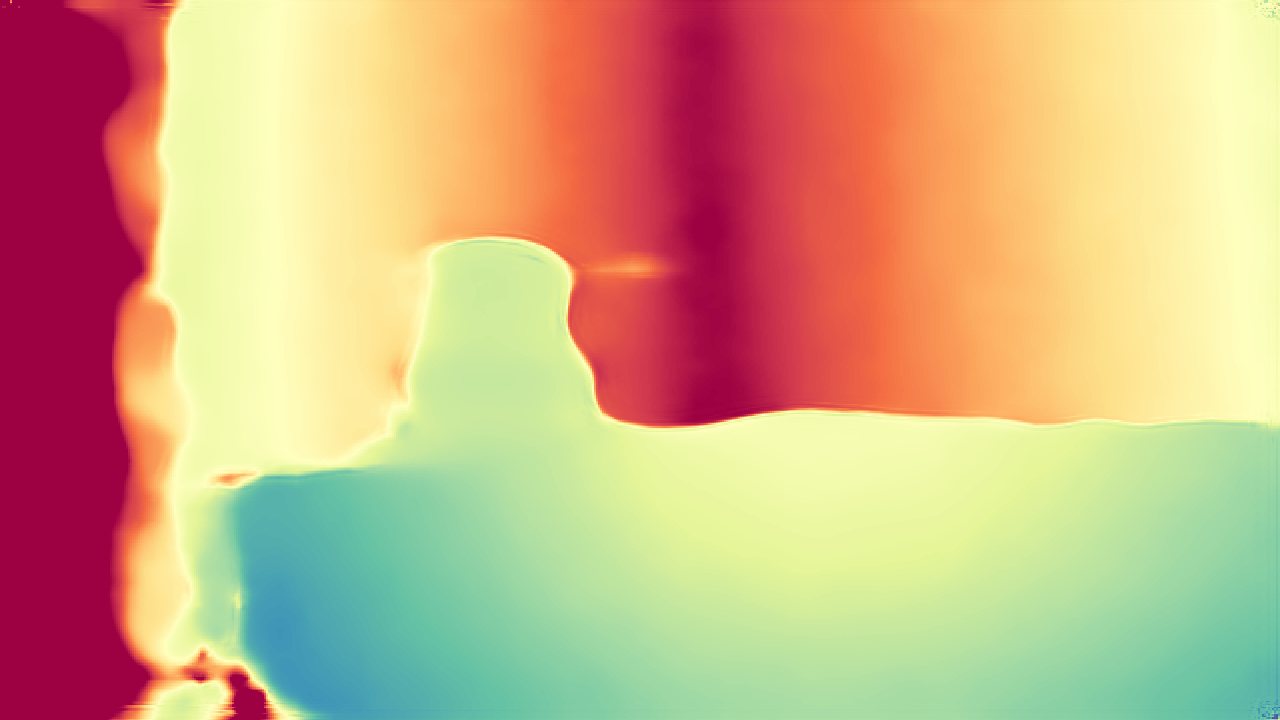} &
        \includegraphics[width=0.0182\textwidth,keepaspectratio]{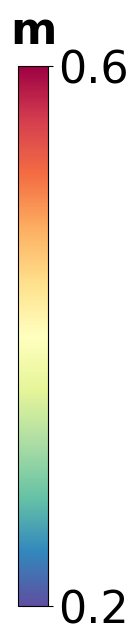}
        \\[2.5pt]

        \multicolumn{8}{c}{\includegraphics[width=1.0\textwidth]{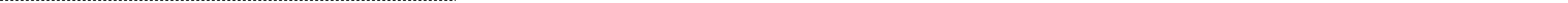}} \\

        \includegraphics[width=0.138\textwidth,keepaspectratio]{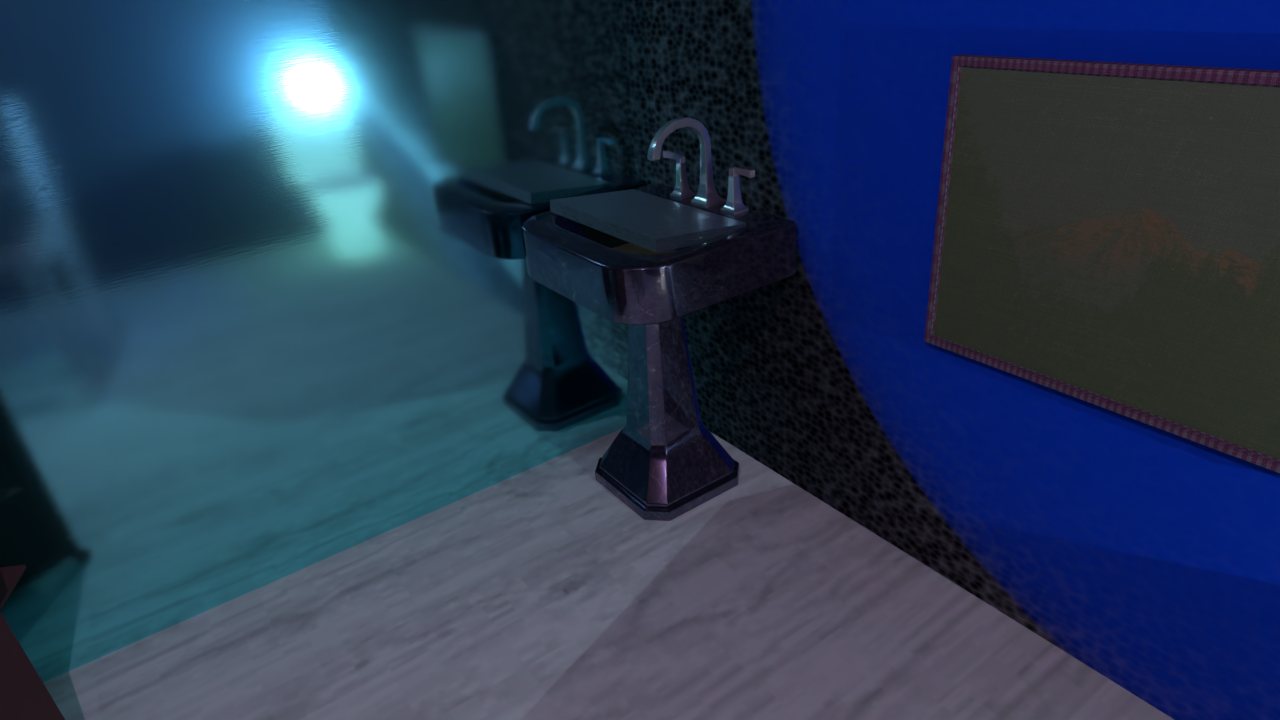}&
        \includegraphics[width=0.138\textwidth,keepaspectratio]{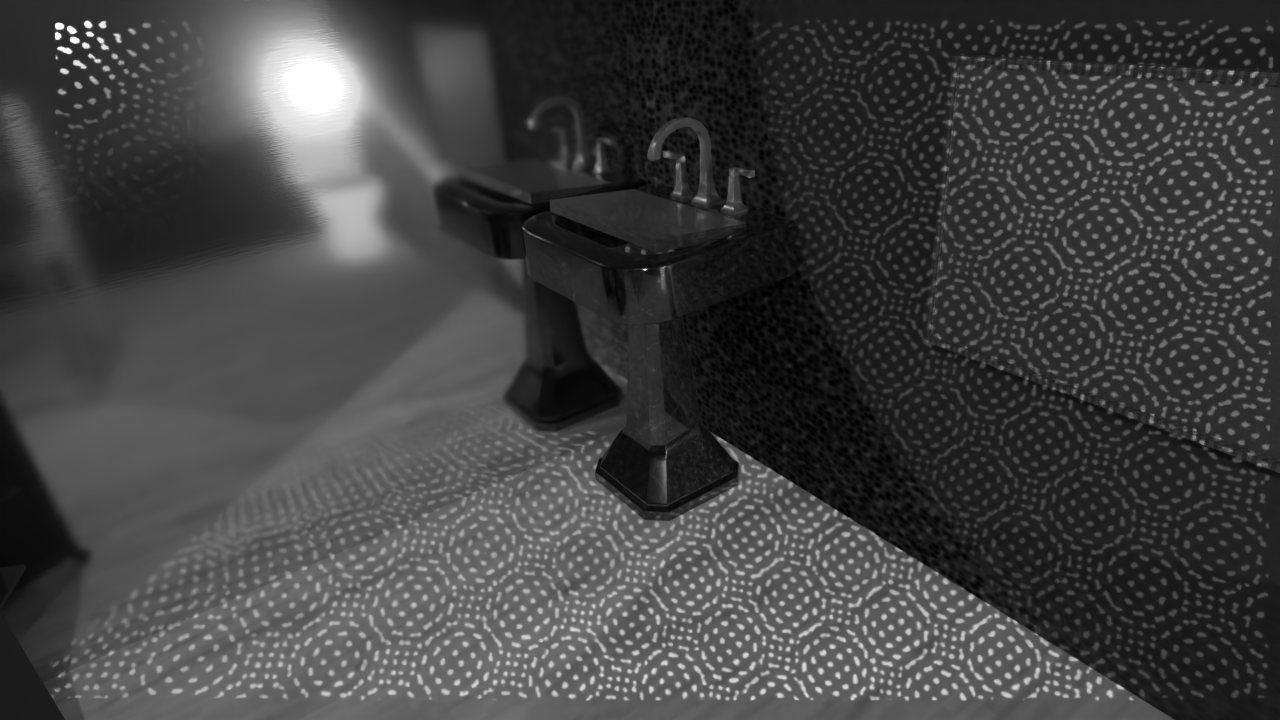} &
        \includegraphics[width=0.138\textwidth,keepaspectratio]{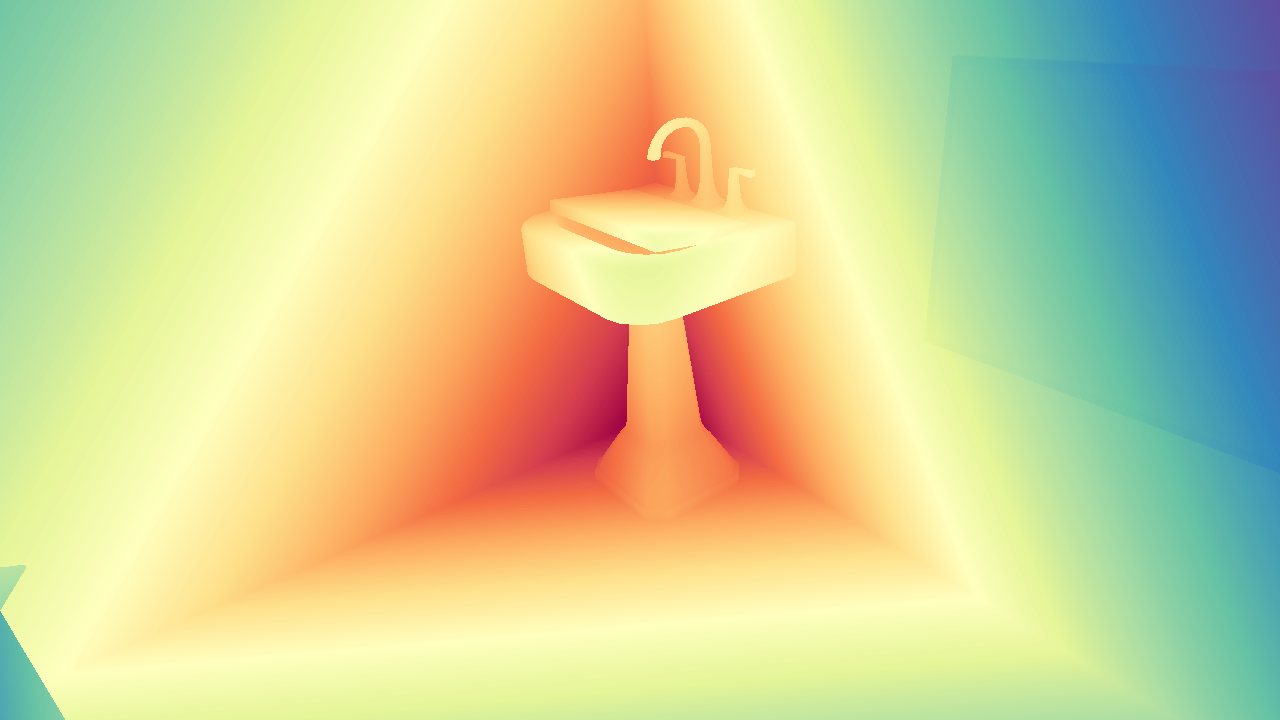} &
        \includegraphics[width=0.138\textwidth,keepaspectratio]{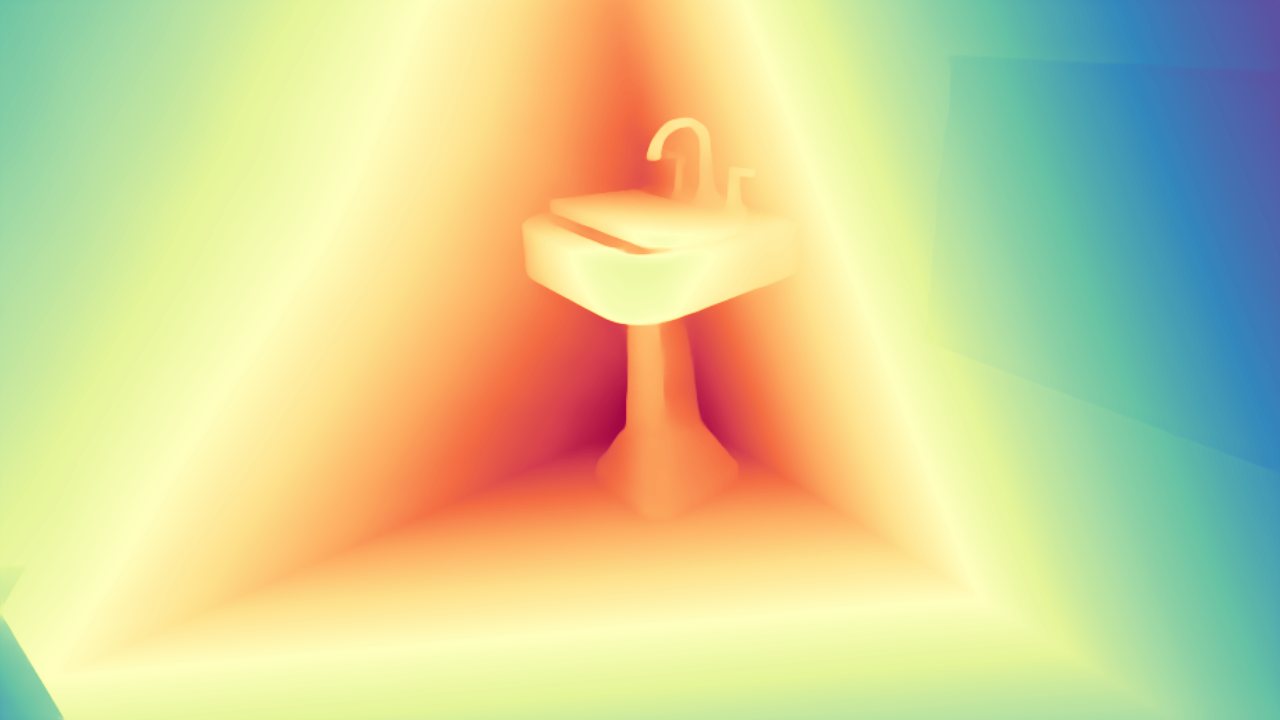} &
        \includegraphics[width=0.138\textwidth,keepaspectratio]{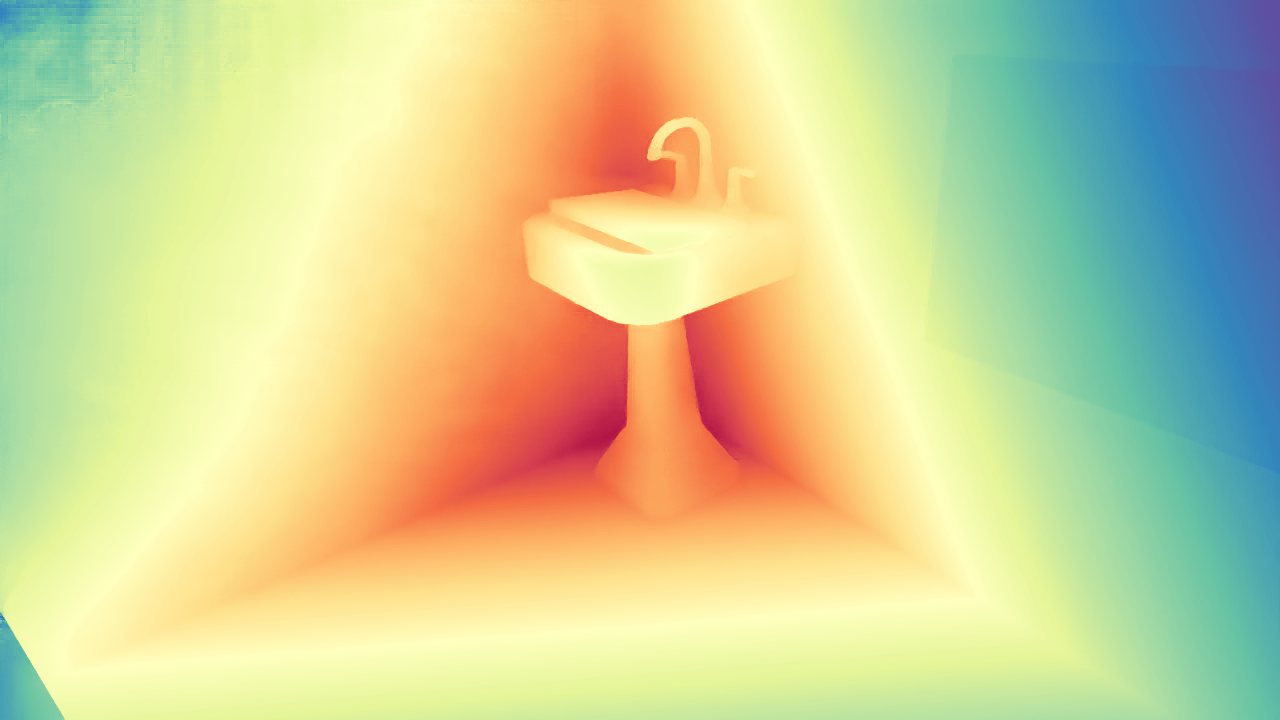} &
        \includegraphics[width=0.138\textwidth,keepaspectratio]{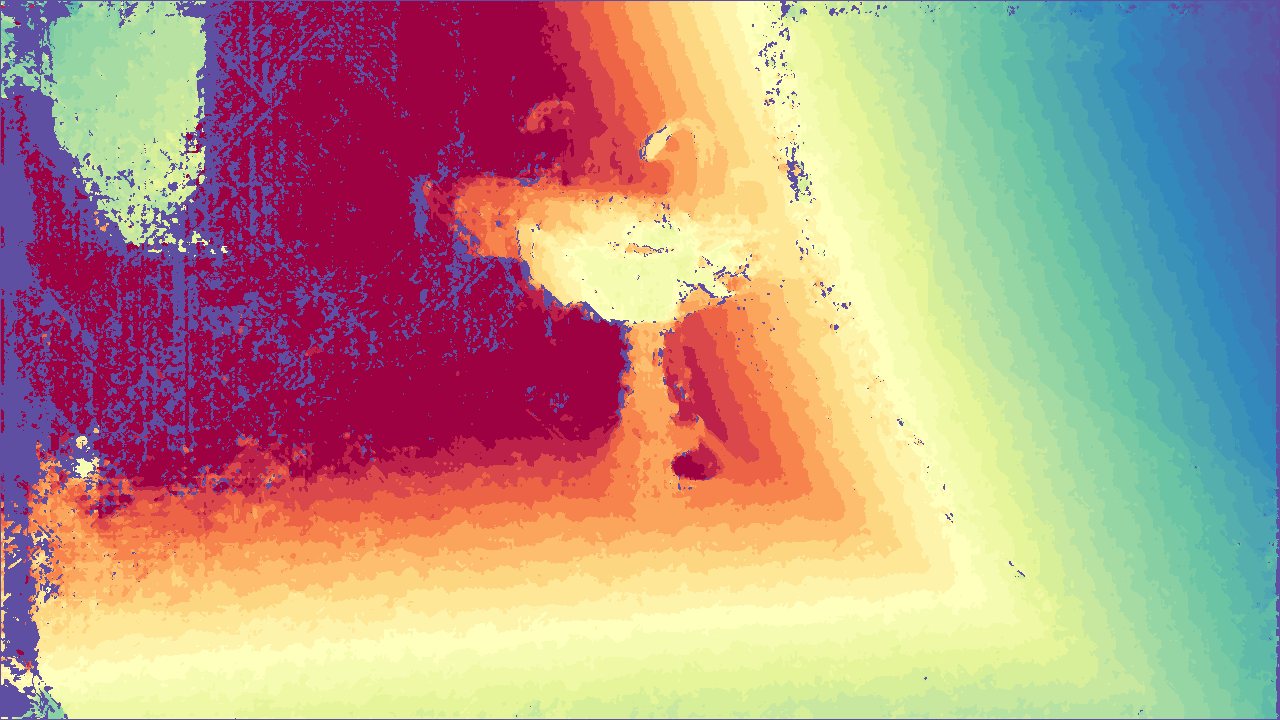} &
        \includegraphics[width=0.138\textwidth,keepaspectratio]{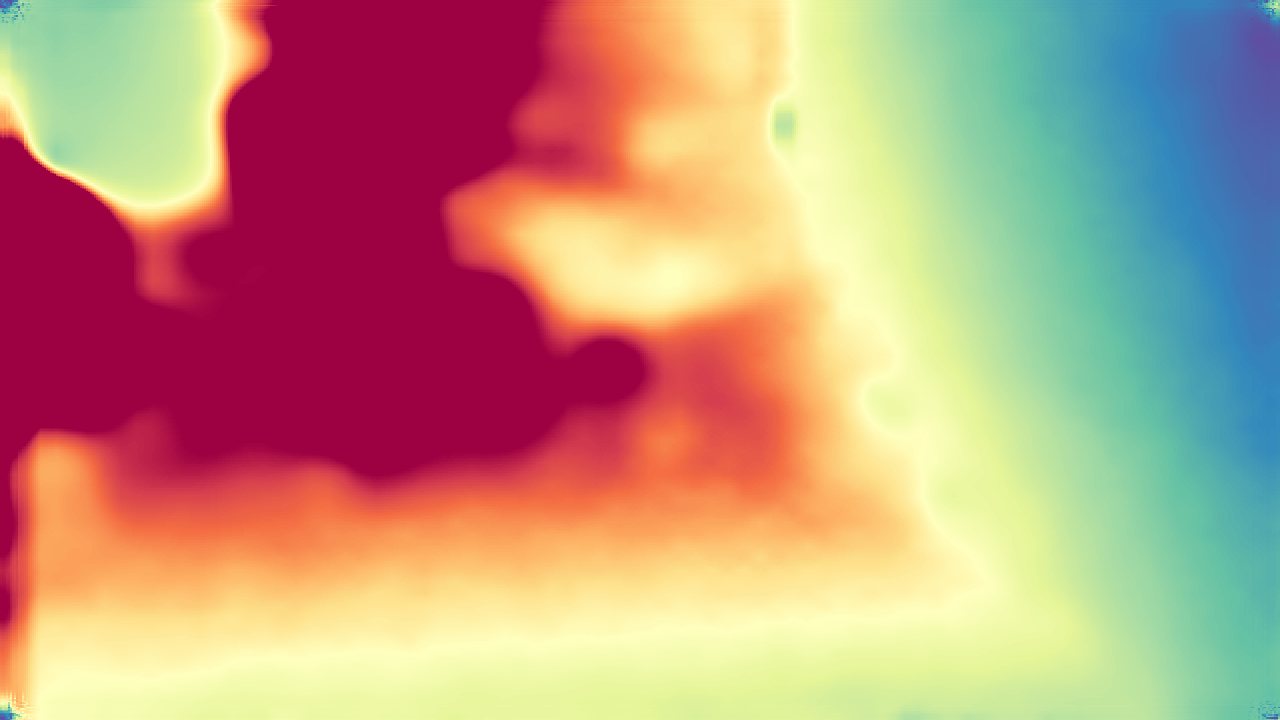} &
        \includegraphics[width=0.0182\textwidth,keepaspectratio]{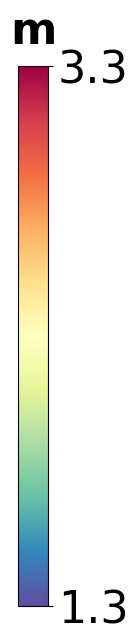} \\
        
        \includegraphics[width=0.138\textwidth,keepaspectratio]{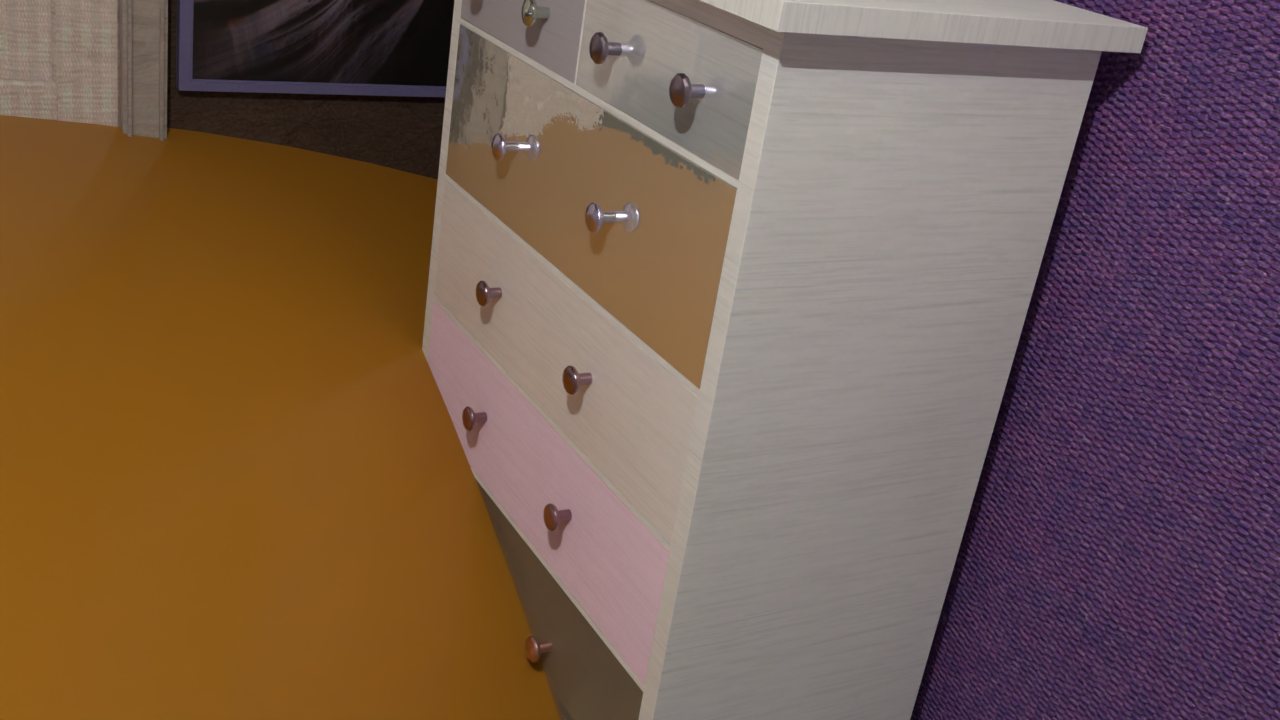} &
        \includegraphics[width=0.138\textwidth,keepaspectratio]{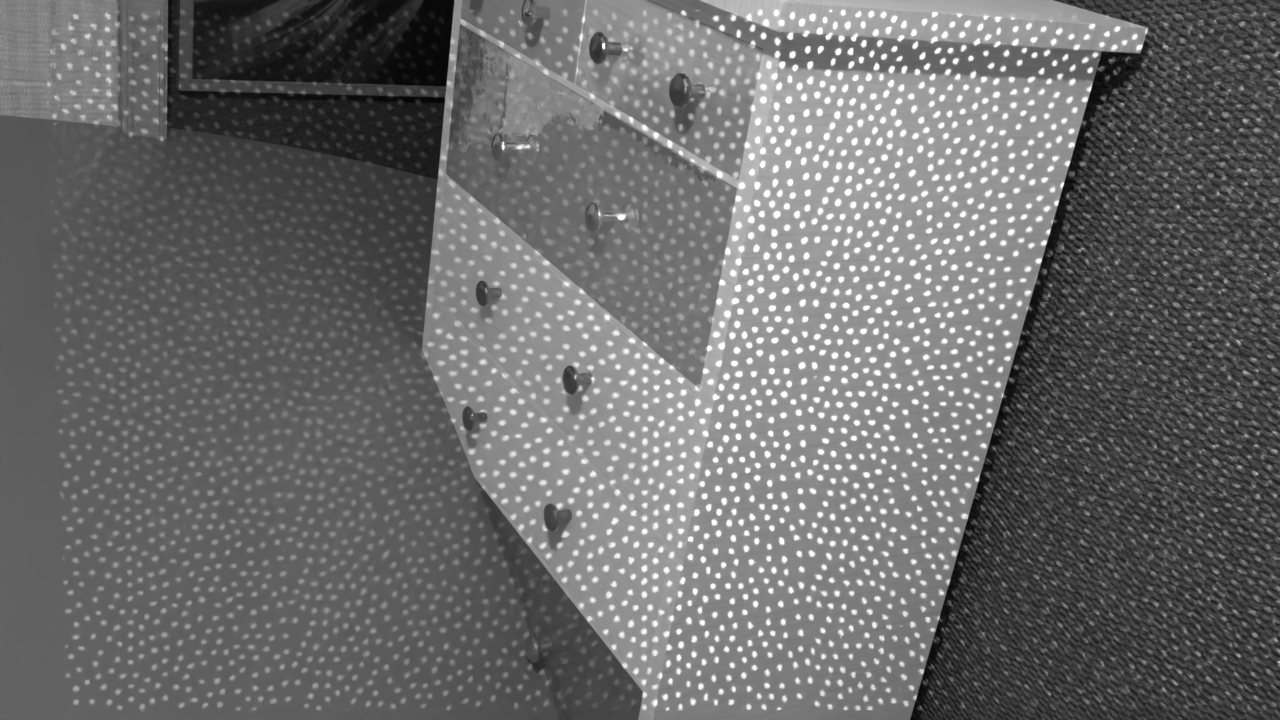} &
        \includegraphics[width=0.138\textwidth,keepaspectratio]{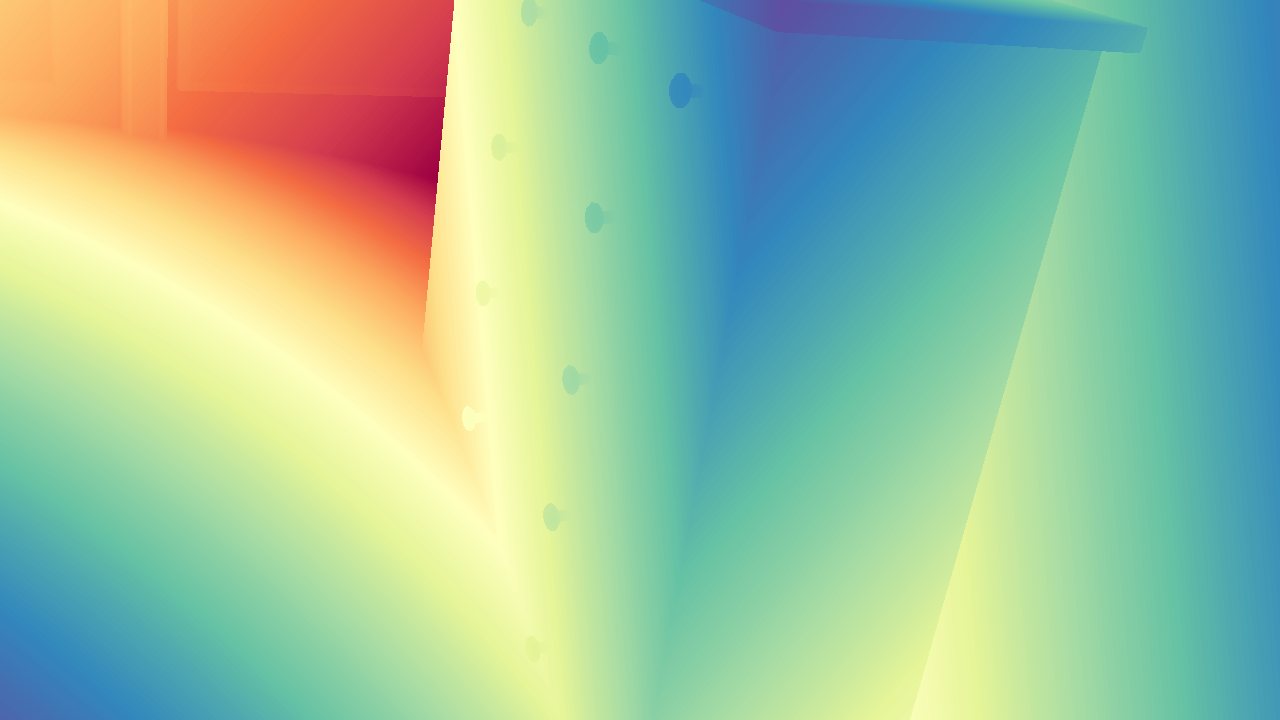} &
        \includegraphics[width=0.138\textwidth,keepaspectratio]{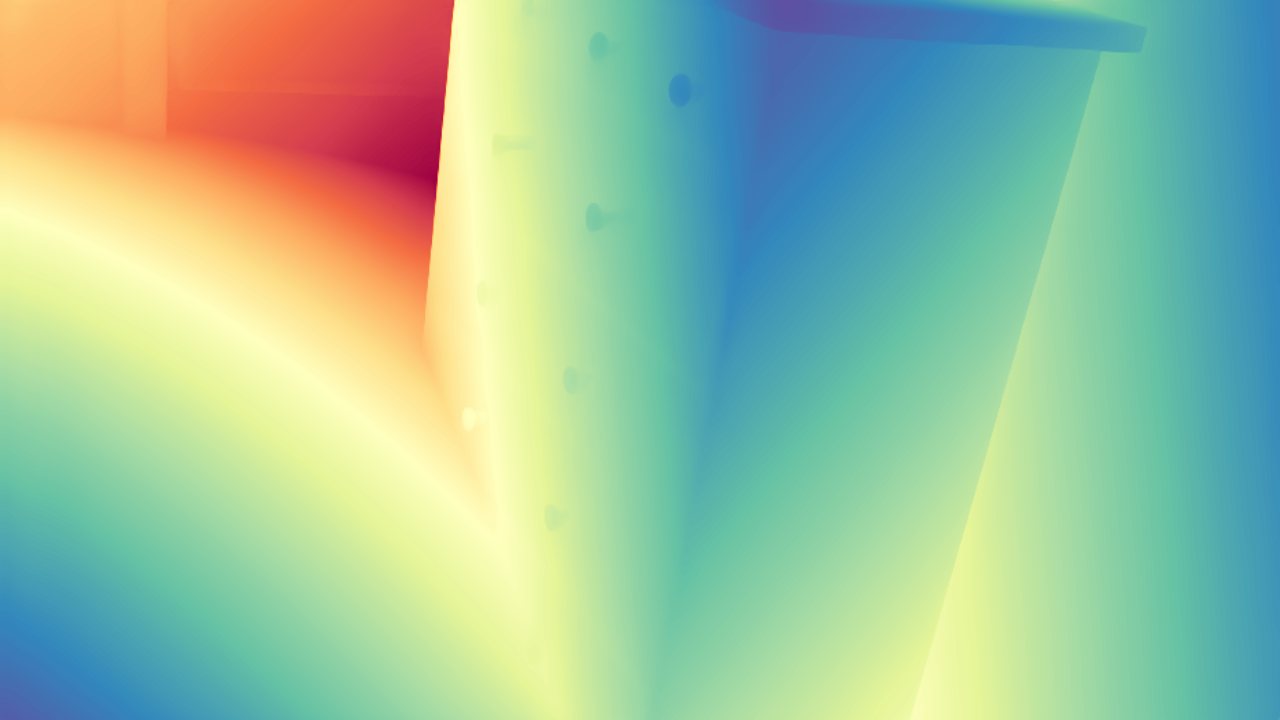} &
        \includegraphics[width=0.138\textwidth,keepaspectratio]{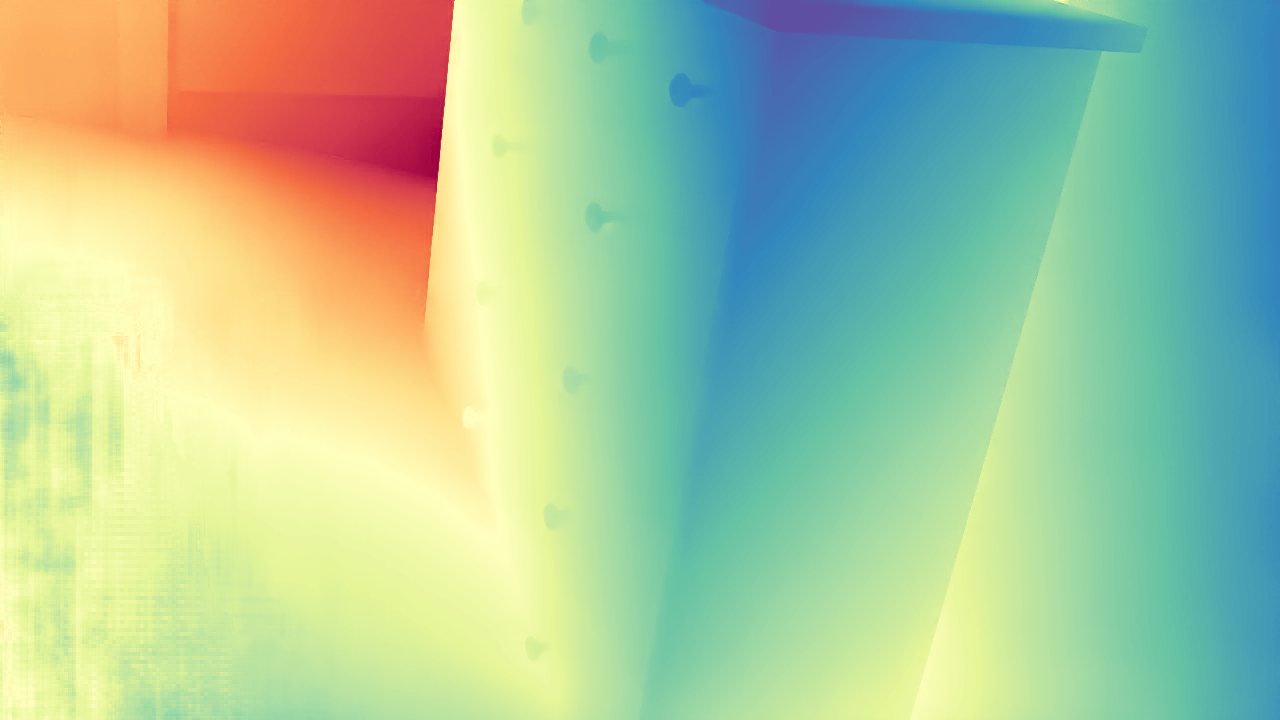} &
        \includegraphics[width=0.138\textwidth,keepaspectratio]{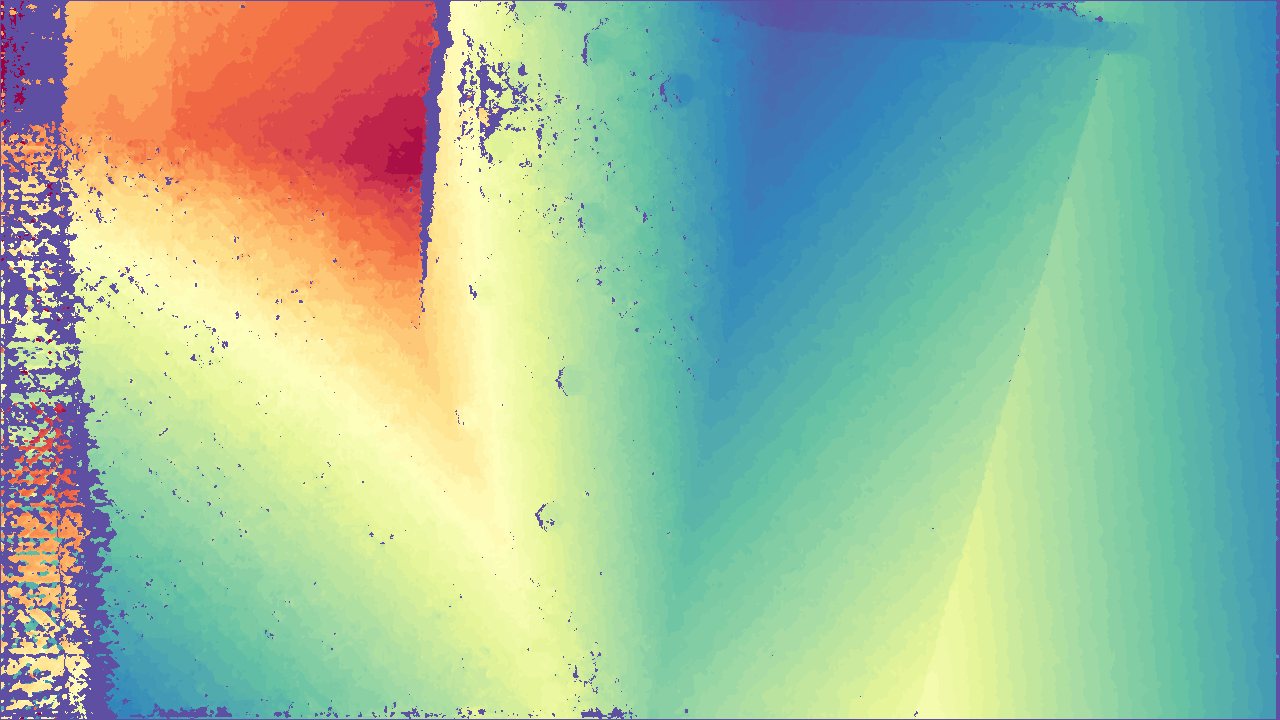} &
        \includegraphics[width=0.138\textwidth,keepaspectratio]{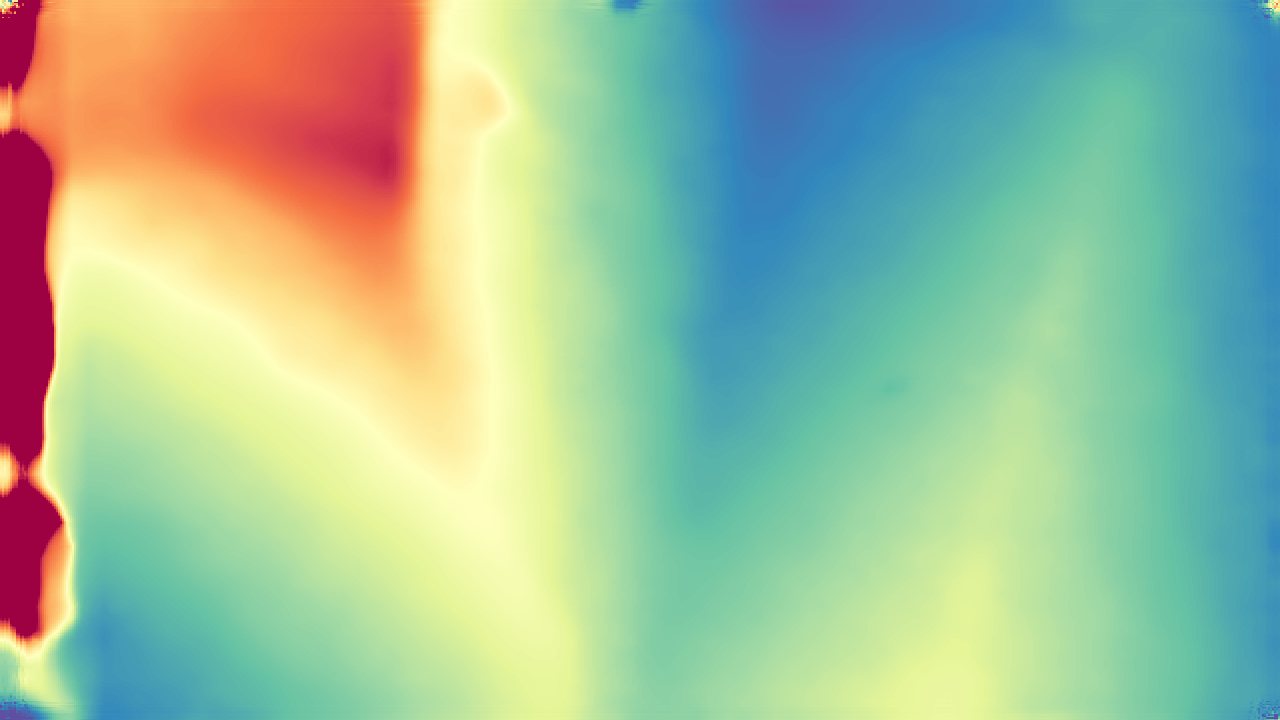} &
        \includegraphics[width=0.0182\textwidth,keepaspectratio]{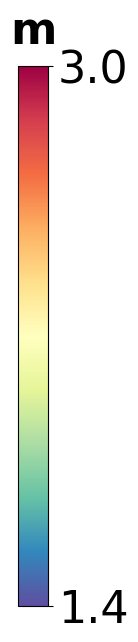} \\
      \end{tabular}
    }
    \caption{\textbf{Qualitative results on our synthetic dataset.}
    Our method (top two rows: $\namem$(single IR+pattern); bottom two: $\nameb$(dual IR+pattern)) is compared with MonSter (dual RGB), ActiveStereoNet (dual IR) and traditional methods (dual IR). It produces more continuous, high-fidelity depth with finer details.
    }
  \label{fig:singleircomp}
\end{figure*}

Qualitative results are shown in Figure~\ref{fig:singleircomp}. Compared to the baselines, our method produces sharper object boundaries, finer structural details, smoother surfaces, and more accurate recovery in textureless and specular regions. Traditional decoding methods rely on low-level pixel matching, which is easily disrupted and unstable under weak signal conditions. ActiveStereoNet fails to resolve fine structures, often producing oversmoothed and indistinct results due to ineffective feature matching. MonSter, lacking pattern-specific local cues, exhibits noticeable depth shifts and missing geometry. In contrast, \name\ maintains strong structural consistency and captures fine geometric details, demonstrating the effectiveness of pattern-aware stereo decoding.

\subsection{Ablation Studies} \label{sec:ablation}

We conduct a series of ablation experiments to analyze the impact of different input configurations, the effectiveness of the monocular depth refinement module, and the generalization capability of our neural decoder across various structured light patterns.

\begin{table}[htbp]
\caption{Comparison of Different Input Configurations on Our Dataset.}
\centering
\resizebox{0.48\textwidth}{!}{%
    \renewcommand{\arraystretch}{1.2}
    \setlength{\tabcolsep}{2pt}
    \begin{tabular}{p{3.0cm} | ccc | ccc}
    \toprule[0.9pt]
    \textbf{Method} & \textbf{MAE}(m) ↓ & \textbf{RMSE} ↓ & \textbf{REL} ↓ & $\boldsymbol{\delta_{1.25}}$ ↑ & $\boldsymbol{\delta_{1.10}}$ ↑ & $\boldsymbol{\delta_{1.05}}$ ↑ \\
    \hline
    \multicolumn{7}{c}{\textbf{Stage 1 without depth refinement module}} \\
    \hline
    $\namem$ {\small(IR+Pat.)} & 0.0599 & 0.3340 & 0.0637 & 0.9612	& 0.9387 & 0.9159 \\
    $\names$ {\small(Dual IR)} & 0.0347 &  0.1005 & 0.0433 & {0.9718} & 0.9539  & 0.9351  \\
    $\nameb$ {\small(Dual IR+Pat.)} & \cellcolor{highlightorange} 0.0253 & 0.1073 & 0.0299 & \cellcolor{highlightorange} 0.9788 & \cellcolor{highlightred} 0.9675 & \cellcolor{highlightred} 0.9560  \\
    \hline
    \multicolumn{7}{c}{\textbf{Stage 2 with depth refinement module}} \\
    \hline
    $\namem$ {\small(IR+Pat.)}  & 0.0408 & 0.0907 & 0.0357 & 0.9704 & 0.9437 & 0.9142  \\
    $\names$ {\small(Dual IR)} & 0.0303 & \cellcolor{highlightorange} 0.0751 & \cellcolor{highlightorange} 0.0251 & 0.9764 & 0.9526 & 0.9258  \\
    $\nameb$ {\small(Dual IR+Pat.)} & \cellcolor{highlightred} 0.0237 & \cellcolor{highlightred} 0.0634 & \cellcolor{highlightred} 0.0210 & \cellcolor{highlightred} 0.9817 & \cellcolor{highlightorange} 0.9666 & \cellcolor{highlightorange} 0.9502 \\
    \bottomrule[0.9pt]
    \end{tabular}
}
\label{tab:ours_vs_variants}
\end{table}

\paragraph{Impact of Input Configurations.}
Our method is very flexible and can work under various settings. To evaluate the importance of different input settings and the role of the structured light pattern in decoding, besides $\namem$ (monucular, single IR + pattern) and $\nameb$ (binocular, dual IR + pattern), we further train a new model that only takes dual IR as input (only with captured SL images, without SL pattern), noted as \textbf{$\names$ (stereo, dual IR only)}.  
As shown in Table~\ref{tab:ours_vs_variants}, regardless of whether the monocular depth refinement module is included, the trend consistently holds: $\texttt{IR + Pattern} < \texttt{Dual IR} < \texttt{Dual IR + Pattern}$. 

Notably, even in the stereo setup where complete binocular cues are available, incorporating the projected pattern still brings a clear performance gain. This demonstrates that patterns are not redundant—even with dual-view geometry—and should not be ignored.

\renewcommand{\arraystretch}{0.05}

\setlength{\imgwidth}{0.162\textwidth}

\renewcommand{\formattedgraphics}[1]{%
      \includegraphics[width=\imgwidth,keepaspectratio]{#1}%
}

\setlength{\colorbarwidth}{0.0182\textwidth} 
\renewcommand{\colorbar}[1]{%
      \includegraphics[width=\colorbarwidth, keepaspectratio]{#1}%
}

\begin{figure}[htb]
  \centering
  \resizebox{0.5\textwidth}{!}{%
      \begin{tabular}{cccc}
    
        \textbf{Scene} & \textbf{IR} & \textbf{GT Depth} \\[0.8pt]
        \includegraphics[width=0.162\textwidth,keepaspectratio]{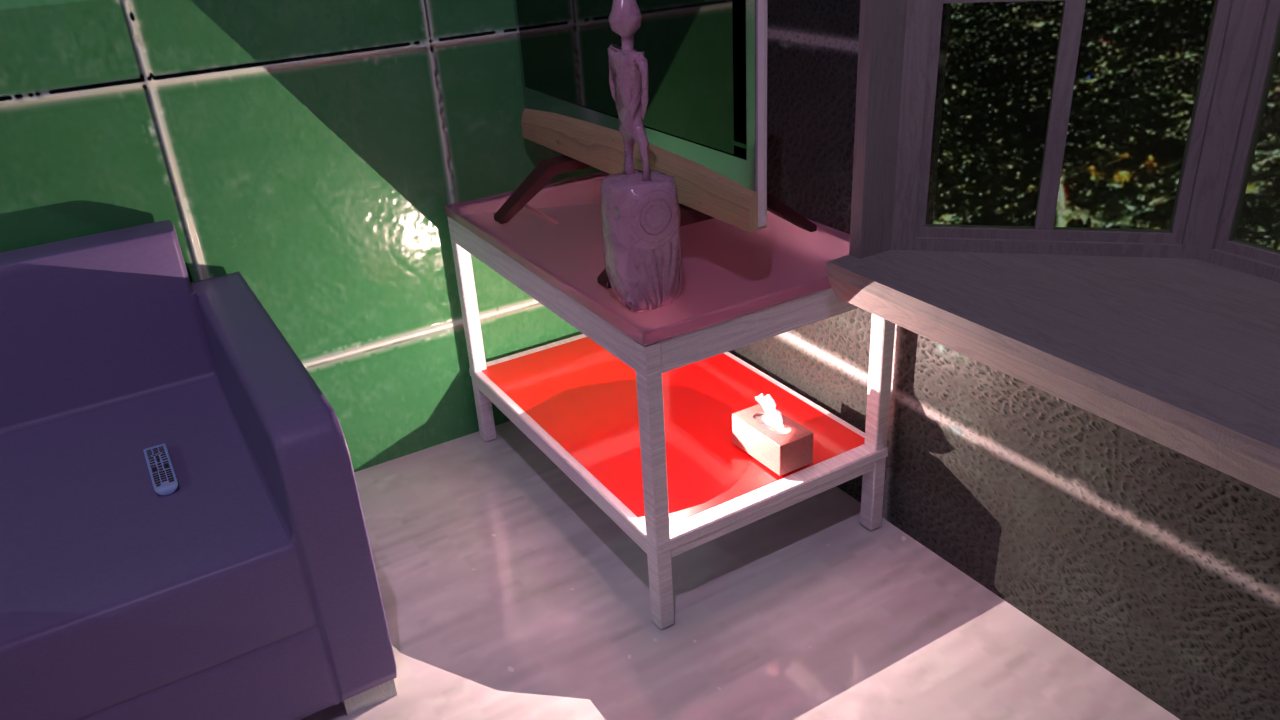} &
        \includegraphics[width=0.162\textwidth,keepaspectratio]{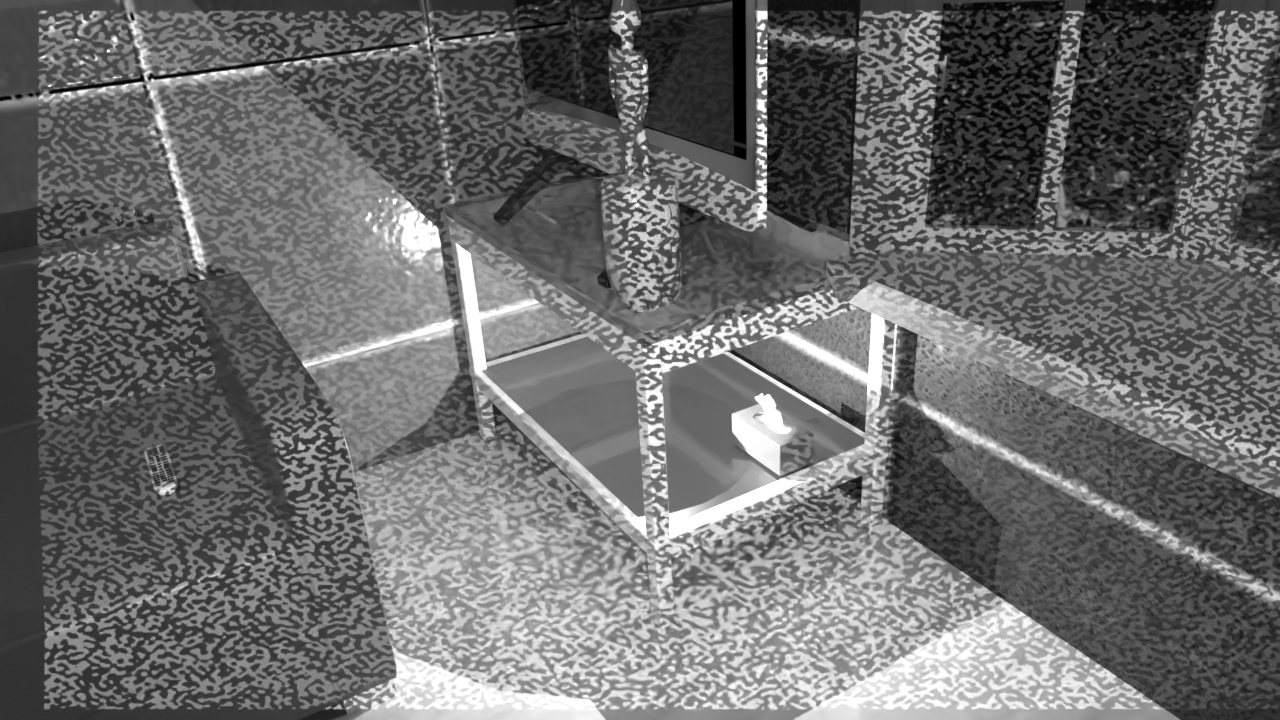} &
        \includegraphics[width=0.162\textwidth,keepaspectratio]{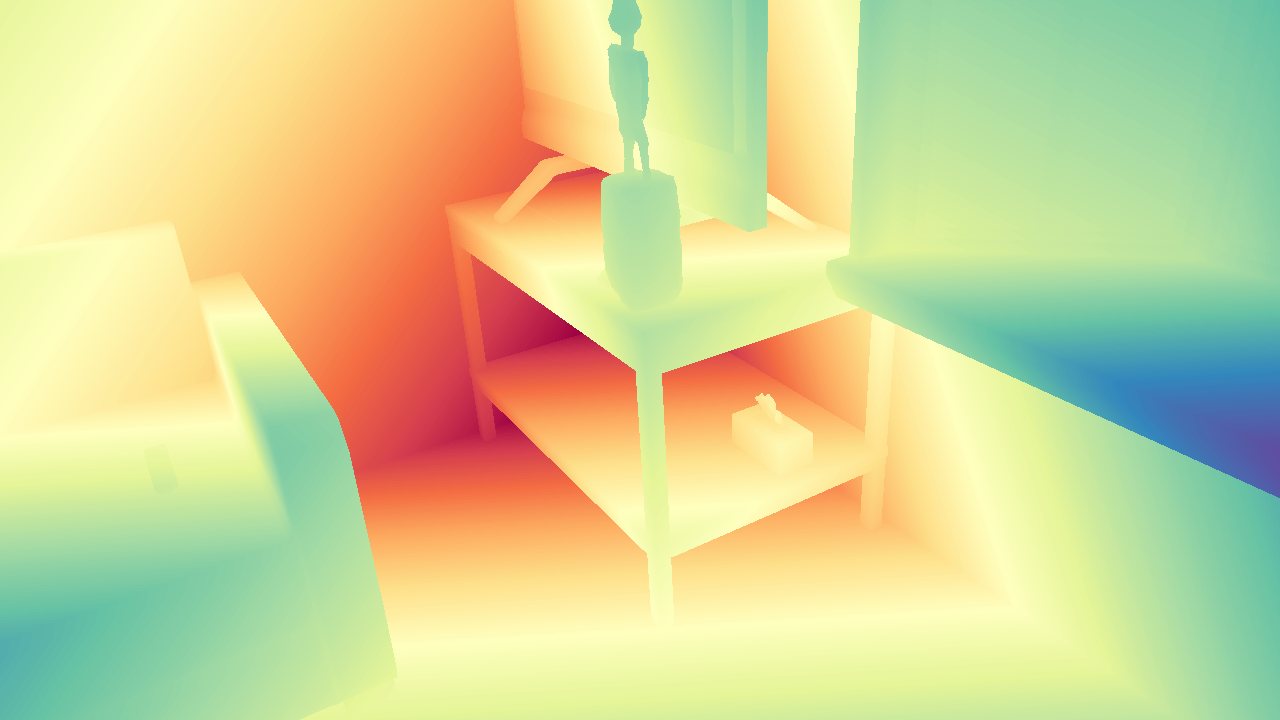} &
        \includegraphics[width=0.0182\textwidth,keepaspectratio]{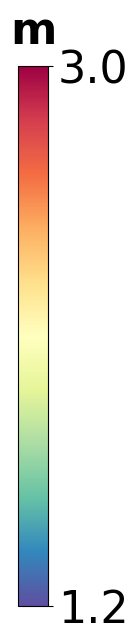}
        \\[0.8pt]
        \textbf{Single IR+Pattern} & \textbf{Dual IR} & \textbf{Dual IR+Pattern}
        \\[0.8pt]
        \includegraphics[width=0.162\textwidth,keepaspectratio]{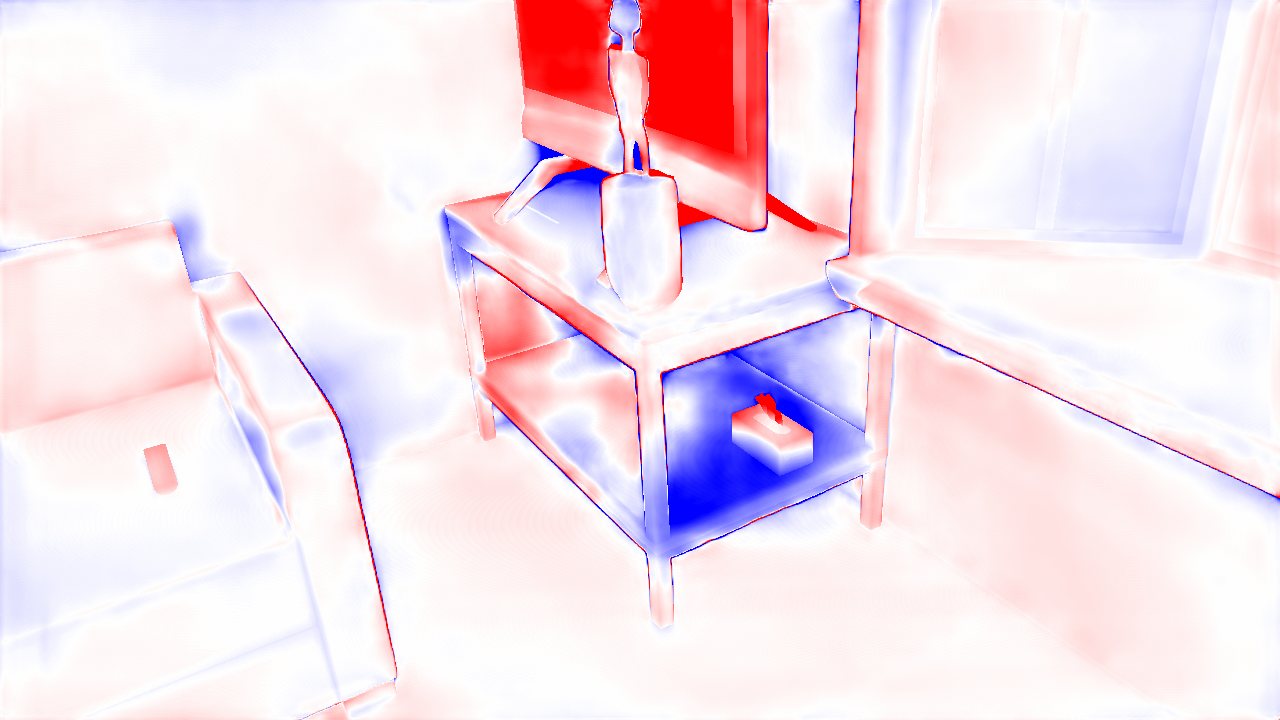} &
        \includegraphics[width=0.162\textwidth,keepaspectratio]{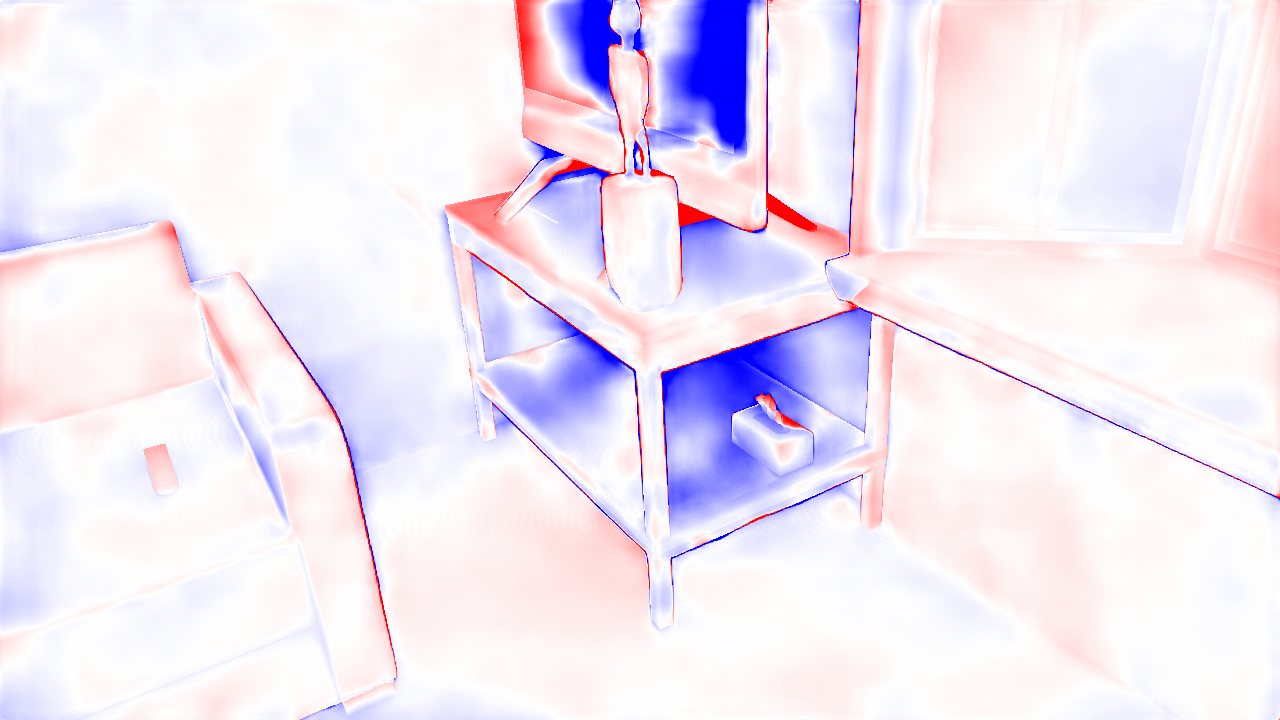} &
        \includegraphics[width=0.162\textwidth,keepaspectratio]{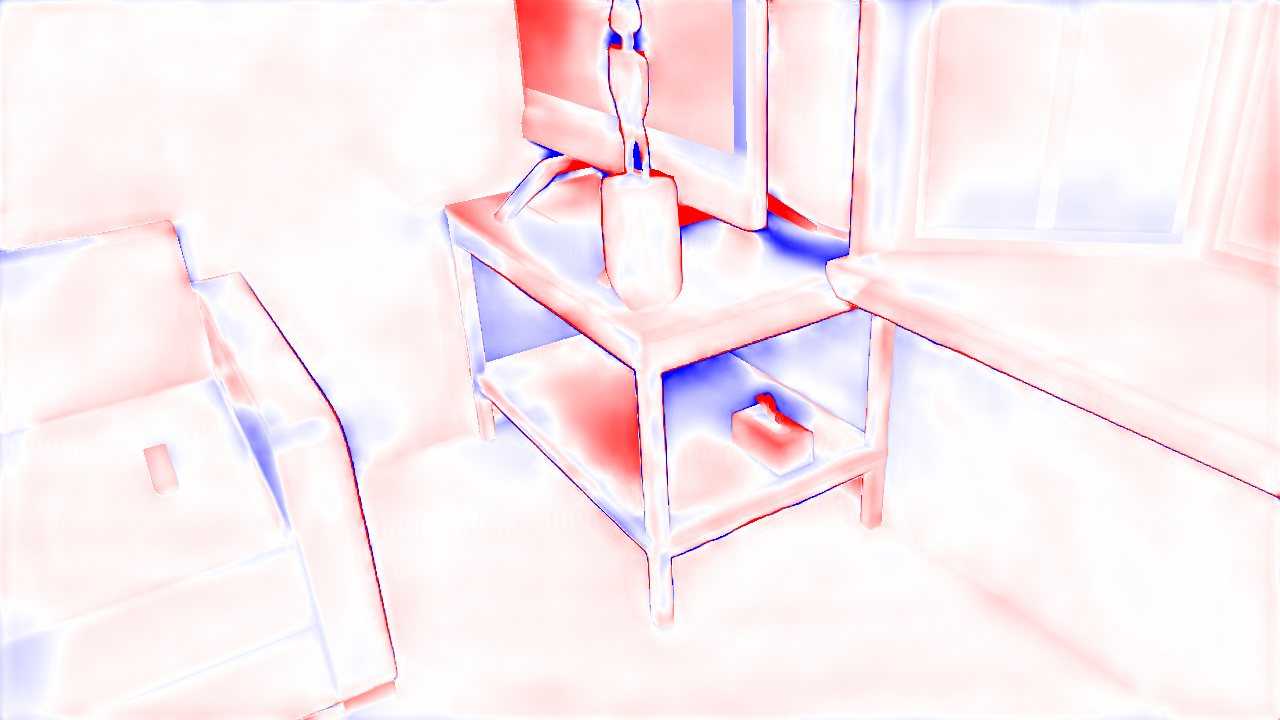} &
        \includegraphics[width=0.0182\textwidth,keepaspectratio]{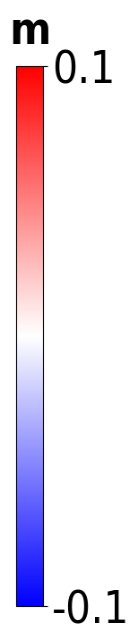}
        \\
      \end{tabular}
  }
  \caption{\textbf{Comparison of Different Input of \name.} The second row shows error maps. The depth error with pattern input (dual IR+Pattern) is lower than without (dual IR), validating the effectiveness of incorporating pattern features for structured light decoding.}
  \label{fig:abal-diff-input}
\end{figure}
\renewcommand{\arraystretch}{1}

\paragraph{Effectiveness of Monocular Depth Refinement.}
We compare results with and without the monocular refinement module across all three input settings (Rows 1–3 vs. 4–6 in Table~\ref{tab:ours_vs_variants}; see also Figure~\ref{fig:abla_mdr_module}). The refinement consistently improves performance, especially in the monocular IR + pattern setting, where global priors help compensate for missing stereo cues.
A slight drop in $\delta_{1.1}$ and $\delta_{1.05}$ is observed, likely due to bias in the initial estimate $\depthInit$. While the module reduces MAE and RMSE by refining geometry and suppressing outliers (indicated by decreased MAE and RMSE), it may also propagate initial bias more broadly—highlighting a direction for future improvement.

\paragraph{Generalization Across Patterns.}
During training, we do not distinguish between structured light patterns and mix all 6 training patterns to improve robustness. To evaluate generalization, we test the trained model separately on each pattern in the test set, including two novel patterns (\texttt{Kinect} and \texttt{RandomSquare}) that were not seen during training. Results are summarized in Table~\ref{tab:performance_comparison}. Full table can be found in the appendix. We observe only mild performance variation across patterns, indicating that our model has generalization capability to unseen patterns.

\begin{table}
\centering
\caption{\namem's Performance Comparison of Different Patterns.}
\resizebox{0.48\textwidth}{!}{%
    \renewcommand{\arraystretch}{1.2}
    \setlength{\tabcolsep}{2pt}
    \begin{tabular}{p{2.0cm} | ccc | ccc}
    \toprule[0.9pt]
    \textbf{Method} & \textbf{MAE(m)} ↓ & \textbf{RMSE} ↓ & \textbf{REL} ↓ & $\boldsymbol{\delta_{1.25}}$ ↑ & $\boldsymbol{\delta_{1.10}}$ ↑ & $\boldsymbol{\delta_{1.05}}$ ↑ \\
    \hline
    \multicolumn{7}{c}{\textbf{Training Patterns} } \\
    \hline
    D415  & 0.0388 & 0.0877 & 0.0354 & 0.9716 & 0.9458 & 0.9180 \\
    
    D435 & 0.0389 & 0.0878 & 0.0361 & 0.9721 & 0.9474 & 0.9195 \\
    
    Alacarte & 0.0415 & 0.0913 & 0.0356 & 0.9661 & 0.9370 & 0.9068 \\
    \hline
    \multicolumn{7}{c}{\textbf{Testing Patterns} } \\
    \hline
    Randsquare & 0.0395 & 0.0883 & 0.0338 & 0.9709 & 0.9450 & 0.9157 \\
    \bottomrule[0.9pt]
    \end{tabular}
}
\label{tab:performance_comparison}
\end{table}

\subsection{Real-World Evaluation} \label{sec:real_results}
\paragraph{Hardware Setup.}
To validate the effectiveness of our method in real-world settings, we build a custom structured light capture system. The setup includes a Lenovo T8s projector and HikVision CS050-10UC left and right cameras. All components are calibrated using standard multi-view calibration techniques~\cite{opencv} to estimate intrinsic and extrinsic parameters.

In generating synthetic dual-view structured light data, we assume ideal optical axis alignment between the projector and both cameras, enabling perfect epipolar rectification for both dual IR and pattern inputs. However, such precise alignment is difficult to achieve with off-the-shelf projector-camera hardware. As a result, we evaluate only the Single IR + pattern and Dual IR settings on real-world captures.

In all real-data experiments, our method uses the model trained solely on the proposed synthetic dataset, without any additional training or fine-tuning. \added{The evaluation resolution is $2040\times1050$.}

\renewcommand{\arraystretch}{0.05}

\setlength{\imgwidth}{0.2\textwidth}

\renewcommand{\formattedgraphics}[1]{%
      \includegraphics[width=\imgwidth,keepaspectratio]{#1}%
}

\begin{figure}[htb]
  \centering
  \begin{tabular}{cc}
    \textbf{Setup} & \textbf{Scene}\\[0.8pt]
    \includegraphics[width=0.2\textwidth,keepaspectratio]{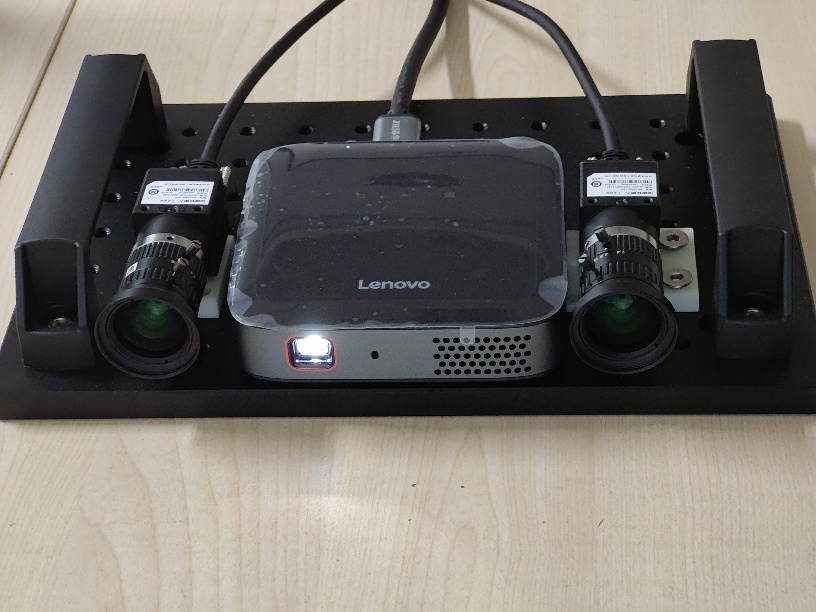} &
    \includegraphics[width=0.2\textwidth,keepaspectratio]{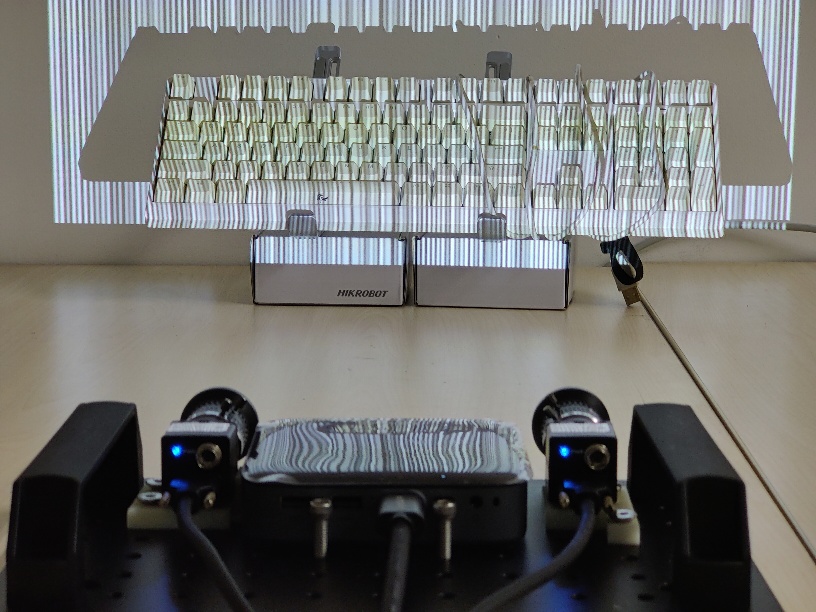} \\
  \end{tabular}
  \caption{\textbf{Hardware.} Our system consists of a Lenovo T8s projector and HikVision CS050-10UC left and right cameras.}
  \label{fig:hardware}
\end{figure}

\paragraph{Quantitative Real Results.}
We place the hardware on the desk and capture 11 real-world indoor scenes. We first apply traditional multi-shot structured light methods to estimate the GT depth. Pseudo ground truth depth maps were obtained by projecting 32 alacarte patterns and decoding using a ZNCC-based method~\cite{alacarte}.  We further create a GT mask to get rid of the occluded regions and outliers.

\begin{figure*}[!t]
\centering
\includegraphics[width=0.95\textwidth]{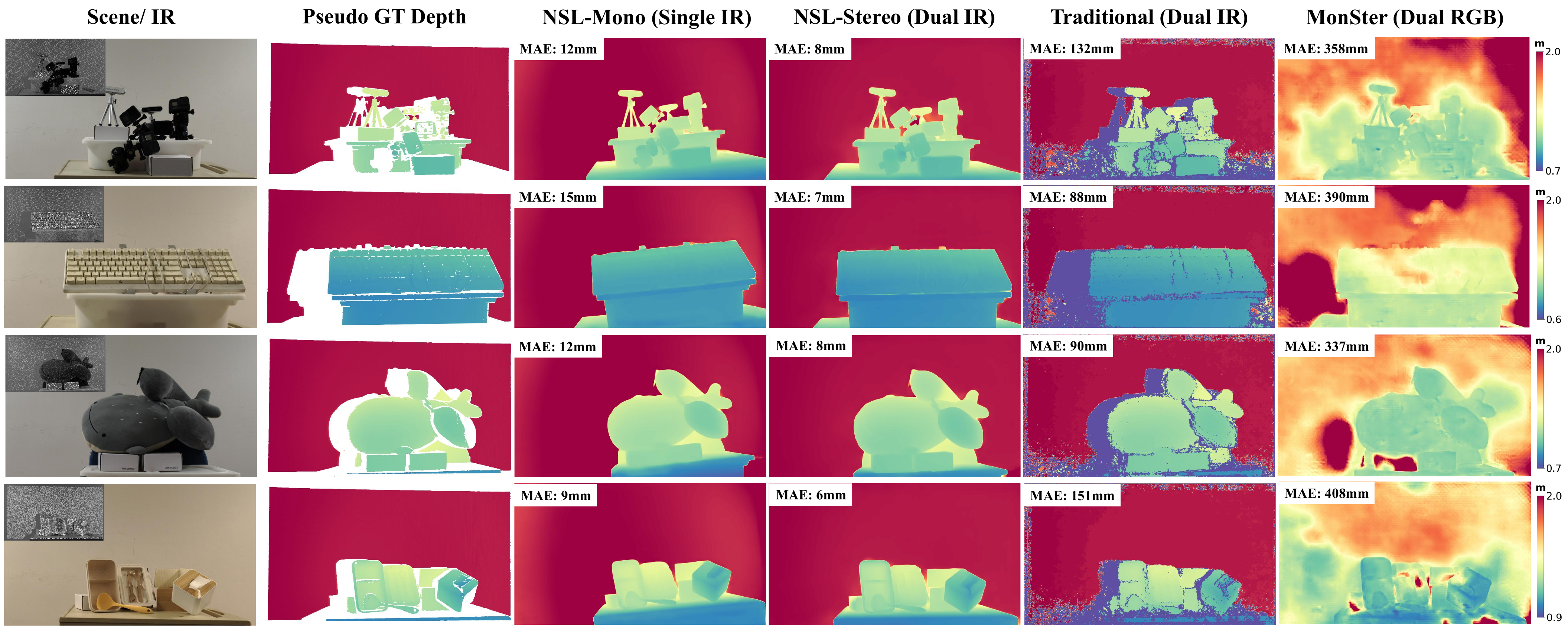}
\caption{\textbf{Quantitative results on real scenes.} Our method achieves 14mm ($\namem$) and 10mm ($\names$) accuracy in known-depth regions. It also delivers better details, smoother surfaces, and more complete depth in depth-unknown regions. The viewpoint difference between single IR and dual IR results stems from distinct epipolar rectifications: one for the camera-projector pair (monocular SL) and the other for the stereo IR camera setup (binocular SL).
}
\label{fig:real_quan}
\end{figure*}

For evaluation, we captured all 8 structured light patterns from our dataset for each scene, resulting in a total of 88 structured light image pairs (11 * 8 = 88). All pattern types were included in the evaluation. Additionally, we captured passive stereo RGB pairs without pattern projection for each scene to enable comparison with RGB-based stereo methods.
To ensure a fair comparison, we report both the zero-shot (z.s.) and fine-tuned (f.t.) results of MonSter.
Fig.~\ref{fig:real_quan} and Tbl.~\ref{tab:real_quan} present the results. 

Several interesting points can be observed. First, traditional decoding performs relatively well here compared to synthetic settings, potentially because pseudo ground truth is only available in unoccluded regions with non-reflective, non-transparent surfaces, and we evaluate only on these regions. 
Both ActiveStereoNet and MonSter fail under real-world conditions—the latter in particular suffers from severe scale inconsistencies (notably low $\delta_{1.05}$), which cannot be attributed to fine-tuning on synthetic data. In contrast, our method, even trained on synthetic data, demonstrates accurate, robust, and generalizable performance.

\begin{table}[t!] 
\centering
\caption{Quantitative Results on Real Data.}
\vspace*{-0.3cm}
\label{tab:real_quan} 
\renewcommand{\arraystretch}{1}
\resizebox{0.48\textwidth}{!}{%
    \begin{tabular}{p{1.5cm} c c c | p{1.5cm} c c c}
    \toprule
    Methods & \textbf{MAE(m)} & \textbf{$\delta_{1.05}$} & \added{EPE} & Methods & \textbf{MAE(m)} & \textbf{$\delta_{1.05}$} & \added{EPE} \\
    \midrule
    \textbf{\small $\namem$} & \cellcolor{highlightorange} 0.014 & \cellcolor{highlightorange} 0.964 & \cellcolor{highlightorange} \added{1.80} & TM & 0.111 & 0.901 & \added{12.55} \\
    \textbf{\small $\names$} & \cellcolor{highlightred} 0.010 & \cellcolor{highlightred} 0.987 & \cellcolor{highlightred} \added{1.74} & ASN & 0.382 & 0.588 & \added{60.71} \\
    MonSter(z.s.) & 0.849 & 0.008 & \added{228.8} & MonSter(f.t.) & 0.379 & 0.092 & \added{58.45} \\
    \bottomrule
    \end{tabular}
}
\end{table}

\begin{figure*}[!t]
\centering
\includegraphics[width=0.95\textwidth]{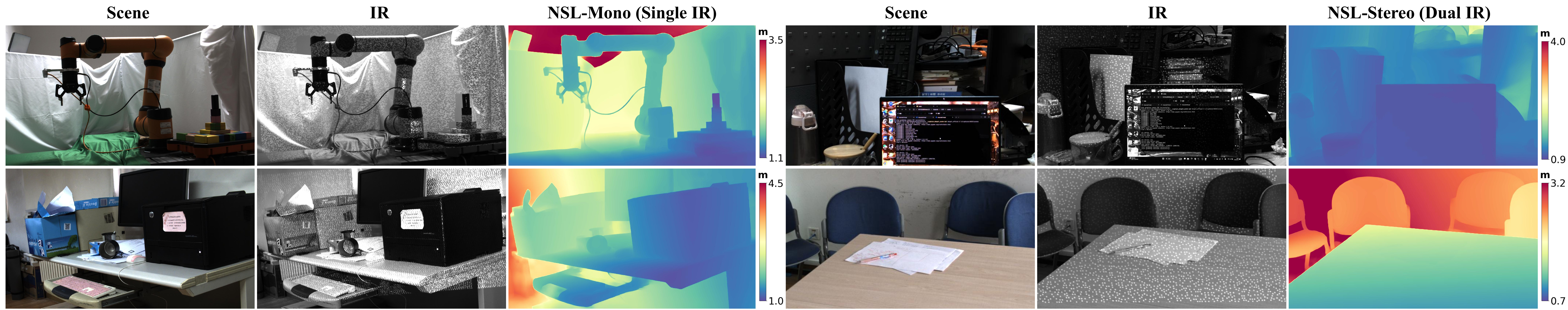}
\caption{\textbf{More qualitative results on real scenes captured with the device held by hand.} No pseudo ground truth available, unlike the fixed setup.
}
\label{fig:real_more_scene}
\end{figure*}

\begin{figure*}[!t]
\centering
\includegraphics[width=0.97\textwidth]{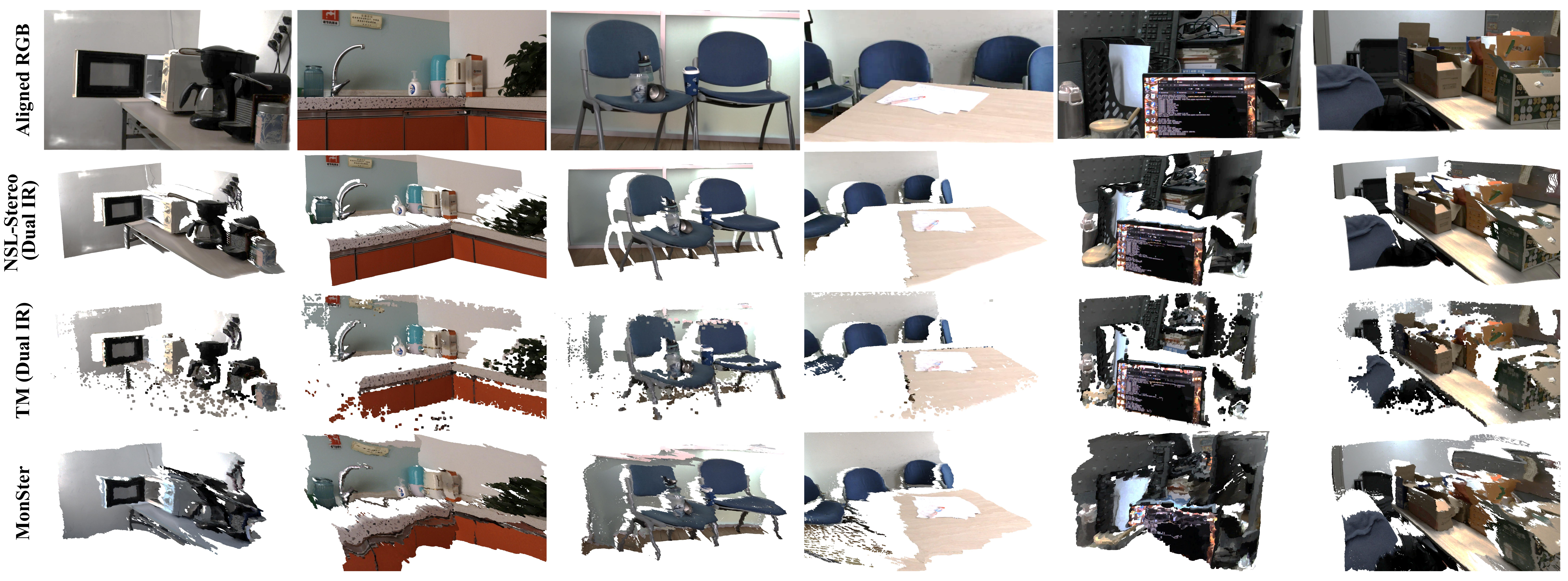}
\caption{\textbf{Point cloud quality comparison of random scene reconstruction under handheld capture}. The same denoising procedure is applied to all point clouds for fair visualization. (TM's point clouds are too noisy to be directly visualized). RGB images are warped onto the IR viewpoint.
}
\label{fig:pointcloud_comparison}
\end{figure*}

\paragraph{Qualitative Real Results.}
We further evaluate all methods under more challenging and diverse real-world scenes using hand-held capture conditions, and present qualitative results by visualizing depth maps across a broader range of scenarios. Representative examples are shown in Figure~\ref{fig:teaser} and Figure~\ref{fig:real_more_scene}. \added{We also show the point clouds comparison in Figure~\ref{fig:pointcloud_comparison}. Since the traditional decoder's point clouds are too noisy to be directly visualized, we apply the same point cloud denoising by removing points with anomalous depth gradient to all point clouds.}
Compared to the traditional decoding approach and state-of-the-art passive RGB stereo methods, our approach produces cleaner and more complete depth maps with fewer artifacts and sharper boundaries, particularly in difficult regions such as textureless surfaces and reflective or occluded areas.

\begin{figure*}[!t]
\centering
\includegraphics[width=0.96\textwidth, height=0.48\textwidth]{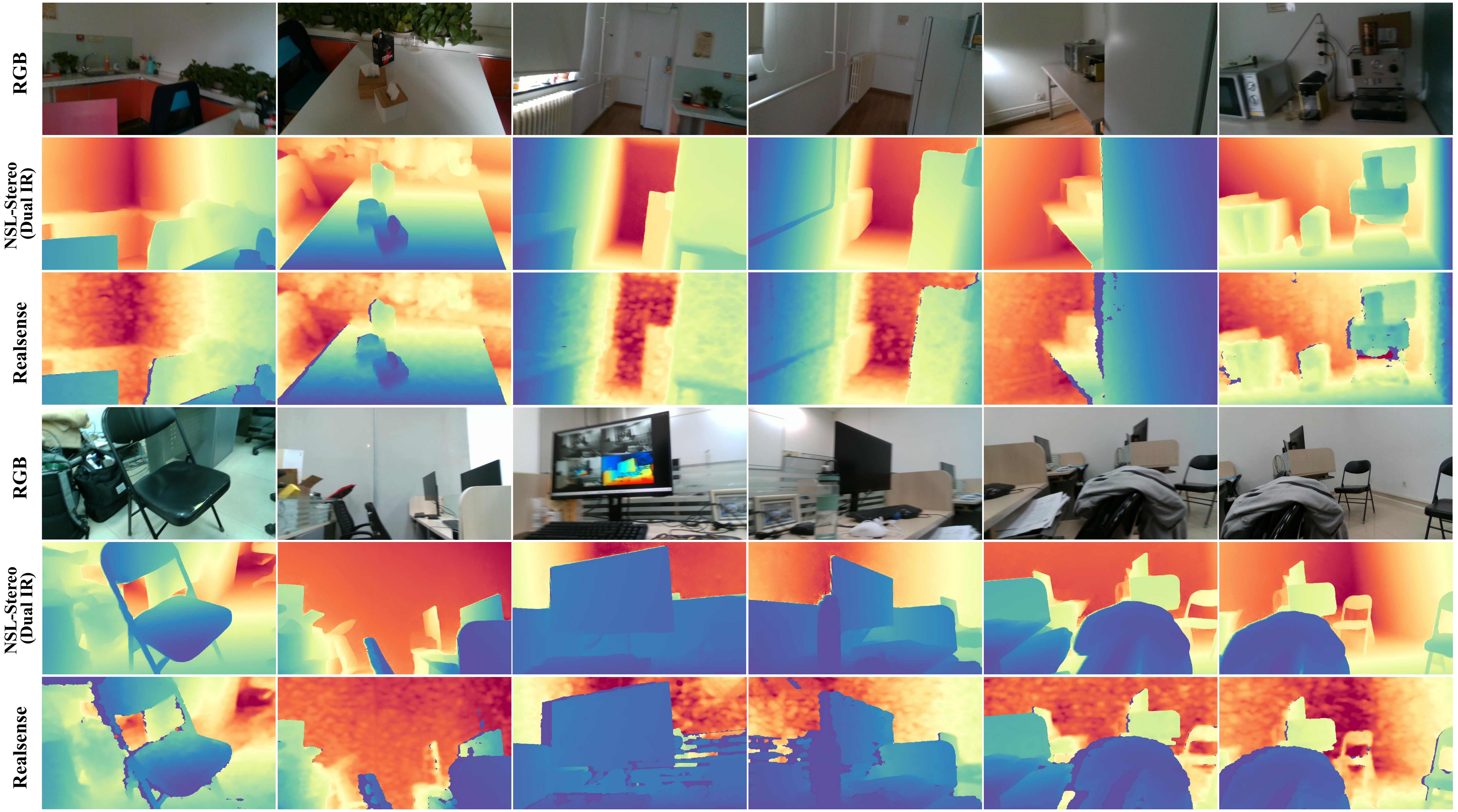}
\caption{\textbf{Qualitative results of {\name} on RealSense dual IR data.} Trained only on synthetic data, {\name} generalizes well to real IR and outperforms RealSense. Due to lower input quality, These depth maps are slightly inferior to those from our self-built device.
}
\label{fig:realsense_dualir}
\end{figure*}

Additionally, we evaluate our method on video sequences randomly recorded using a RealSense D435 depth camera, which demonstrates our method significantly outperforms commercial structured light cameras (Figure~\ref{fig:realsense_dualir})

\added{\paragraph{Limitation.} Though our method generalizes well to high-quality inputs, its performance may degrade on lower-quality real data affected by factors such as motion-blur or hardware imperfections (e.g. lens blur). These are not considered during data synthesizing and result in a remaining sim2real gap.}

\begin{figure*}[!t]
\centering
\includegraphics[width=0.97\textwidth]{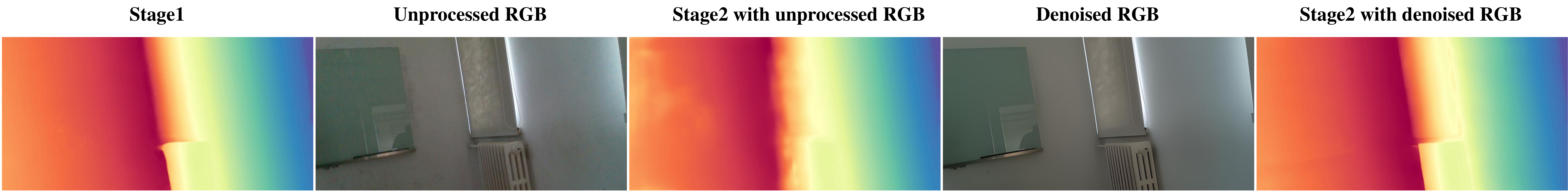}
\caption{The second-stage refinement may produces oversmoothed depth with noisy images from Realsense D435. Applying denoising and deblurring mitigates the issue. For clearer visualization, RGB images are used as inputs to the monocular depth refiner here.
}
\label{fig:realsense_qual_monodepth}
\end{figure*}

{Monocular depth refiner on low-quality images}. While it performs well on real data from our high-end device, its effectiveness declines under low-quality imaging conditions like RealSense. Specifically, due to the absence of motion blur, noise, and defocus in the synthetic training data, the refiner tends to oversmooth depth predictions on RealSense D435 inputs. In contrast, the first-stage neural feature matching remains robust even under such degradations. As shown in Figure~\ref{fig:realsense_qual_monodepth}, applying deblurring and denoising as preprocessing may help mitigate this issue.

\added{{Object boundaries}. Some artifacts can be observed on boundaries at depth discontinuities. Apart from the outliers caused by reprojection from IR view to RGB view in Realsense results, there may be oversmooth boundaries, especially under handheld condition where motion-blur is severe. We will deal with non-ideal capture conditions in the future works.}

\subsection{\added{Runtime Analysis}}
\label{sec:runtime}

\added{Table~\ref{tab:inference_time} reports the inference time on a 640$\times$480 image pair with a single RTX4090 GPU. Though our model is not yet real-time (e.g., 30fps) due to its complexity and lack of speed optimization, it could be accelerated by standard techniques (e.g., quantization and distillation). 
ActiveStereoNet is faster but produces inaccurate depth with over-smooth details. Traditional method (TM) is ultra-fast but lacks robustness, producing noisy results. Passive RGB stereo baseline, MonSter, lags in both speed and quality.}

\begin{table}[htbp] 
\centering
\caption{Runtime.}
\vspace*{-0.3cm}
\label{tab:inference_time} 
\renewcommand{\arraystretch}{1}
    \begin{tabular}{p{3.3cm} c}
    \toprule
    Methods & \textbf{FPS} \\
    \midrule
    \textbf{$\namem$}/\textbf{$\names$} & 18.94 \\ 
    \textbf{$\nameb$} & 15.38 \\
    TM & 1098.89 \\
    ActiveStereoNet & 80.39 \\ 
    MonSter & 3.63 \\
    \bottomrule
    \end{tabular}
\end{table}

\section{Conclusions}

We present \name, a neural decoding framework for single-shot structured light that performs correspondence matching in feature space rather than in the pixel domain. Our method significantly improves robustness and accuracy across both monocular and binocular structured light setups, outperforming traditional decoders and even RGB-based stereo methods. 
In future work, we aim to extend \name\ to handle dynamic scenes, uncalibrated settings, and jointly learn pattern design and decoding in an end-to-end manner.
\begin{acks}
This research is an achievement of the Key Laboratory of Science, Technology and Standard in Press Industry (Key Laboratory of Intelligent Press Media Techonology). We gratefully acknowledge the support from Jiiov Technology for providing computing resources.
\end{acks}

\setlength{\tabcolsep}{1.5pt}

\renewcommand{\arraystretch}{0.02}

\setlength{\imgwidth}{0.138\textwidth}

\renewcommand{\formattedgraphics}[1]{%
    \includegraphics[width=\imgwidth,keepaspectratio]{#1}%
}


\setlength{\colorbarwidth}{0.0182\textwidth} 
\renewcommand{\colorbar}[1]{%
    \includegraphics[width=\colorbarwidth, keepaspectratio]{#1}%
}

\bibliographystyle{ACM-Reference-Format}
\bibliography{references}

@inproceedings{lipson2021raft,
  title={Raft-stereo: Multilevel recurrent field transforms for stereo matching},
  author={Lipson, Lahav and Teed, Zachary and Deng, Jia},
  booktitle={2021 International Conference on 3D Vision (3DV)},
  pages={218--227},
  year={2021},
  organization={IEEE}
}

@inproceedings{lin2024promptda,
  title={Prompting Depth Anything for 4K Resolution Accurate Metric Depth Estimation},
  author={Lin, Haotong and Peng, Sida and Chen, Jingxiao and Peng, Songyou and Sun, Jiaming and Liu, Minghuan and Bao, Hujun and Feng, Jiashi and Zhou, Xiaowei and Kang, Bingyi},
  journal={arXiv},
  year={2024}
}

@inproceedings{procthor,
  author={Matt Deitke and Eli VanderBilt and Alvaro Herrasti and
          Luca Weihs and Jordi Salvador and Kiana Ehsani and
          Winson Han and Eric Kolve and Ali Farhadi and
          Aniruddha Kembhavi and Roozbeh Mottaghi},
  title={{ProcTHOR: Large-Scale Embodied AI Using Procedural Generation}},
  booktitle={NeurIPS},
  year={2022},
  note={Outstanding Paper Award}
}

@inproceedings{tao2015depth,
  title={Depth from shading, defocus, and correspondence using light-field angular coherence},
  author={Tao, Michael W and Srinivasan, Pratul P and Malik, Jitendra and Rusinkiewicz, Szymon and Ramamoorthi, Ravi},
  booktitle={Proceedings of the IEEE Conference on Computer Vision and Pattern Recognition},
  pages={1940--1948},
  year={2015}
}

@inproceedings{godard2017unsupervised,
  title={Unsupervised monocular depth estimation with left-right consistency},
  author={Godard, Cl{\'e}ment and Mac Aodha, Oisin and Brostow, Gabriel J},
  booktitle={Proceedings of the IEEE conference on computer vision and pattern recognition},
  pages={270--279},
  year={2017}
}

@inproceedings{hazirbas2019deep,
  title={Deep depth from focus},
  author={Hazirbas, Caner and Soyer, Sebastian Georg and Staab, Maximilian Christian and Leal-Taix{\'e}, Laura and Cremers, Daniel},
  booktitle={Computer Vision--ACCV 2018: 14th Asian Conference on Computer Vision, Perth, Australia, December 2--6, 2018, Revised Selected Papers, Part III 14},
  pages={525--541},
  year={2019},
  organization={Springer}
}

@inproceedings{kadambi2015polarized,
  title={Polarized 3d: High-quality depth sensing with polarization cues},
  author={Kadambi, Achuta and Taamazyan, Vage and Shi, Boxin and Raskar, Ramesh},
  booktitle={Proceedings of the IEEE international conference on computer vision},
  pages={3370--3378},
  year={2015}
}

@article{laurentini1994visual,
  title={The visual hull concept for silhouette-based image understanding},
  author={Laurentini, Aldo},
  journal={IEEE Transactions on pattern analysis and machine intelligence},
  volume={16},
  number={2},
  pages={150--162},
  year={1994},
  publisher={IEEE}
}

@article{mildenhall2021nerf,
  title={Nerf: Representing scenes as neural radiance fields for view synthesis},
  author={Mildenhall, Ben and Srinivasan, Pratul P and Tancik, Matthew and Barron, Jonathan T and Ramamoorthi, Ravi and Ng, Ren},
  journal={Communications of the ACM},
  volume={65},
  number={1},
  pages={99--106},
  year={2021},
  publisher={ACM New York, NY, USA}
}

@incollection{levoy2023light,
  title={Light field rendering},
  author={Levoy, Marc and Hanrahan, Pat},
  booktitle={Seminal Graphics Papers: Pushing the Boundaries, Volume 2},
  pages={441--452},
  year={2023}
}

@InProceedings{alacarte,
author = {Mirdehghan, Parsa and Chen, Wenzheng and Kutulakos, Kiriakos N.},
title = {Optimal Structured Light à La Carte},
booktitle = {Proceedings of the IEEE Conference on Computer Vision and Pattern Recognition (CVPR)},
month = {June},
year = {2018}
}

@article{geng2011structured,
  title={Structured-light 3D surface imaging: a tutorial},
  author={Geng, Jason},
  journal={Advances in optics and photonics},
  volume={3},
  number={2},
  pages={128--160},
  year={2011},
  publisher={Optical Society of America}
}

@inproceedings{sun2023consistent,
  title={Consistent direct time-of-flight video depth super-resolution},
  author={Sun, Zhanghao and Ye, Wei and Xiong, Jinhui and Choe, Gyeongmin and Wang, Jialiang and Su, Shuochen and Ranjan, Rakesh},
  booktitle={Proceedings of the ieee/cvf conference on computer vision and pattern recognition},
  pages={5075--5085},
  year={2023}
}

@inproceedings{huang2023neurallidar,
  title={Neural lidar fields for novel view synthesis},
  author={Huang, Shengyu and Gojcic, Zan and Wang, Zian and Williams, Francis and Kasten, Yoni and Fidler, Sanja and Schindler, Konrad and Litany, Or},
  booktitle={Proceedings of the IEEE/CVF International Conference on Computer Vision},
  pages={18236--18246},
  year={2023}
}

@misc{AzureKinectDK,
  title = {Azure Kinect DK – Develop AI Models},
  author = {Microsoft},
  year = {2025},
  url = {https://azure.microsoft.com/en-us/products/kinect-dk},
  note = {Accessed: 2025-04-08}
}

@misc{FaceID,
  title = {Apple Face ID},
  author = {Apple},
  year = {2025},
  url = {https://support.apple.com/en-us/102381},
  note = {Accessed: 2025-04-08}
}

@misc{IntelRealSense,
  title = {Intel RealSense – Computer Vision and Depth Tracking Cameras},
  author = {Intel Corporation},
  year = {2025},
  url = {https://www.intelrealsense.com/},
  note = {Accessed: 2025-04-08}
}

@article{will1971grid,
  title={Grid coding: A preprocessing technique for robot and machine vision},
  author={Will, Peter M and Pennington, Keith S},
  journal={Artificial Intelligence},
  volume={2},
  number={3-4},
  pages={319--329},
  year={1971},
  publisher={Elsevier}
}

@article{posdamer1982surface,
  title={Surface measurement by space-encoded projected beam systems},
  author={Posdamer, Jeffrey L and Altschuler, Martin D},
  journal={Computer graphics and image processing},
  volume={18},
  number={1},
  pages={1--17},
  year={1982},
  publisher={Elsevier}
}

@article{salvi2010state,
  title={A state of the art in structured light patterns for surface profilometry},
  author={Salvi, Joaquim and Fernandez, Sergio and Pribanic, Tomislav and Llado, Xavier},
  journal={Pattern recognition},
  volume={43},
  number={8},
  pages={2666--2680},
  year={2010},
  publisher={Elsevier}
}

@inproceedings{scharstein2003high,
  title={High-accuracy stereo depth maps using structured light},
  author={Scharstein, Daniel and Szeliski, Richard},
  booktitle={2003 IEEE Computer Society Conference on Computer Vision and Pattern Recognition, 2003. Proceedings.},
  volume={1},
  pages={I--I},
  year={2003},
  organization={IEEE}
}

@inproceedings{aliaga2008photogeometric,
  title={Photogeometric structured light: A self-calibrating and multi-viewpoint framework for accurate 3d modeling},
  author={Aliaga, Daniel G and Xu, Yi},
  booktitle={2008 IEEE Conference on Computer Vision and Pattern Recognition},
  pages={1--8},
  year={2008},
  organization={IEEE}
}

@inproceedings{kawasaki2008dynamic,
  title={Dynamic scene shape reconstruction using a single structured light pattern},
  author={Kawasaki, Hiroshi and Furukawa, Ryo and Sagawa, Ryusuke and Yagi, Yasushi},
  booktitle={2008 IEEE conference on computer vision and pattern recognition},
  pages={1--8},
  year={2008},
  organization={Ieee}
}

@article{koninckx2006real,
  title={Real-time range acquisition by adaptive structured light},
  author={Koninckx, Thomas P and Van Gool, Luc},
  journal={IEEE transactions on pattern analysis and machine intelligence},
  volume={28},
  number={3},
  pages={432--445},
  year={2006},
  publisher={IEEE}
}

@inproceedings{sagawa2011dense,
  title={Dense one-shot 3D reconstruction by detecting continuous regions with parallel line projection},
  author={Sagawa, Ryusuke and Kawasaki, Hiroshi and Kiyota, Shota and Furukawa, Ryo},
  booktitle={2011 International Conference on Computer Vision},
  pages={1911--1918},
  year={2011},
  organization={IEEE}
}

@inproceedings{taguchi2012motion,
  title={Motion-aware structured light using spatio-temporal decodable patterns},
  author={Taguchi, Yuichi and Agrawal, Amit and Tuzel, Oncel},
  booktitle={Computer Vision--ECCV 2012: 12th European Conference on Computer Vision, Florence, Italy, October 7-13, 2012, Proceedings, Part V 12},
  pages={832--845},
  year={2012},
  organization={Springer}
}

@article{maruyama1993range,
  title={Range sensing by projecting multiple slits with random cuts},
  author={Maruyama, Minoru and Abe, Shigeru},
  journal={IEEE Transactions on Pattern Analysis and Machine Intelligence},
  volume={15},
  number={6},
  pages={647--651},
  year={1993},
  publisher={IEEE}
}

@inproceedings{zhang2002rapid,
  title={Rapid shape acquisition using color structured light and multi-pass dynamic programming},
  author={Zhang, Li and Curless, Brian and Seitz, Steven M},
  booktitle={Proceedings. First International Symposium on 3D Data Processing Visualization and Transmission},
  pages={24--36},
  year={2002},
  organization={IEEE}
}

@article{le1988structured,
  title={Structured light patterns for robot mobility},
  author={Le Moigne, JJ and Waxman, Allen Mark},
  journal={IEEE Journal on Robotics and Automation},
  volume={4},
  number={5},
  pages={541--548},
  year={1988},
  publisher={IEEE}
}

@article{vuylsteke1990range,
  title={Range image acquisition with a single binary-encoded light pattern},
  author={Vuylsteke, Piet and Oosterlinck, Andr{\'e}},
  journal={IEEE transactions on pattern analysis and machine intelligence},
  volume={12},
  number={2},
  pages={148--164},
  year={1990},
  publisher={IEEE}
}

@inproceedings{martinez2013kinect,
  title={Kinect Unleashed: Getting Control over High Resolution Depth Maps.},
  author={Martinez, Manuel and Stiefelhagen, Rainer},
  booktitle={MVA},
  pages={247--250},
  year={2013}
}

@inproceedings{chang2018pyramid,
  title={Pyramid stereo matching network},
  author={Chang, Jia-Ren and Chen, Yong-Sheng},
  booktitle={Proceedings of the IEEE conference on computer vision and pattern recognition},
  pages={5410--5418},
  year={2018}
}

@inproceedings{guo2019group,
  title={Group-wise correlation stereo network},
  author={Guo, Xiaoyang and Yang, Kai and Yang, Wukui and Wang, Xiaogang and Li, Hongsheng},
  booktitle={Proceedings of the IEEE/CVF conference on computer vision and pattern recognition},
  pages={3273--3282},
  year={2019}
}

@inproceedings{xu2020aanet,
  title={Aanet: Adaptive aggregation network for efficient stereo matching},
  author={Xu, Haofei and Zhang, Juyong},
  booktitle={Proceedings of the IEEE/CVF conference on computer vision and pattern recognition},
  pages={1959--1968},
  year={2020}
}

@inproceedings{zhang2019ga,
  title={Ga-net: Guided aggregation net for end-to-end stereo matching},
  author={Zhang, Feihu and Prisacariu, Victor and Yang, Ruigang and Torr, Philip HS},
  booktitle={Proceedings of the IEEE/CVF conference on computer vision and pattern recognition},
  pages={185--194},
  year={2019}
}

@inproceedings{chen2024mocha,
  title={Mocha-stereo: Motif channel attention network for stereo matching},
  author={Chen, Ziyang and Long, Wei and Yao, He and Zhang, Yongjun and Wang, Bingshu and Qin, Yongbin and Wu, Jia},
  booktitle={Proceedings of the IEEE/CVF Conference on Computer Vision and Pattern Recognition},
  pages={27768--27777},
  year={2024}
}

@inproceedings{jing2023uncertainty,
  title={Uncertainty guided adaptive warping for robust and efficient stereo matching},
  author={Jing, Junpeng and Li, Jiankun and Xiong, Pengfei and Liu, Jiangyu and Liu, Shuaicheng and Guo, Yichen and Deng, Xin and Xu, Mai and Jiang, Lai and Sigal, Leonid},
  booktitle={Proceedings of the IEEE/CVF International Conference on Computer Vision},
  pages={3318--3327},
  year={2023}
}

@inproceedings{li2024local,
  title={LoS: Local structure-guided stereo matching},
  author={Li, Kunhong and Wang, Longguang and Zhang, Ye and Xue, Kaiwen and Zhou, Shunbo and Guo, Yulan},
  booktitle={Proceedings of the IEEE/CVF Conference on Computer Vision and Pattern Recognition},
  pages={19746--19756},
  year={2024}
}

@inproceedings{xu2023iterative,
  title={Iterative geometry encoding volume for stereo matching},
  author={Xu, Gangwei and Wang, Xianqi and Ding, Xiaohuan and Yang, Xin},
  booktitle={Proceedings of the IEEE/CVF conference on computer vision and pattern recognition},
  pages={21919--21928},
  year={2023}
}

@article{tolgyessy2021evaluation,
  title={Evaluation of the azure kinect and its comparison to kinect v1 and kinect v2},
  author={T{\"o}lgyessy, Michal and Dekan, Martin and Chovanec, L'ubo{\v{s}} and Hubinsk{\`y}, Peter},
  journal={Sensors},
  volume={21},
  number={2},
  pages={413},
  year={2021},
  publisher={MDPI}
}

@article{yang2024dav2,
  title={Depth anything v2},
  author={Yang, Lihe and Kang, Bingyi and Huang, Zilong and Zhao, Zhen and Xu, Xiaogang and Feng, Jiashi and Zhao, Hengshuang},
  journal={Advances in Neural Information Processing Systems},
  volume={37},
  pages={21875--21911},
  year={2024}
}

@article{oquabdinov2,
  author       = {Maxime Oquab and
                  Timoth{\'{e}}e Darcet and
                  Th{\'{e}}o Moutakanni and
                  Huy V. Vo and
                  Marc Szafraniec and
                  Vasil Khalidov and
                  Pierre Fernandez and
                  Daniel Haziza and
                  Francisco Massa and
                  Alaaeldin El{-}Nouby and
                  Mido Assran and
                  Nicolas Ballas and
                  Wojciech Galuba and
                  Russell Howes and
                  Po{-}Yao Huang and
                  Shang{-}Wen Li and
                  Ishan Misra and
                  Michael Rabbat and
                  Vasu Sharma and
                  Gabriel Synnaeve and
                  Hu Xu and
                  Herv{\'{e}} J{\'{e}}gou and
                  Julien Mairal and
                  Patrick Labatut and
                  Armand Joulin and
                  Piotr Bojanowski},
  title        = {DINOv2: Learning Robust Visual Features without Supervision},
  journal      = {Trans. Mach. Learn. Res.},
  volume       = {2024},
  year         = {2024},
  url          = {https://openreview.net/forum?id=a68SUt6zFt},
  bibsource    = {dblp computer science bibliography, https://dblp.org}
}

@inproceedings{ranftl2021vision,
  title={Vision transformers for dense prediction},
  author={Ranftl, Ren{\'e} and Bochkovskiy, Alexey and Koltun, Vladlen},
  booktitle={Proceedings of the IEEE/CVF international conference on computer vision},
  pages={12179--12188},
  year={2021}
}

@misc{libSGM,
  author       = {Fixstars},
  title        = {libSGM},
  year         = {2025},
  url          = {https://github.com/fixstars/libSGM},
  note         = {Accessed: 2025-04-10}
}

@misc{opencv,
  author       = {OpenCV Contributors},
  title        = {OpenCV},
  year         = {2025},
  url          = {https://opencv.org/},
  note         = {Accessed: 2025-04-10}
}

@inproceedings{zhang2018activestereonet,
  title={Activestereonet: End-to-end self-supervised learning for active stereo systems},
  author={Zhang, Yinda and Khamis, Sameh and Rhemann, Christoph and Valentin, Julien and Kowdle, Adarsh and Tankovich, Vladimir and Schoenberg, Michael and Izadi, Shahram and Funkhouser, Thomas and Fanello, Sean},
  booktitle={Proceedings of the european conference on computer vision (ECCV)},
  pages={784--801},
  year={2018}
}

@inproceedings{baek2021polka,
  title={Polka lines: Learning structured illumination and reconstruction for active stereo},
  author={Baek, Seung-Hwan and Heide, Felix},
  booktitle={Proceedings of the IEEE/CVF conference on computer vision and pattern recognition},
  pages={5757--5767},
  year={2021}
}

@inproceedings{riegler2019connecting,
  title={Connecting the dots: Learning representations for active monocular depth estimation},
  author={Riegler, Gernot and Liao, Yiyi and Donne, Simon and Koltun, Vladlen and Geiger, Andreas},
  booktitle={Proceedings of the IEEE/CVF conference on computer vision and pattern recognition},
  pages={7624--7633},
  year={2019}
}

@inproceedings{xu2022monobino,
  title={Depth estimation by combining binocular stereo and monocular structured-light},
  author={Xu, Yuhua and Yang, Xiaoli and Yu, Yushan and Jia, Wei and Chu, Zhaobi and Guo, Yulan},
  booktitle={Proceedings of the IEEE/CVF Conference on Computer Vision and Pattern Recognition},
  pages={1746--1755},
  year={2022}
}

@article{cheng2025monster,
  title={MonSter: Marry Monodepth to Stereo Unleashes Power},
  author={Cheng, Junda and Liu, Longliang and Xu, Gangwei and Wang, Xianqi and Zhang, Zhaoxing and Deng, Yong and Zang, Jinliang and Chen, Yurui and Cai, Zhipeng and Yang, Xin},
  journal={arXiv preprint arXiv:2501.08643},
  year={2025}
}

@inproceedings{fanello2016hyperdepth,
  title={Hyperdepth: Learning depth from structured light without matching},
  author={Fanello, Sean Ryan and Rhemann, Christoph and Tankovich, Vladimir and Kowdle, Adarsh and Escolano, Sergio Orts and Kim, David and Izadi, Shahram},
  booktitle={Proceedings of the IEEE conference on computer vision and pattern recognition},
  pages={5441--5450},
  year={2016}
}

@inproceedings{fanello2017ultrastereo,
  title={Ultrastereo: Efficient learning-based matching for active stereo systems},
  author={Fanello, Sean Ryan and Valentin, Julien and Rhemann, Christoph and Kowdle, Adarsh and Tankovich, Vladimir and Davidson, Philip and Izadi, Shahram},
  booktitle={2017 IEEE Conference on Computer Vision and Pattern Recognition (CVPR)},
  pages={6535--6544},
  year={2017},
  organization={IEEE}
}

@inproceedings{zhan2018unsupervised,
  author       = {Huangying Zhan and
                  Ravi Garg and
                  Chamara Saroj Weerasekera and
                  Kejie Li and
                  Harsh Agarwal and
                  Ian D. Reid},
  title        = {Unsupervised Learning of Monocular Depth Estimation and Visual Odometry
                  With Deep Feature Reconstruction},
  booktitle    = {2018 {IEEE} Conference on Computer Vision and Pattern Recognition,
                  {CVPR}},
  pages        = {340--349},
  publisher    = {Computer Vision Foundation / {IEEE} Computer Society},
  year         = {2018},
}

@techreport{shapenet2015,
  title       = {{ShapeNet: An Information-Rich 3D Model Repository}},
  author      = {Chang, Angel X. and Funkhouser, Thomas and Guibas, Leonidas and Hanrahan, Pat and Huang, Qixing and Li, Zimo and Savarese, Silvio and Savva, Manolis and Song, Shuran and Su, Hao and Xiao, Jianxiong and Yi, Li and Yu, Fisher},
  number      = {arXiv:1512.03012 [cs.GR]},
  institution = {Stanford University --- Princeton University --- Toyota Technological Institute at Chicago},
  year        = {2015}
}

@inproceedings{dosovitskiy2021vit,
  author       = {Alexey Dosovitskiy and
                  Lucas Beyer and
                  Alexander Kolesnikov and
                  Dirk Weissenborn and
                  Xiaohua Zhai and
                  Thomas Unterthiner and
                  Mostafa Dehghani and
                  Matthias Minderer and
                  Georg Heigold and
                  Sylvain Gelly and
                  Jakob Uszkoreit and
                  Neil Houlsby},
  title        = {An Image is Worth 16x16 Words: Transformers for Image Recognition
                  at Scale},
  booktitle    = {9th International Conference on Learning Representations, {ICLR} 2021,
                  Virtual Event, Austria, May 3-7, 2021},
  publisher    = {OpenReview.net},
  year         = {2021},
  url          = {https://openreview.net/forum?id=YicbFdNTTy},
  bibsource    = {dblp computer science bibliography, https://dblp.org}
}
\newpage
\appendix
\includepdf[pages=-]{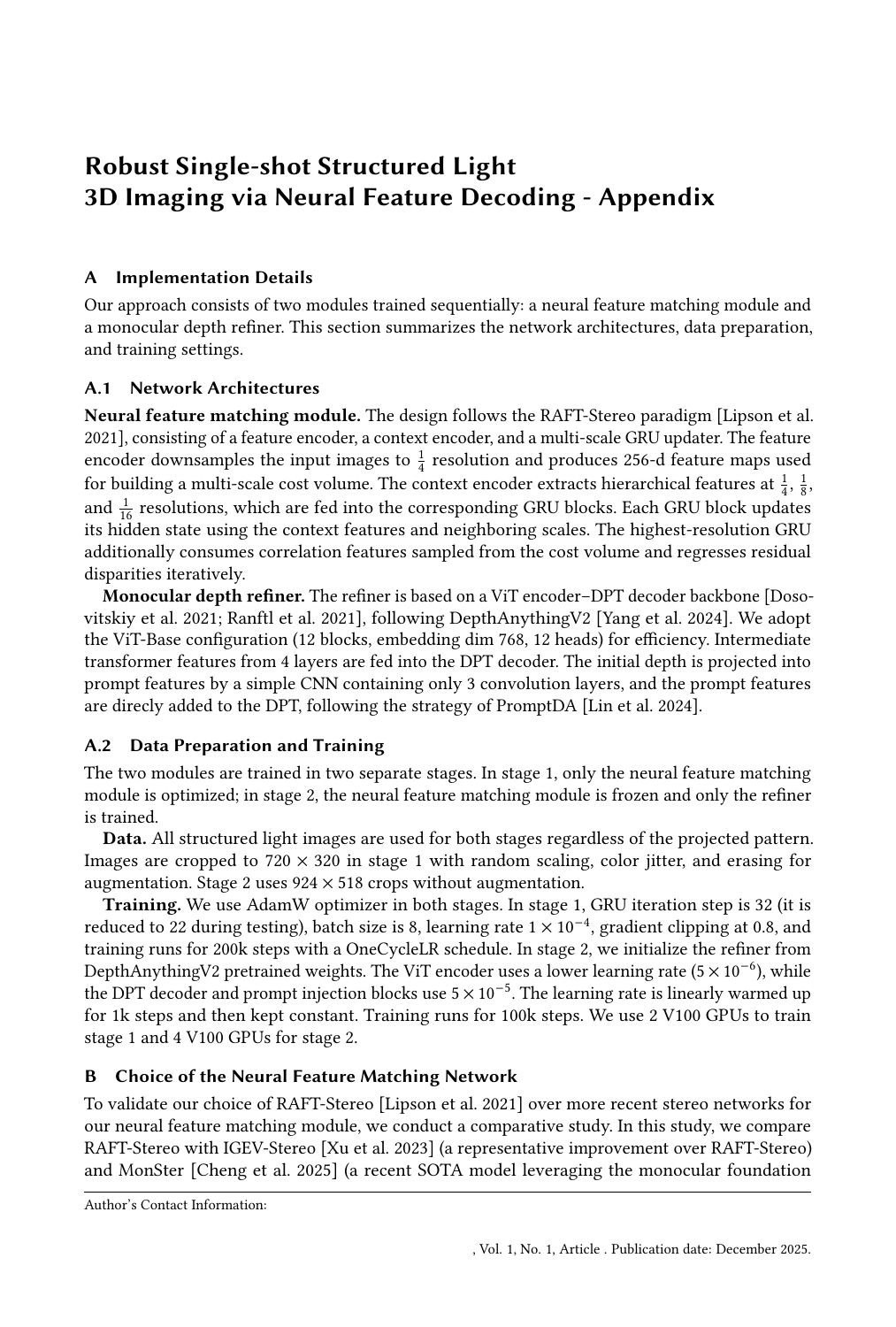}

\end{document}


\title[Robust Single-shot Structured Light 3D Imaging  
via Neural Feature Decoding]{Robust Single-shot Structured Light \\ 3D Imaging via Neural Feature Decoding - Appendix}

\renewcommand{\shortauthors}{Li et al.}

\maketitle

\appendix

\section{\added{Implementation Details}}
\label{sec:supp:impl}

\added{Our approach consists of two modules trained sequentially: a neural feature matching module and a monocular depth refiner. This section summarizes the network architectures, data preparation, and training settings.}

\subsection{\added{Network Architectures}}

\added{\textbf{Neural feature matching module.}
The design follows the RAFT-Stereo paradigm~\cite{lipson2021raft}, consisting of a feature encoder, a context encoder, and a multi-scale GRU updater. 
The feature encoder downsamples the input images to $\tfrac{1}{4}$ resolution and produces 256-d feature maps used for building a multi-scale cost volume.
The context encoder extracts hierarchical features at $\tfrac{1}{4}$, $\tfrac{1}{8}$, and $\tfrac{1}{16}$ resolutions, which are fed into the corresponding GRU blocks.
Each GRU block updates its hidden state using the context features and neighboring scales. 
The highest-resolution GRU additionally consumes correlation features sampled from the cost volume and regresses residual disparities iteratively.}

\added{\textbf{Monocular depth refiner.}
The refiner is based on a ViT encoder--DPT decoder backbone~\cite{dosovitskiy2021vit,ranftl2021vision}, following DepthAnythingV2~\cite{yang2024dav2}.
We adopt the ViT-Base configuration (12 blocks, embedding dim 768, 12 heads) for efficiency. 
Intermediate transformer features from 4 layers are fed into the DPT decoder.  
The initial depth is projected into prompt features by a simple CNN containing only 3 convolution layers, and the prompt features are direcly added to the DPT, following the strategy of PromptDA~\cite{lin2024promptda}. }

\subsection{\added{Data Preparation and Training}}

\added{The two modules are trained in two separate stages. 
In stage 1, only the neural feature matching module is optimized; in stage 2, the neural feature matching module is frozen and only the refiner is trained.}

\added{\textbf{Data.}
All structured light images are used for both stages regardless of the projected pattern.
Images are cropped to $720\times320$ in stage 1 with random scaling, color jitter, and erasing for augmentation.
Stage 2 uses $924\times518$ crops without augmentation.}

\added{\textbf{Training.}
We use AdamW optimizer in both stages.
In stage 1, GRU iteration step is 32 (it is reduced to 22 during testing), batch size is 8, learning rate $1\times10^{-4}$, gradient clipping at 0.8, and training runs for 200k steps with a OneCycleLR schedule.  
In stage 2, we initialize the refiner from DepthAnythingV2 pretrained weights. The ViT encoder uses a lower learning rate ($5\times10^{-6}$), while the DPT decoder and prompt injection blocks use $5\times10^{-5}$. The learning rate is linearly warmed up for 1k steps and then kept constant. Training runs for 100k steps. We use 2 V100 GPUs to train stage 1 and 4 V100 GPUs for stage 2.}

\section{Choice of the Neural Feature Matching Network}
\label{sec:supp:network_choice}

To validate our choice of RAFT-Stereo~\cite{lipson2021raft} over more recent stereo networks for our neural feature matching module, we conduct a comparative study. In this study, we compare RAFT-Stereo with IGEV-Stereo~\cite{xu2023iterative} (a representative improvement over RAFT-Stereo) and MonSter~\cite{cheng2025monster} (a recent SOTA model leveraging the monocular foundation model~\cite{yang2024dav2}) on our dataset. The evaluation is performed under two input settings: standard dual RGB images and monocular IR structured light images, without the subsequent depth refinement module to isolate the matching network's performance.

The results (see Table~\ref{tab:baselinecomp}) show that while IGEV-Stereo and MonSter performed better in the standard dual RGB setting, which demonstrates their advancements on such data, their performance difference compared to RAFT-Stereo was not significant in the structured light setting (MAE difference within 2mm). We hypothesize this is because structured light images primarily provide textural information, lacking the high-level semantic and global scene cues that recent SOTA methods leverage from natural images. Consequently, the architectural advancements in these networks, optimized for such cues, offer limited improvement on structured light data.

In summary, given its performance on structured light data is comparable to more complex recent networks, we select the simpler RAFT-Stereo as a suitable and efficient choice for our neural feature matching module, aligning with our goal of effectively leveraging structured light priors.


\begin{table}[htbp]
\caption{Quantitative comparison of different stereo networks on our dataset.}
\centering
    \renewcommand{\arraystretch}{1.2}
    \setlength{\tabcolsep}{2.5pt}
    \begin{tabular}{p{3cm}|ccc|ccc}
    \toprule[0.9pt]
    \textbf{Methods} & \textbf{MAE}(m) ↓ & \textbf{RMSE} ↓ & \textbf{REL} ↓ & $\boldsymbol{\delta_{1.25}}$ ↑ & $\boldsymbol{\delta_{1.10}}$ ↑ & $\boldsymbol{\delta_{1.05}}$ ↑ \\
    \hline
    \multicolumn{7}{c}{\textbf{Dual RGB Input}} \\
    \hline
    RAFT-Stereo  & 0.0969 & 10.6257 & 0.0897 & 0.9362 & 0.8924 & 0.8454\\
    IGEV-Stereo  & 0.0915 & 0.1460 & 0.1730 & 0.9020 & 0.8525 & 0.8020 \\
    MonSter      & 0.0862 & 0.1457 & 0.1839 & 0.9059 & 0.8611 & 0.8087 \\
    \hline
    \multicolumn{7}{c}{\textbf{Single IR + Pattern Input}} \\
    \hline
    RAFT-Stereo & 0.0599 & 0.3340 & 0.0637 & 0.9612	& 0.9387 & 0.9159 \\
    IGEV-Stereo & 0.0587 & 0.1288 & 0.0817 & 0.9489 & 0.9215 & 0.8951 \\
    MonSter     & 0.0583 & 0.1339 & 0.0737 & 0.9508 & 0.9215 & 0.8897 \\
    
    \bottomrule[0.9pt]
    \end{tabular}
\label{tab:baselinecomp}
\end{table}

\added{\textbf{Additional Experiment: Role of Stage 1 Neural Feature Matching}. To further evaluate the contribution of our Stage 1 neural feature matching module, we perform an additional experiment using only Stage 2. In this setting, Stage 1 is entirely removed (and thus no raw depth is provided as prompts to Stage 2), and we directly fine-tune the monocular backbone DepthAnythingV2~\cite{yang2024dav2} on our structured light dataset.}

\added{This configuration almost completely fails due to scale ambiguity, yielding a mean absolute depth error (MAE) of 34 cm on synthetic data. These results highlight that neural feature matching in Stage 1 is indispensable: it supplies the essential depth cues for our neural structured light system, while Stage 2 primarily acts to refine and optimize the predictions.}
\section{Details of Our Dataset}
\label{sec:supp_dataset}

\begin{figure*}[htbp]
\centering

  \begin{minipage}[c][6cm][c]{1.0\textwidth}
  \centering
  \includegraphics[width=\textwidth]{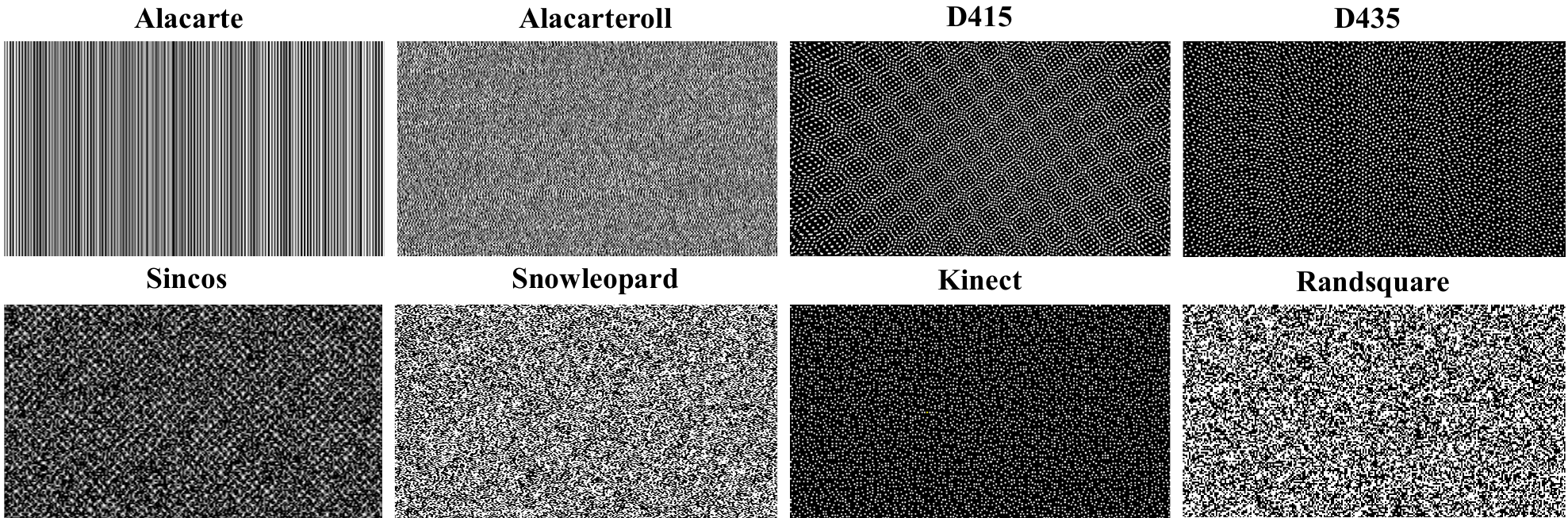}
  \end{minipage}

\caption{\textbf{Patterns used in our dataset.} 6 patterns (Alacarte, Alacarteroll, D415, D415, Sincos, Snowleopard) are used for model training, while 2 (Kinect, Randsquare) are used only during testing.}
\label{fig:pattern}
\end{figure*}

Our dataset is carefully constructed for structured light decoding.
It comprises photorealistic indoor scenes—sourced from the Procther dataset~\cite{procthor} and lit with HDR environment maps to capture authentic lighting variations including brightness differences, color casts, and shadows—populated with thousands of 3D objects from  ShapeNet~\cite{shapenet2015}, featuring a variety of materials including diffuse, specular, and transparent surfaces. 
In addition, the dataset incorporates 8 distinct structured light patterns, with 6 patterns (Alacarte, Alacarte-roll~\footnote{Alacarte is 1D column pattern, where we randomly shift each row to create a 2D pattern}, D415, D415, Sincos
, Snowleopard) allocated for model training and the remaining (Kinect, Randsquare) reserved for testing to evaluate generalization, as shown in Fig.~\ref{fig:pattern}, while the randomization of object placement, pattern selection, and lighting conditions ensures variability that mimics challenges like occlusions, non-Lambertian surfaces, and varying illumination. 
Overall, with approximately 953,000 samples—each sample comprising the left and right RGB images, the pseudo-IR images, a corresponding structured light pattern image, and a ground-truth depth map—this randomized strategy offers a diverse dataset that mirrors real-world structured light scanning scenarios. Data examples are shown in Fig.~\ref{fig:dataset-part1}.

\section{Pointclouds}
\begin{figure*}
    \centering
    \includegraphics[width=1.0\linewidth]{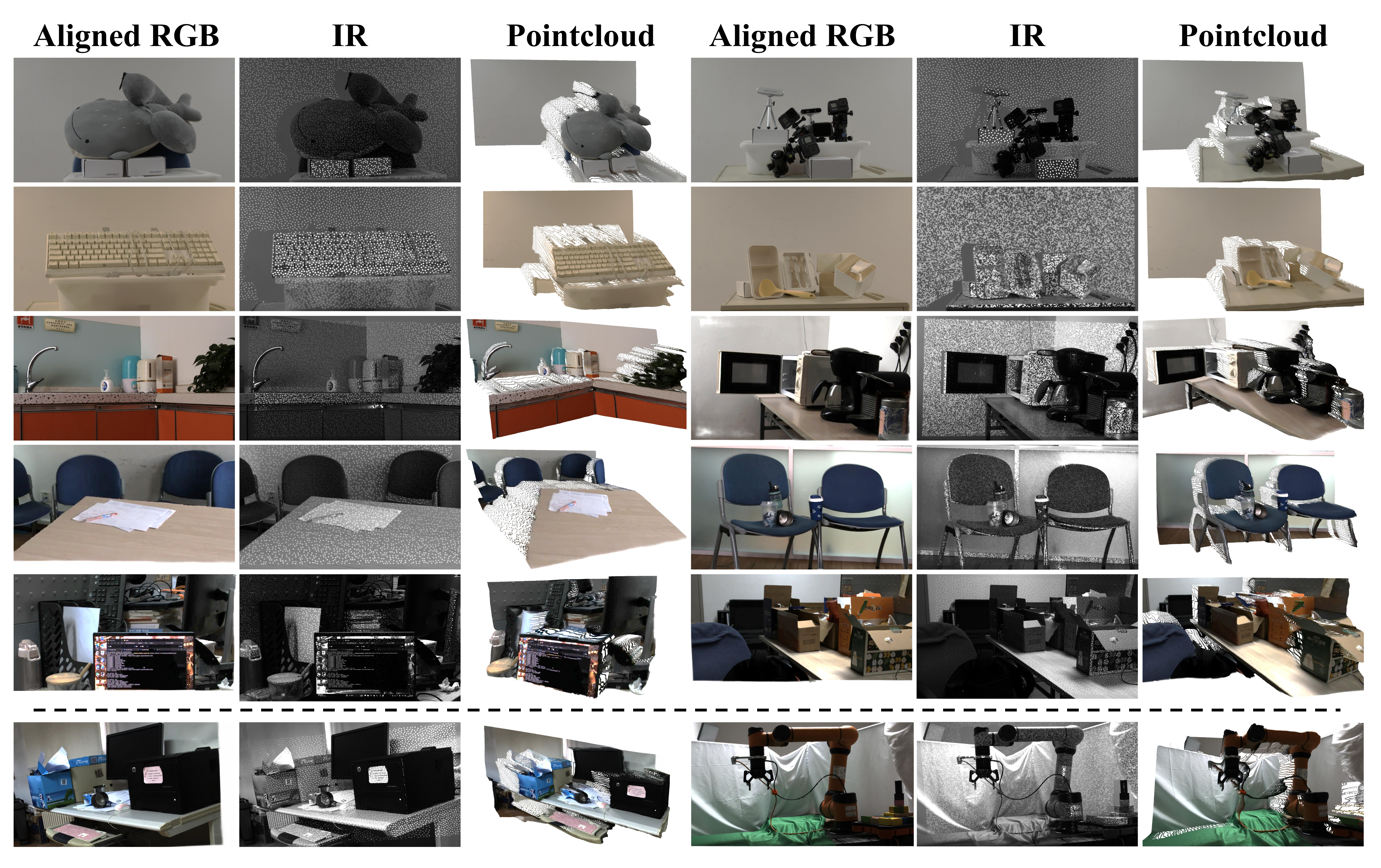}
    \caption{Pointcloud visualizations \textit{without denoising} of our depth results. The first 2 rows are object-level data, while the last 4 rows are random scene-level data captured in handheld condition. The RGB images are warped to align to the IR viewpoint and their brightness are manually increased and contrast reduced for better visualization. The last row shows monocular results obtained by only 1 camera and 1 projector.}
    \label{fig:pointclouds}
\end{figure*}

\added{
To further demonstrate the quality of our model's depth outputs, we present pointcloud visualizations \textit{without any denoising} in Figure~\ref{fig:pointclouds}. The bottom four rows show results of scene-level data captured in handheld condition. Due to handheld acquisition, no RGB images aligns perfectly with the grayscale IR viewpoint, We pre-align and warp RGB image to the IR viewpoint, which may result in incomplete coverage at image boundaries. Additionally, RGB brightness is manually increased and contrast reduced for better visualization, while IR images are unchanged. We also show monocular structured light results at the last row.   
}

\added{
The pointclouds reveal that our method achieves generally high-quality depth estimation even in high-challenging scenes. However, it must be acknowledged the limitations that edge artifacts and over-smooth at depth discontinuities can be observed, which has been discussed in the main text.
}

\section{Generalization Across Patterns and More Results on Real Data}
\label{sec:supp:pattern_generalization}

Table~\ref{tab:pattern_results} reports the evaluation results of all 8 patterns in our dataset under the single IR setting, further demonstrating generalization across pattern types. More details about the patterns are provided in Figure~\ref{fig:pattern}.

\begin{figure*}[t]
\centering
\includegraphics[width=1.0\textwidth]{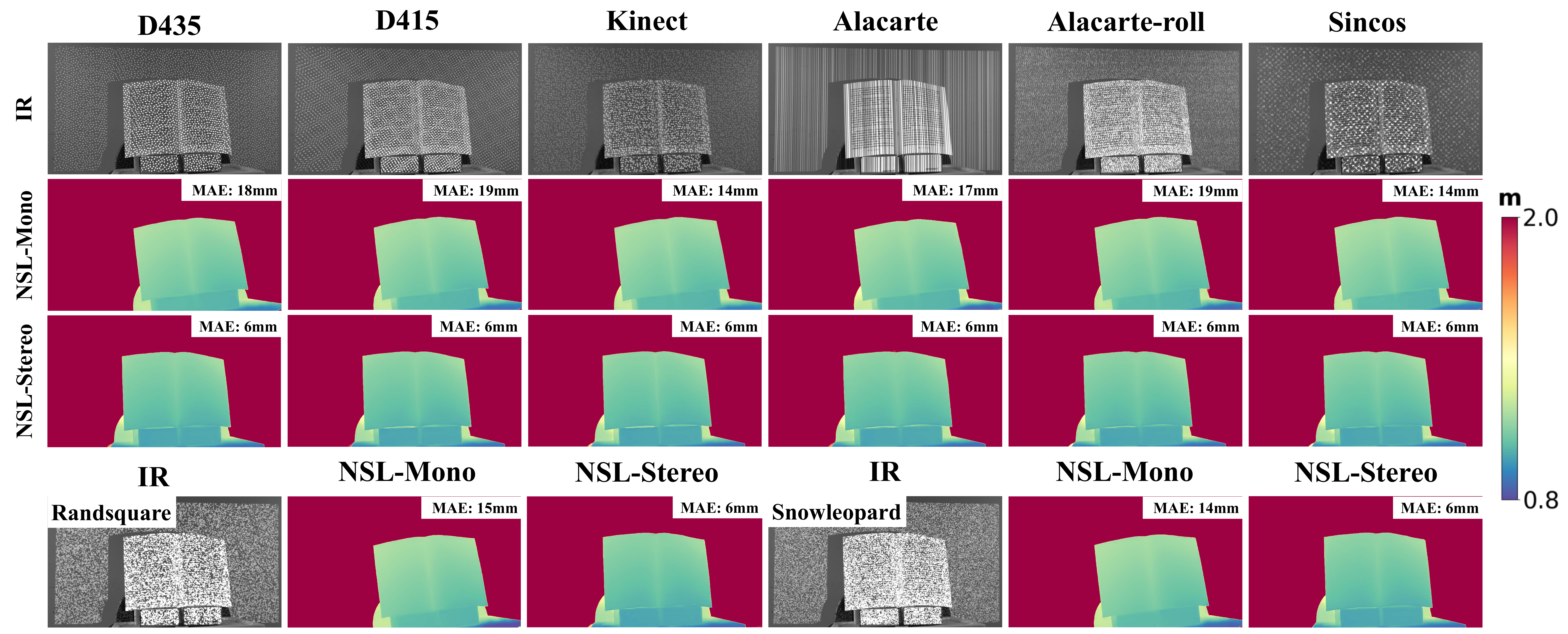}
\caption{\textbf{Depth predictions for the "book" scene captured using all 8 patterns in our dataset.} The results across different patterns are highly consistent. \namem exhibits minor variations comparable to those observed on synthetic data, while \names remains stable. This demonstrates strong pattern generalization of our method on real data. Note that the viewpoint difference between $\namem$ and $\names$ results stems from distinct epipolar rectifications: one for the camera-projector pair (monocular SL) and the other for the stereo IR camera setup (binocular SL).
}
\label{fig:book_8pat}
\end{figure*}

For all 11 scenes with pseudo ground-truth depth, data were acquired using each of the 8 pattern variations. Figure~\ref{fig:book_8pat} shows results from our method on the "\textbf{book}" scene across these patterns. Under the single IR setting (\namem), the variation across patterns is comparable to that observed on the synthetic dataset. In the dual IR setting (\names), the results are highly consistent, highlighting the method’s robustness and generalization to diverse patterns in real-world conditions.

\begin{figure*}[t]
\centering
\includegraphics[width=1.0\textwidth]{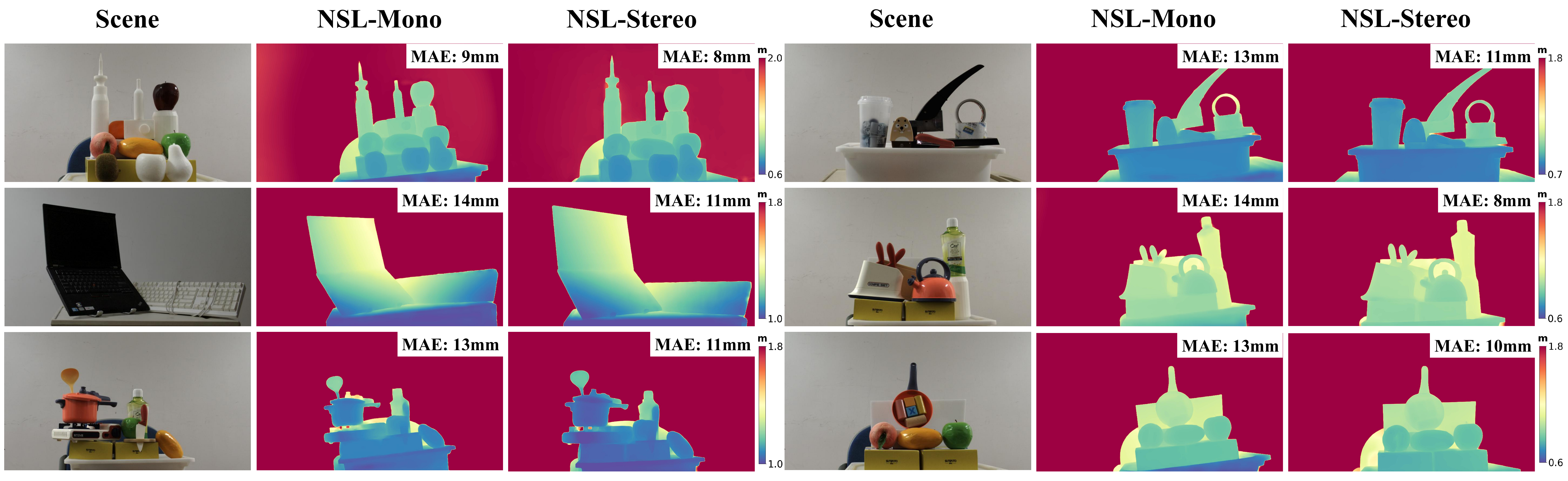}
\caption{\textbf{Results on six additional scenes with pseudo ground-truth depth not shown earlier.} Four scenes appear in the main paper, and one is shown in Figure~\ref{fig:book_8pat} of this appendix. 
}
\label{fig:full_real_with_gt}
\end{figure*}

In addition, Figure~\ref{fig:full_real_with_gt} presents results for the remaining scenes not previously shown, completing the qualitative evaluation across all 11 pseudo ground-truth scenes.

\begin{table}[htbp] 
\caption{Quantitative results for different structured light patterns.} 
\centering
    \renewcommand{\arraystretch}{1.2}
    \setlength{\tabcolsep}{2pt}
    \begin{tabular}{p{2.0cm} | ccc | ccc}
    \toprule[0.9pt] 
    Pattern & MAE(m)↓ & RMSE↓ & REL↓ & $\delta_{1.25}$↑ & $\delta_{1.10}$↑ & $\delta_{1.05}$↑ \\ 
    \hline 
    \multicolumn{7}{c}{Training Patterns } \\
    \hline
    D415 & 0.0388 & 0.0877 & 0.0354 & 0.9716 & 0.9458 & 0.9180 \\
    D435 & 0.0389 & 0.0878 & 0.0361 & 0.9721 & 0.9474 & 0.9195 \\
    Alacarte & 0.0415 & 0.0913 & 0.0356 & 0.9661 & 0.9370 & 0.9068 \\
    Alacarteroll & 0.0437 & 0.0957 & 0.0369 & 0.9685 & 0.9391 & 0.9072 \\
    Snowleopard &  0.0386 & 0.0871 & 0.0356 & 0.9723 & 0.9483 & 0.9207 \\
    Sincos & 0.0428 & 0.0935 & 0.0353 & 0.9712 & 0.9435 & 0.9119 \\
    \hline
    \noalign{\vskip 2.8pt}
    \multicolumn{7}{c}{Testing Patterns } \\
    \hline
    Kinect & 0.0422 & 0.0939 & 0.0370 & 0.9704  & 0.9439 & 0.9139 \\
    Randsquare & 0.0395 & 0.0883 & 0.0338 & 0.9709 & 0.9450 & 0.9157 \\
    \bottomrule[0.9pt] 
    \end{tabular}
\label{tab:pattern_results} 
\end{table}

\section{Monocular structured light using Realsense}
\added{We further evaluate \name's monocular structured light performance on the Realsense D435 by disabling its right IR camera. Unlike a digital projector, the D435 employs a laser source with diffractive optical elements, so the projector pattern in our dataset cannot be directly reused. To obtain the exact pattern of the tested device, we project it onto a white wall at a known distance and capture the wall using the left IR camera. This captured image serves as the input pattern for \name. During testing, scene depths are constrained to be smaller than the wall distance, consistent with the negative disparity assumption in training.}

\begin{figure*}[!t]
\centering
\includegraphics[width=0.95\textwidth]{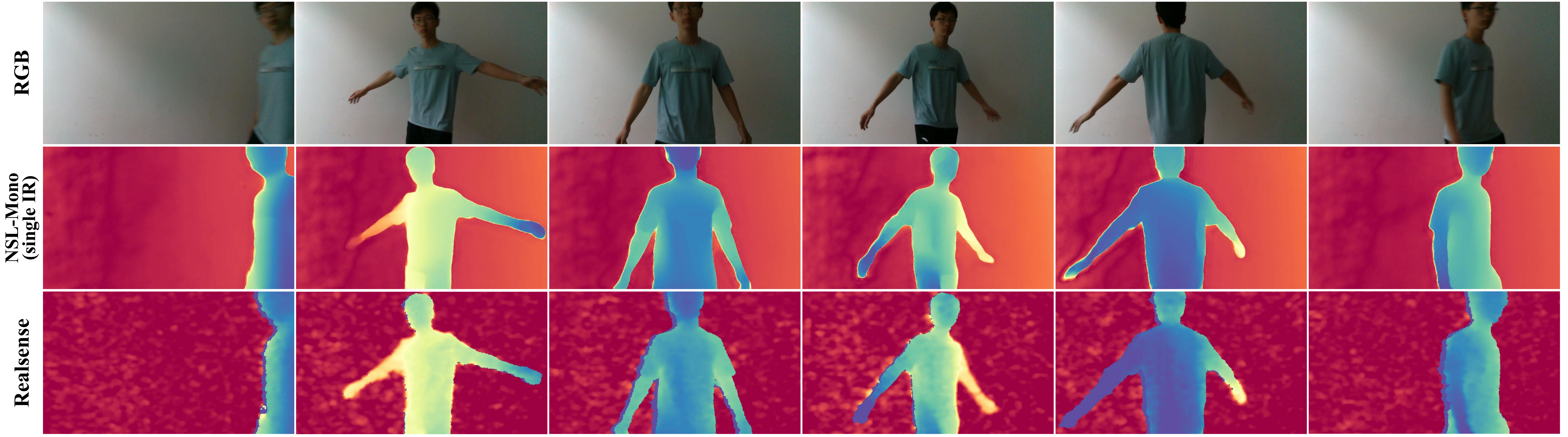}
\caption{\textbf{Qualitative results of {\name} on RealSense data using single IR + pattern.} Depth maps are aligned to RGB for visualization. A reference pattern captured on a planar surface at a known distance enables monocular inference. Despite using only one IR frame, {\name} outperforms RealSense. Note that reference capture inaccuracies may cause depth deviations not attributable to the method itself.
}
\label{fig:realsense_singleir}
\end{figure*}

\added{Qualitative results are shown in Figure~\ref{fig:realsense_singleir}. Due to imperfect calibration (e.g., misalignment of the wall and projector, inaccuracies in wall distance, or dirt on the wall), the quality may be degraded. Interestingly, Figure~\ref{fig:realsense_singleir} shows noisy predictions on the left side of the white wall where the projected pattern is less visible under strong ambient light. Nevertheless, our method still achieves better reconstructions in these regions compared to the original Realsense that uses both IR images.}

\setlength{\tabcolsep}{1pt}

\renewcommand{\arraystretch}{0.1}

\newlength{\imgwidth}
\setlength{\imgwidth}{0.162\textwidth}

\renewcommand{\formattedgraphics}[1]{%
      \includegraphics[width=\imgwidth,keepaspectratio]{#1}%
}

\begin{figure*}[htb]
  \centering
  \begin{tabular}{cccccc}
    \textbf{Left RGB} & \textbf{Left IR} & \textbf{GT Depth} & \textbf{Left RGB} & \textbf{Left IR} & \textbf{GT Depth} \\[1.5pt]
    \formattedgraphics{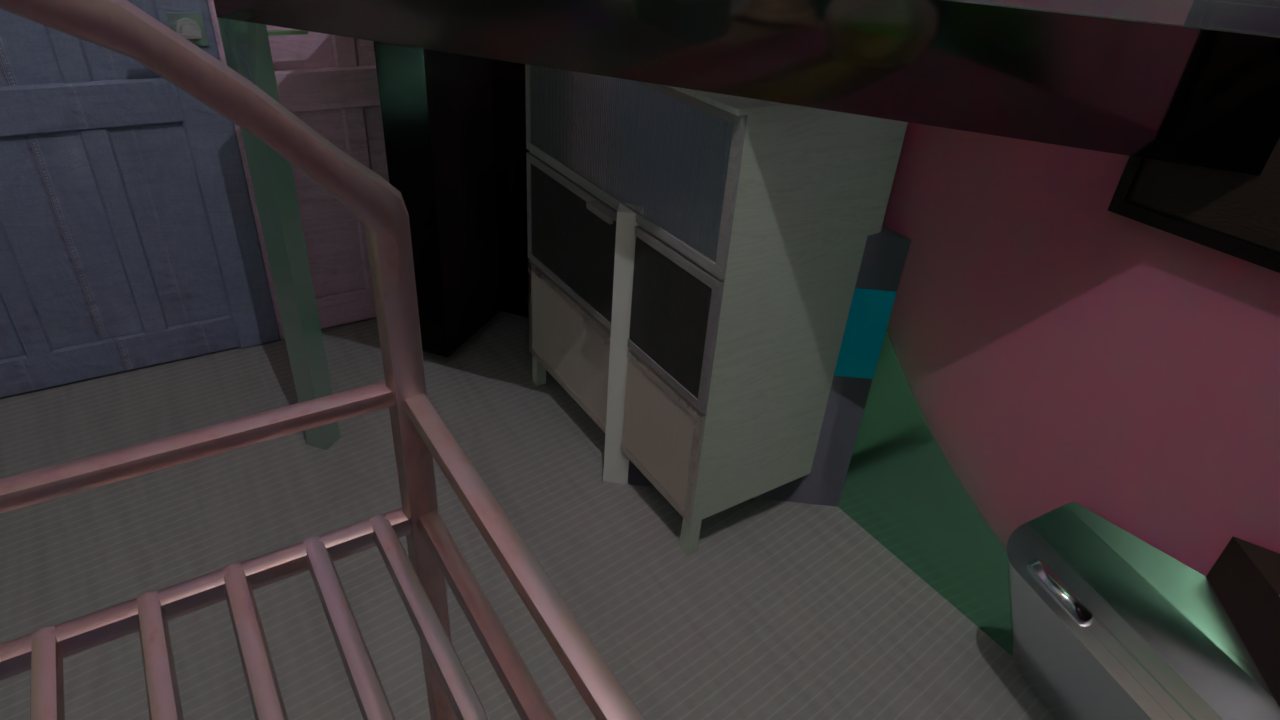} &
    \formattedgraphics{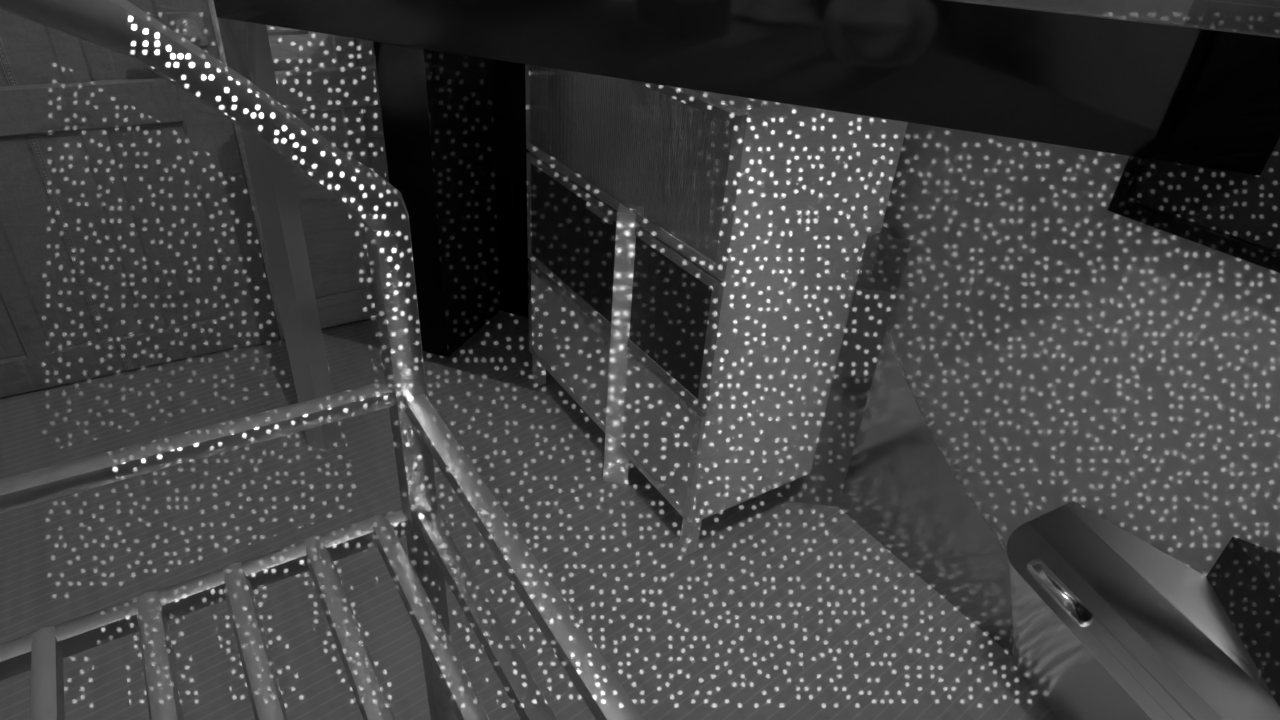} &
    \formattedgraphics{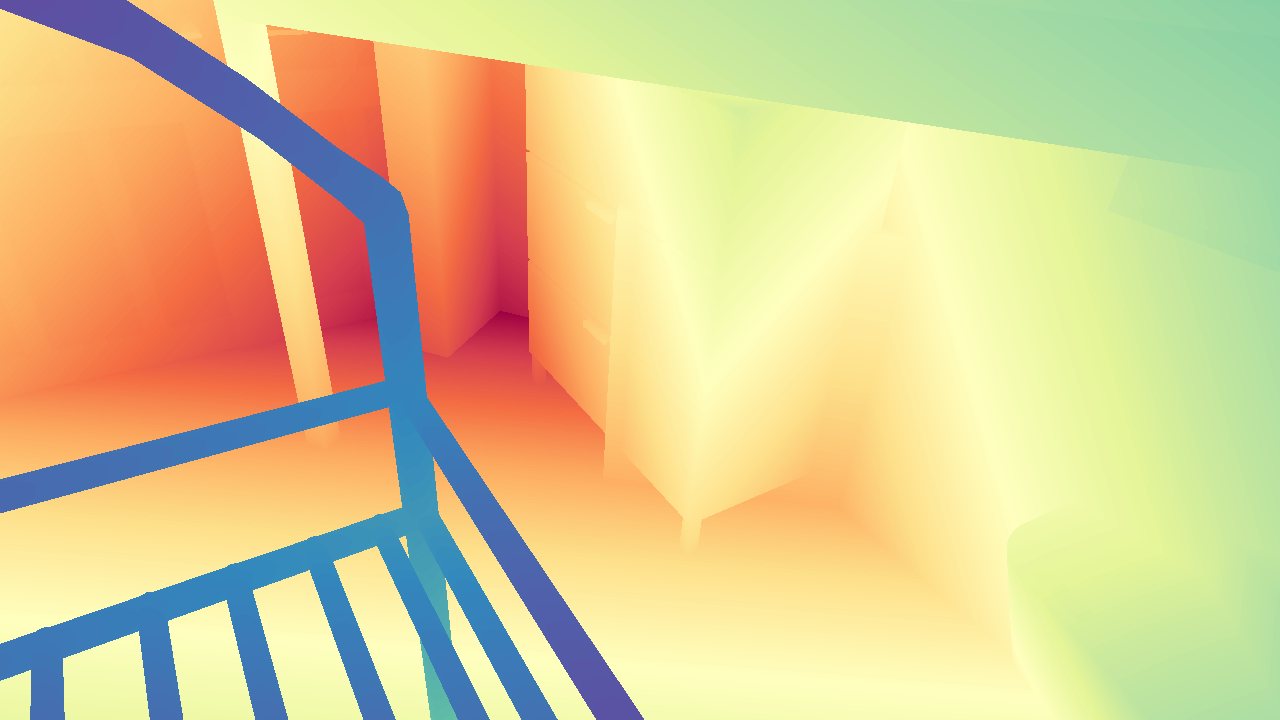} &
    \formattedgraphics{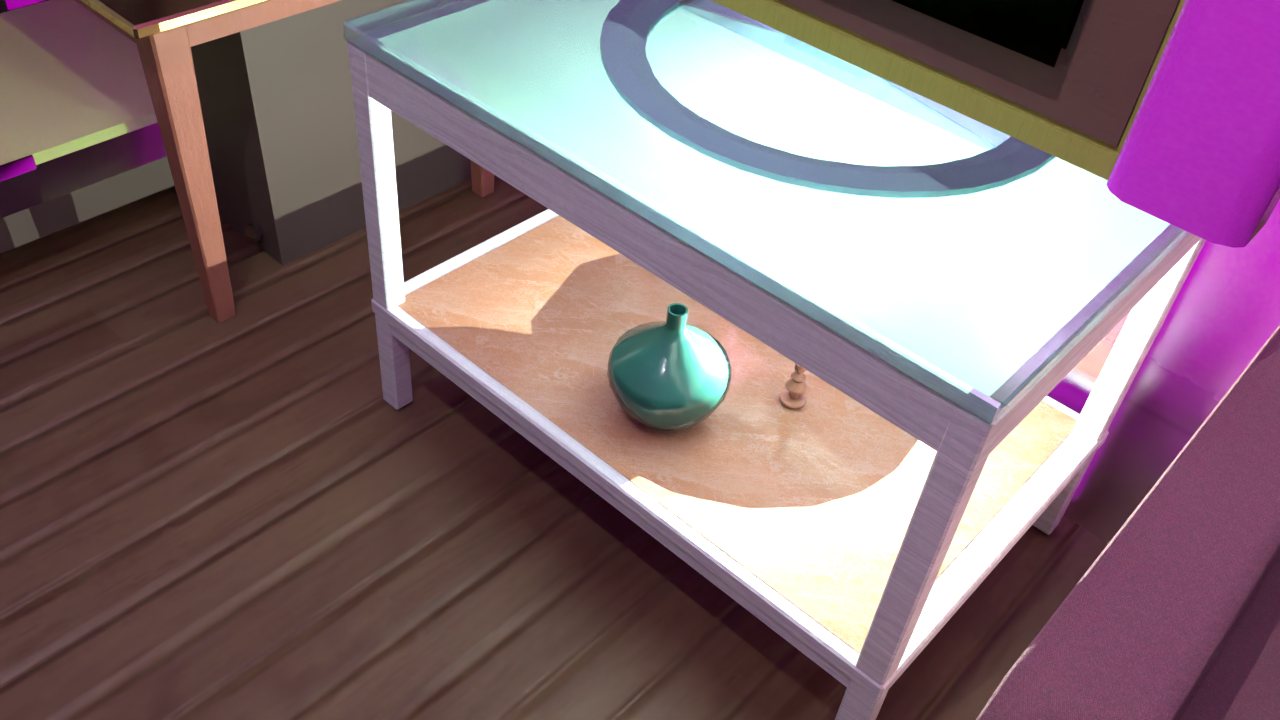} &
    \formattedgraphics{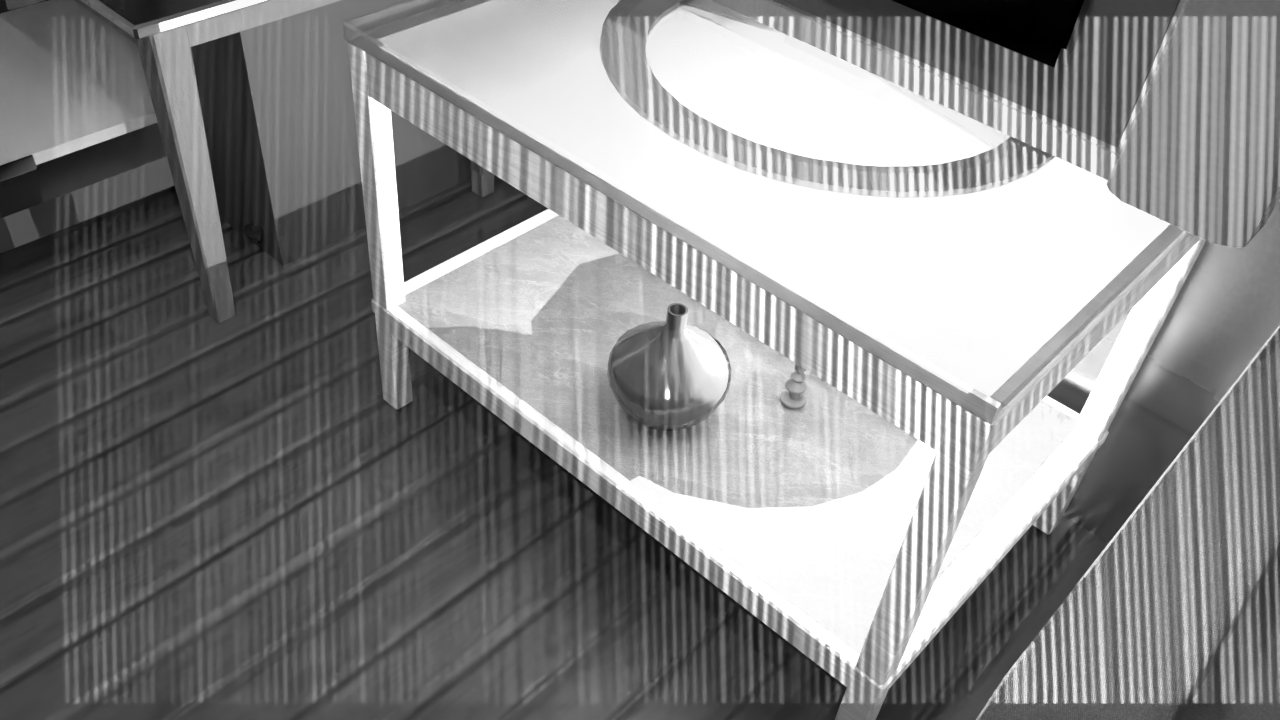} &
    \formattedgraphics{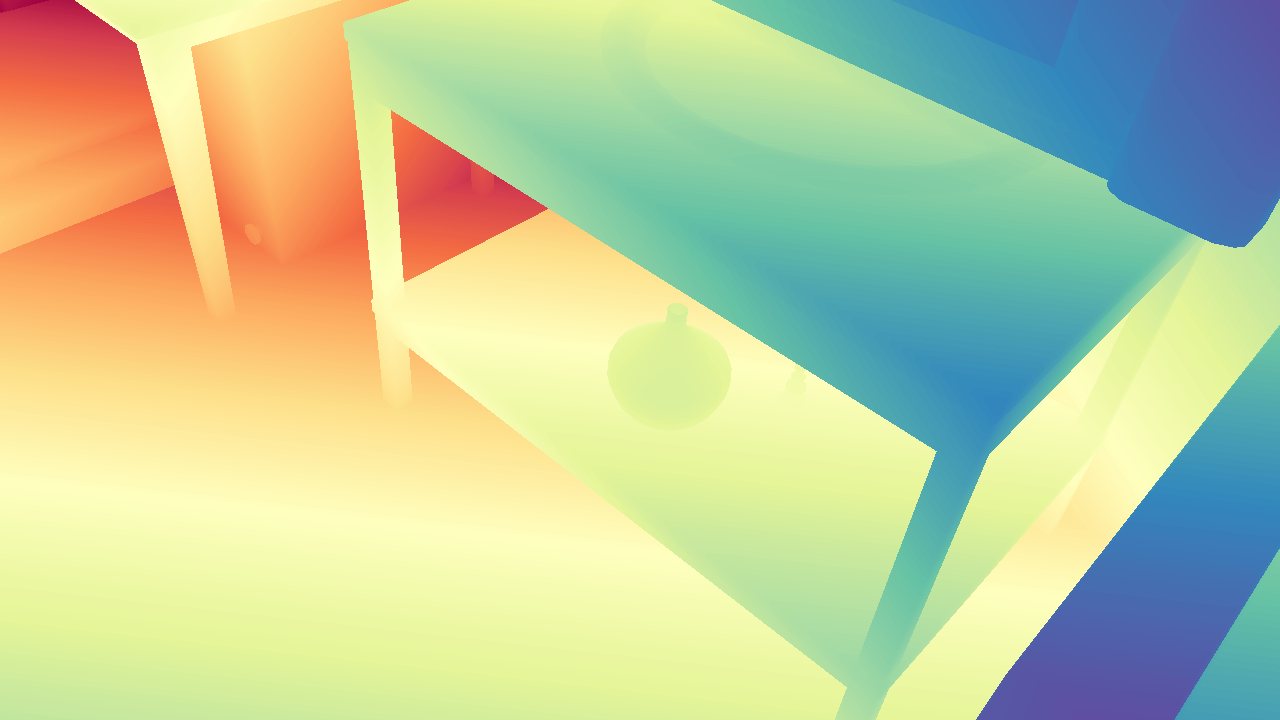} \\[1pt]
    \formattedgraphics{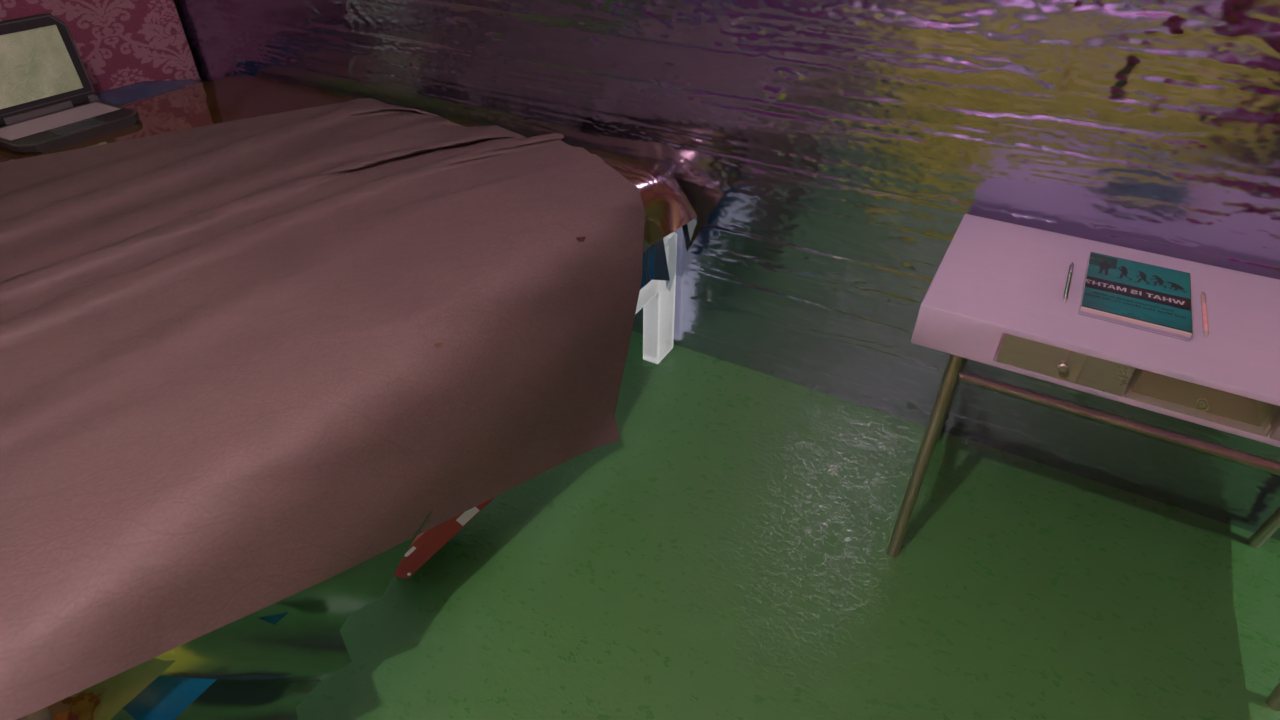} &
    \formattedgraphics{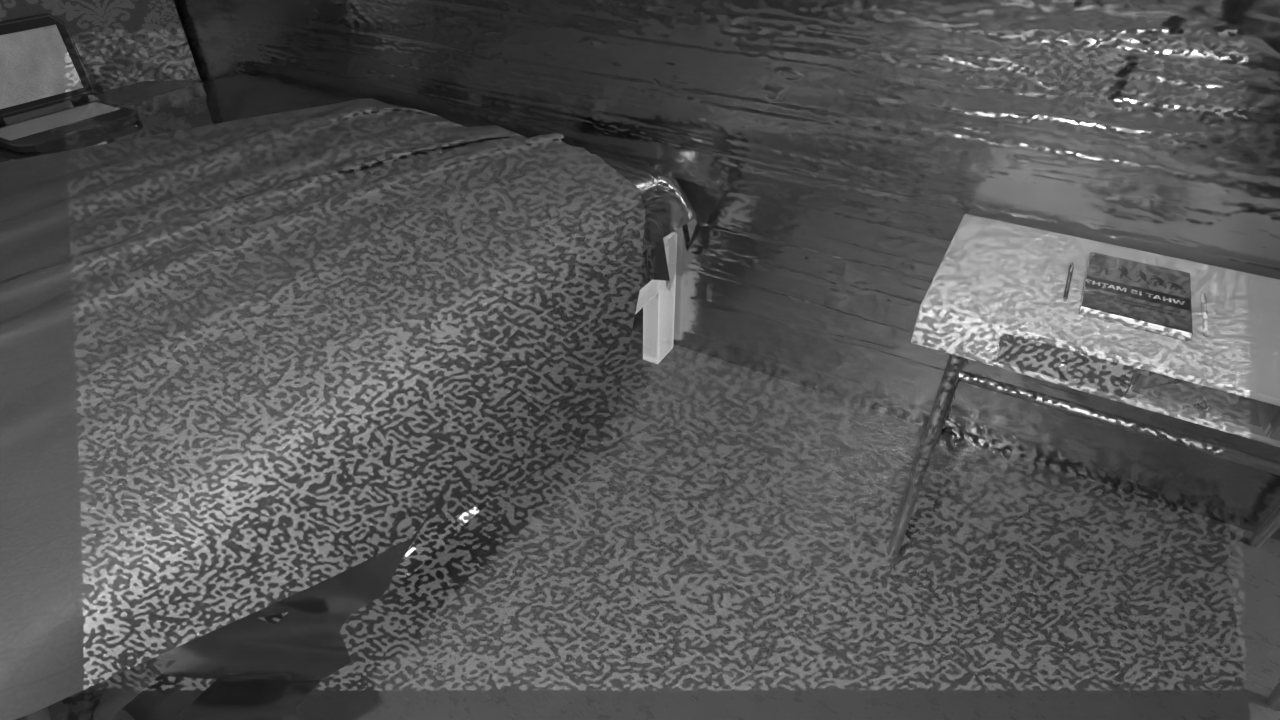} &
    \formattedgraphics{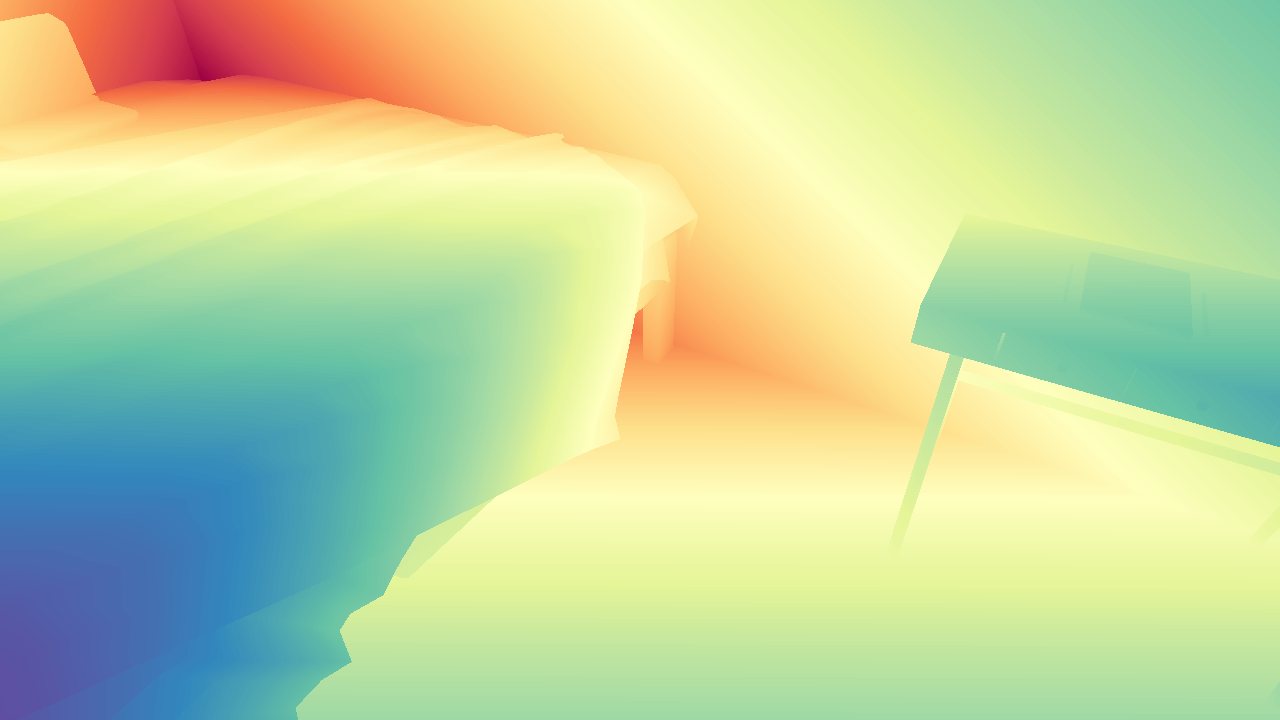} &
    \formattedgraphics{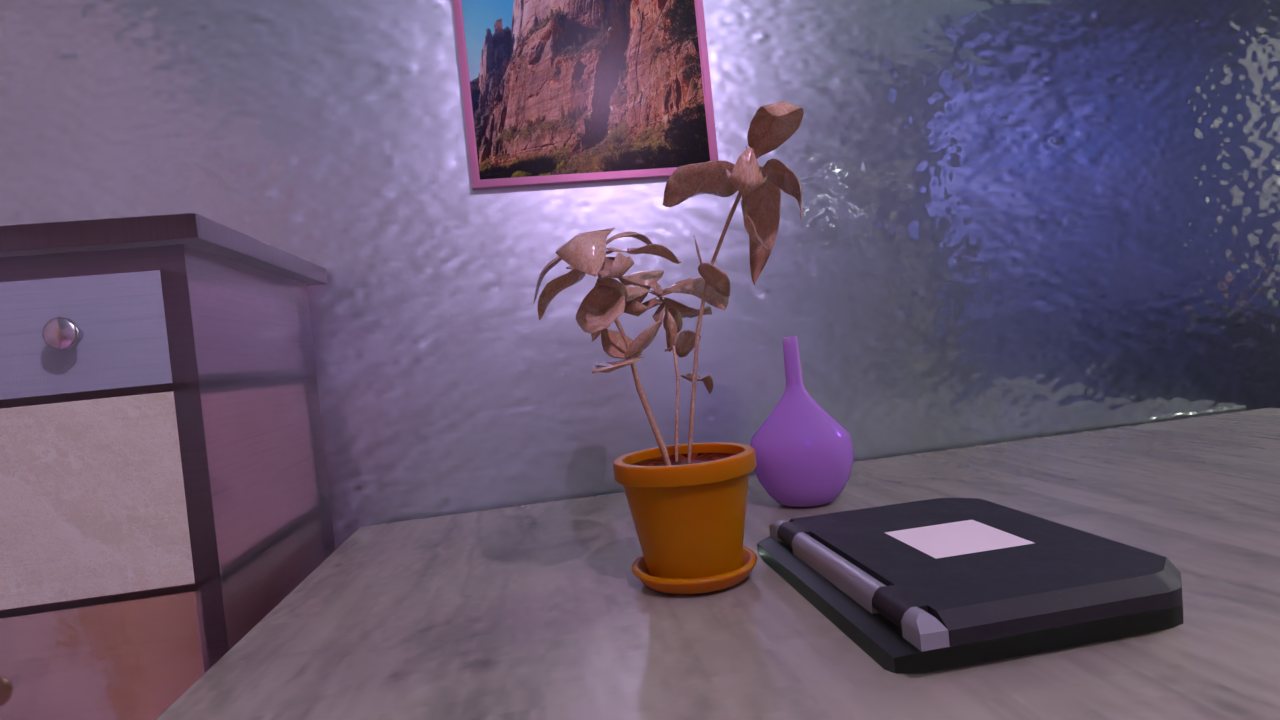} &
    \formattedgraphics{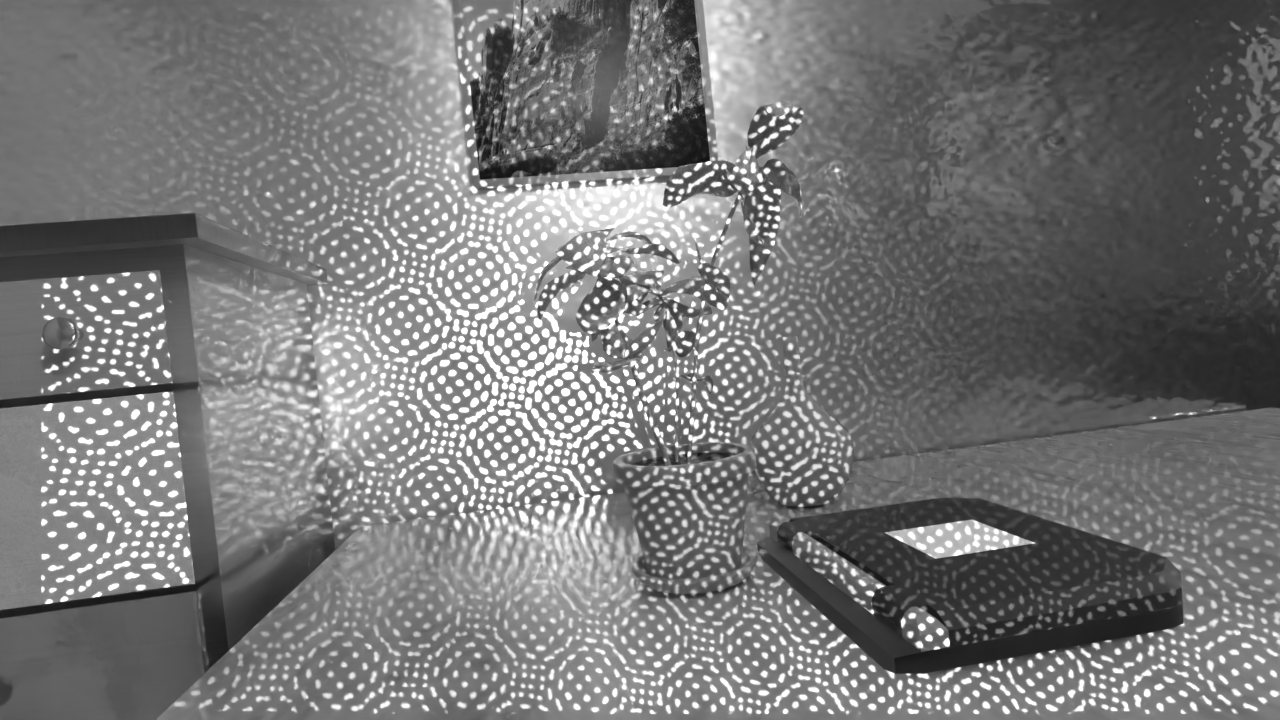} &
    \formattedgraphics{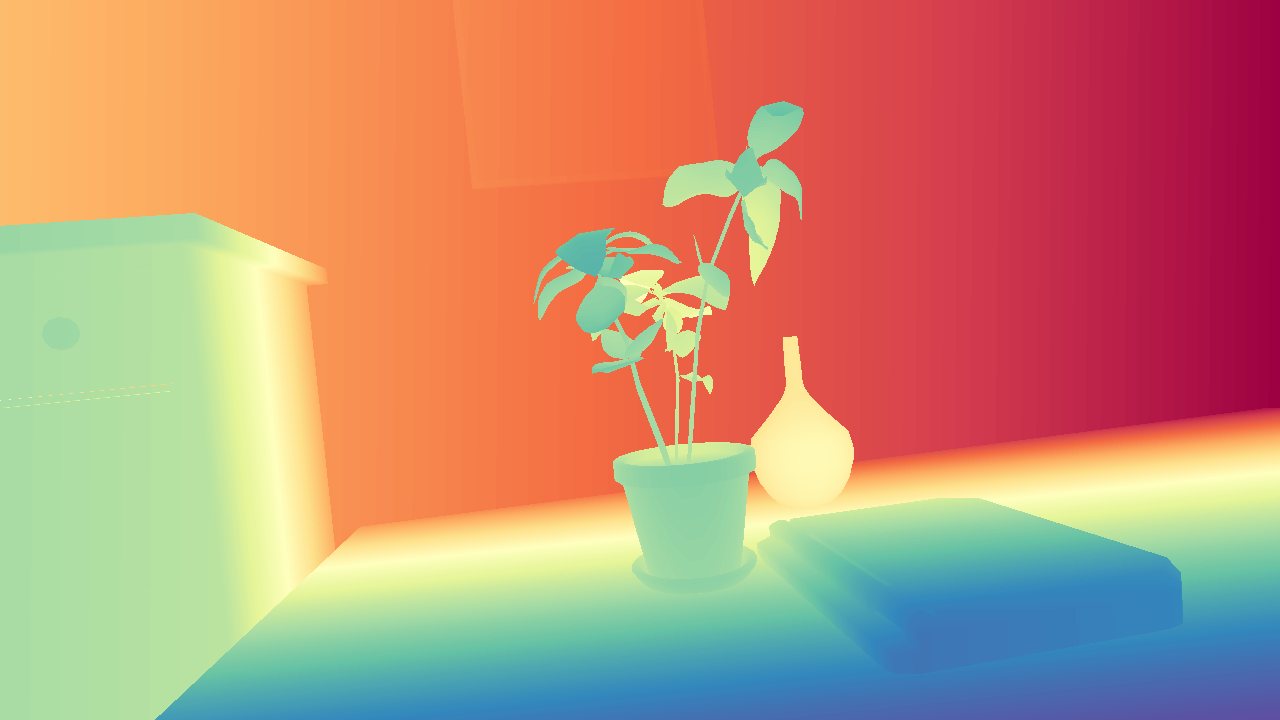} \\[1pt]
    \formattedgraphics{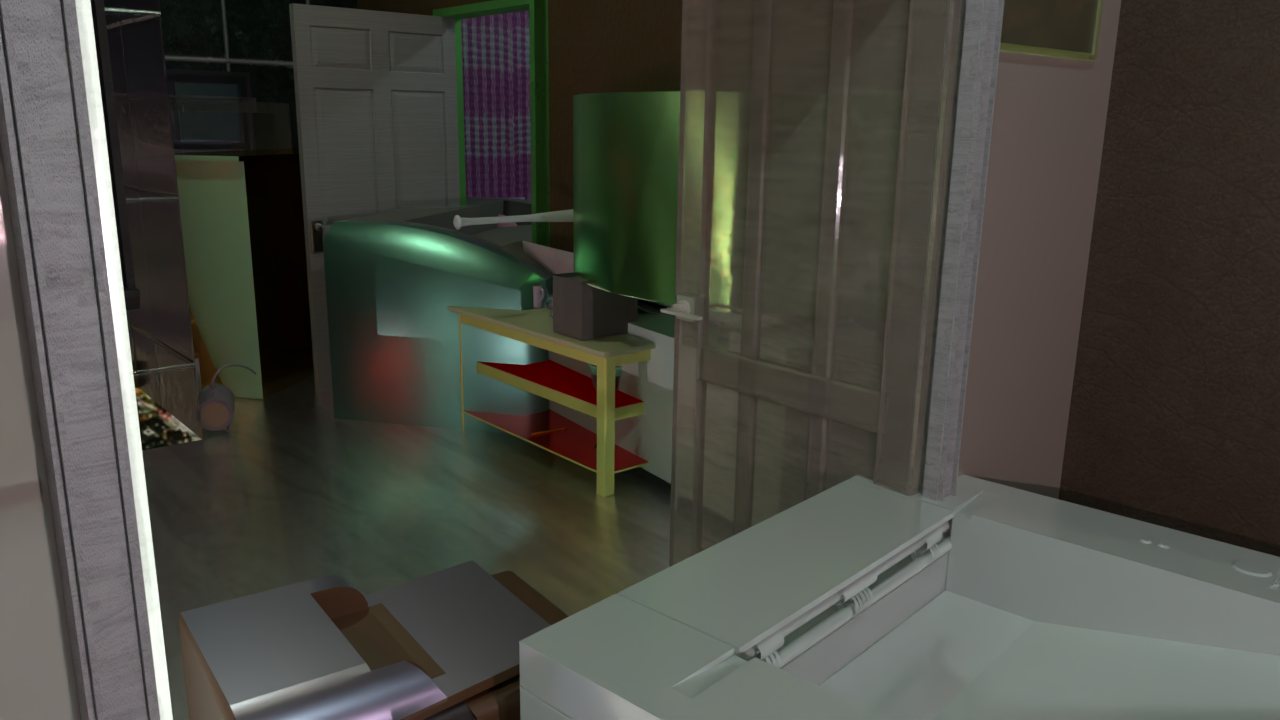} &
    \formattedgraphics{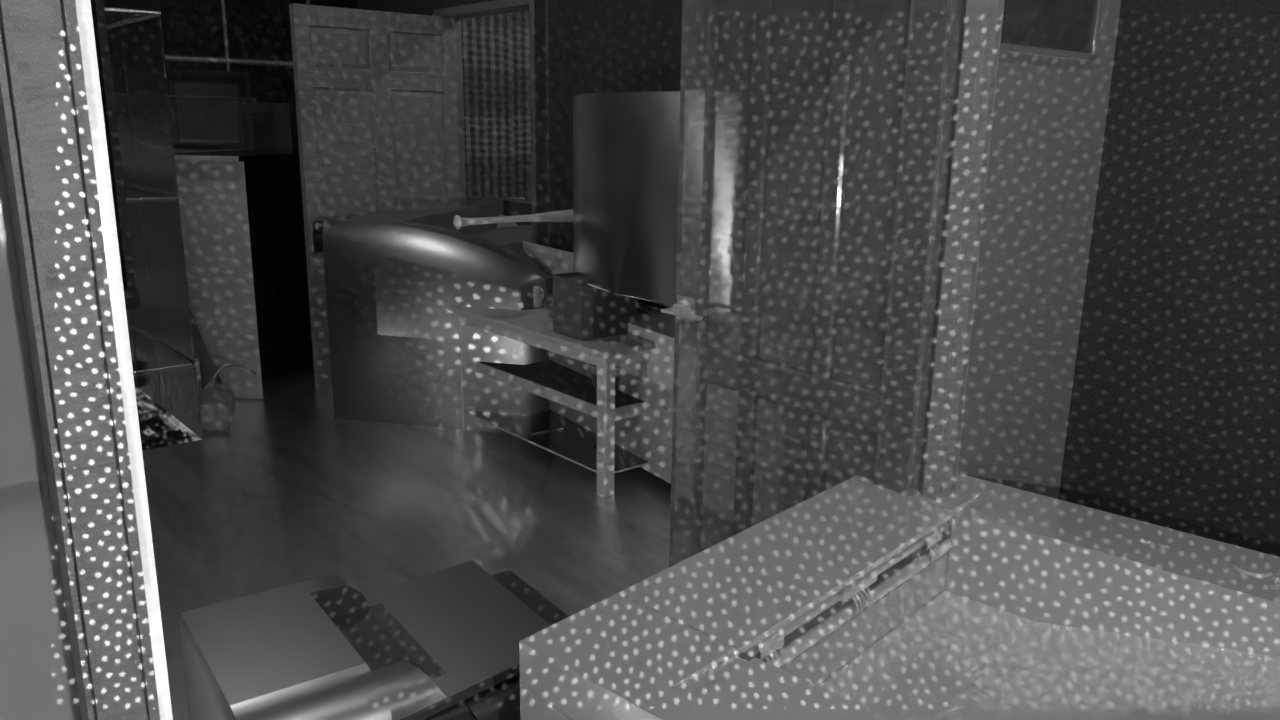} &
    \formattedgraphics{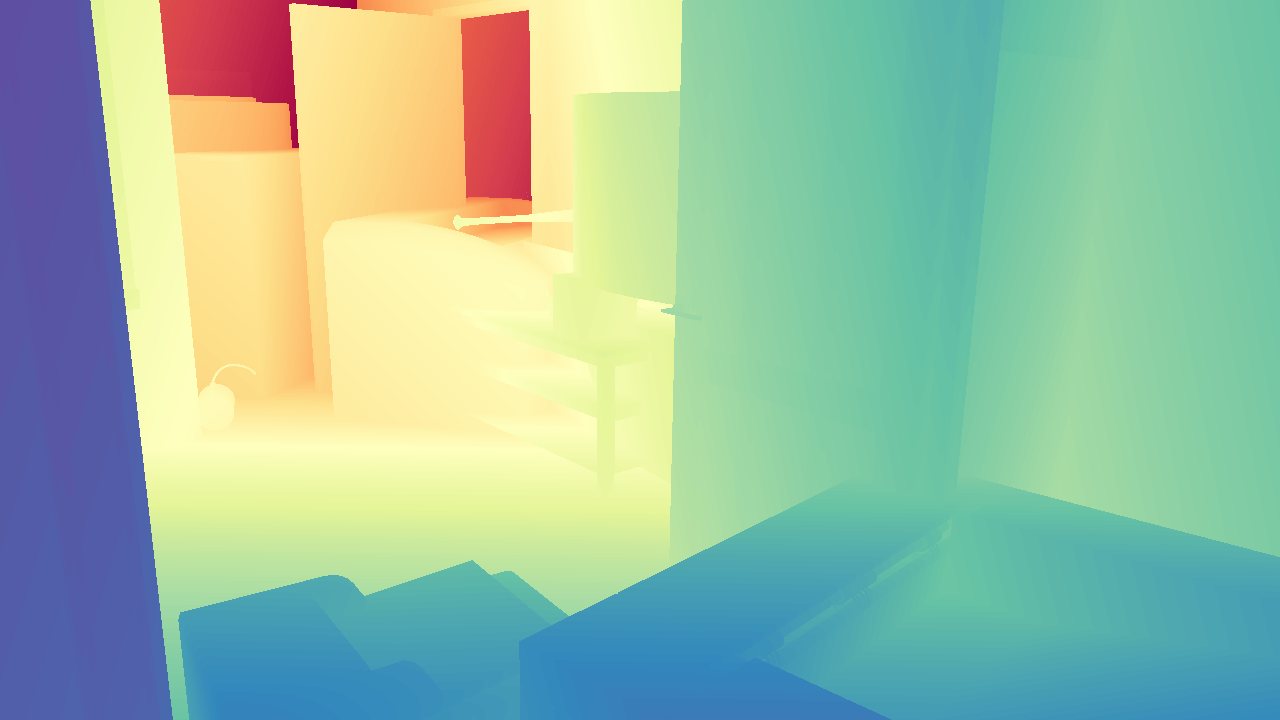} &
    \formattedgraphics{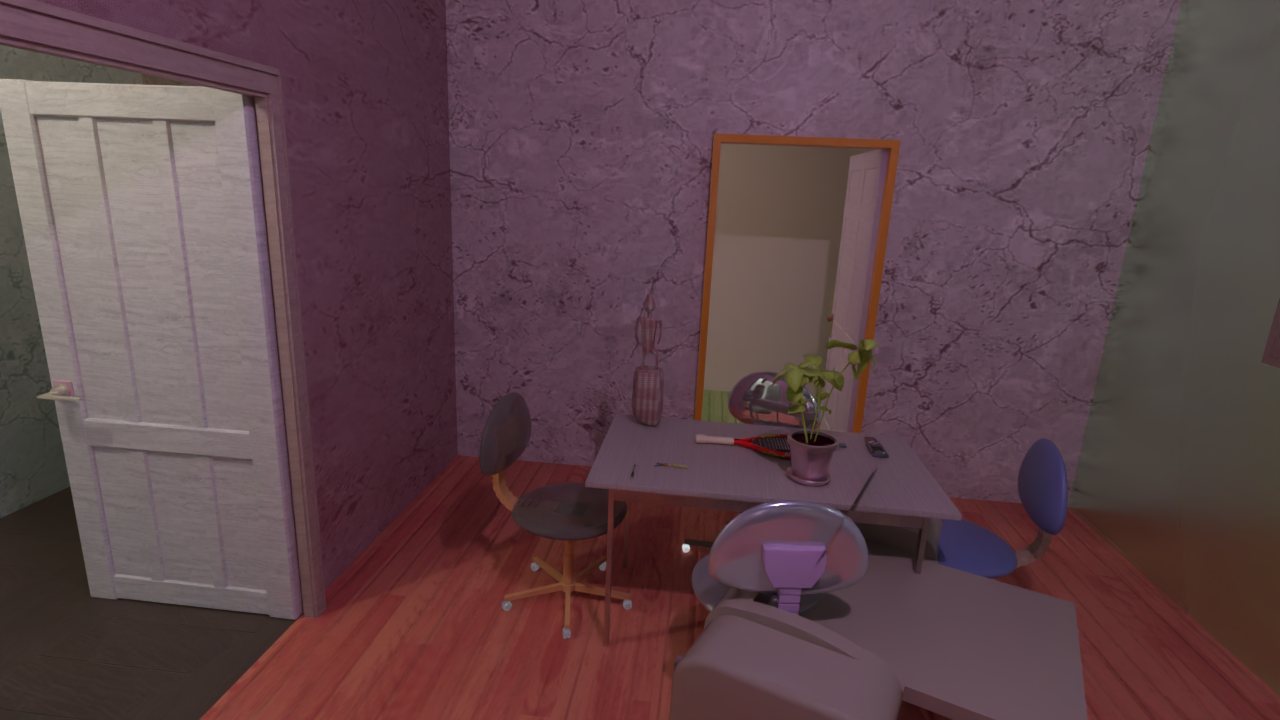} &
    \formattedgraphics{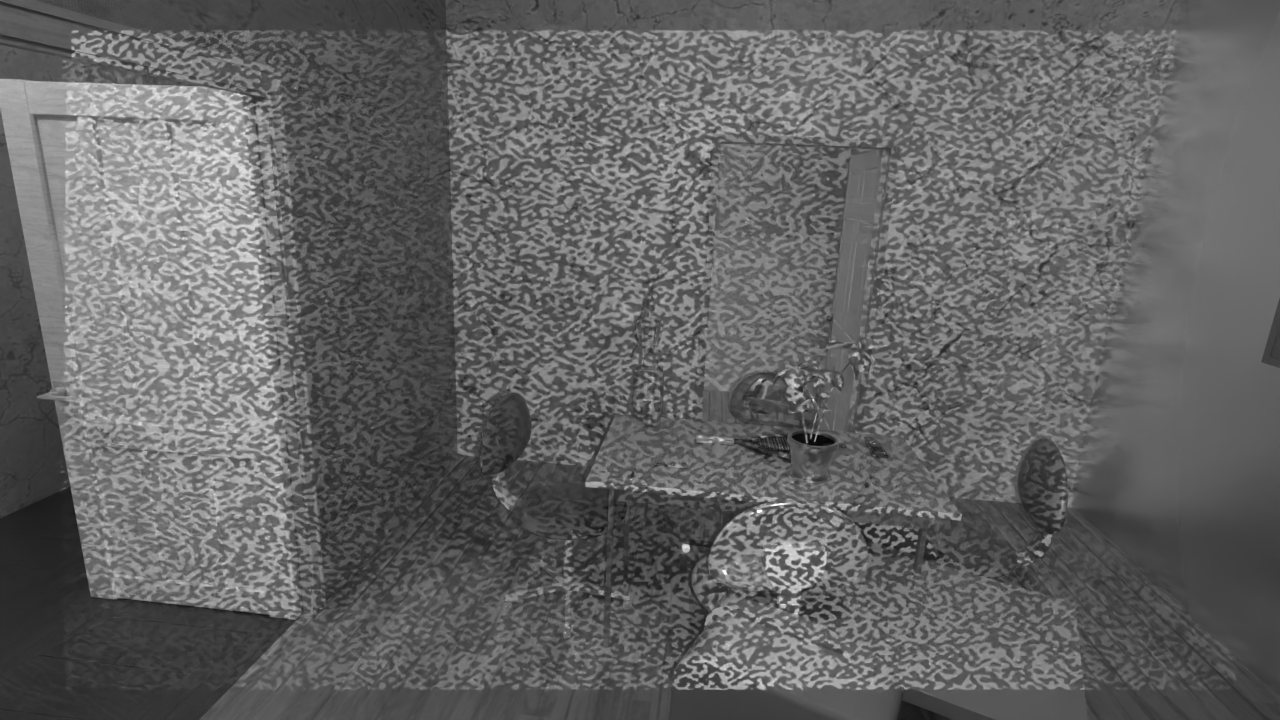} &
    \formattedgraphics{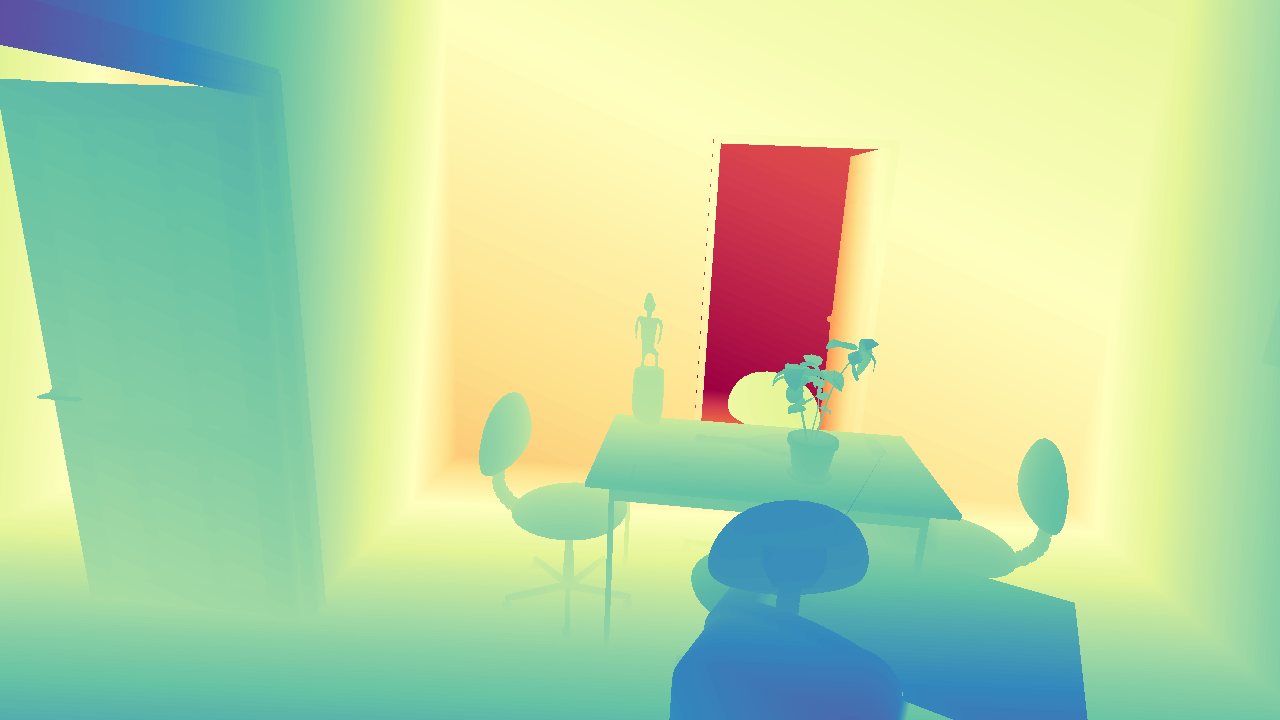} \\[1pt]
    \formattedgraphics{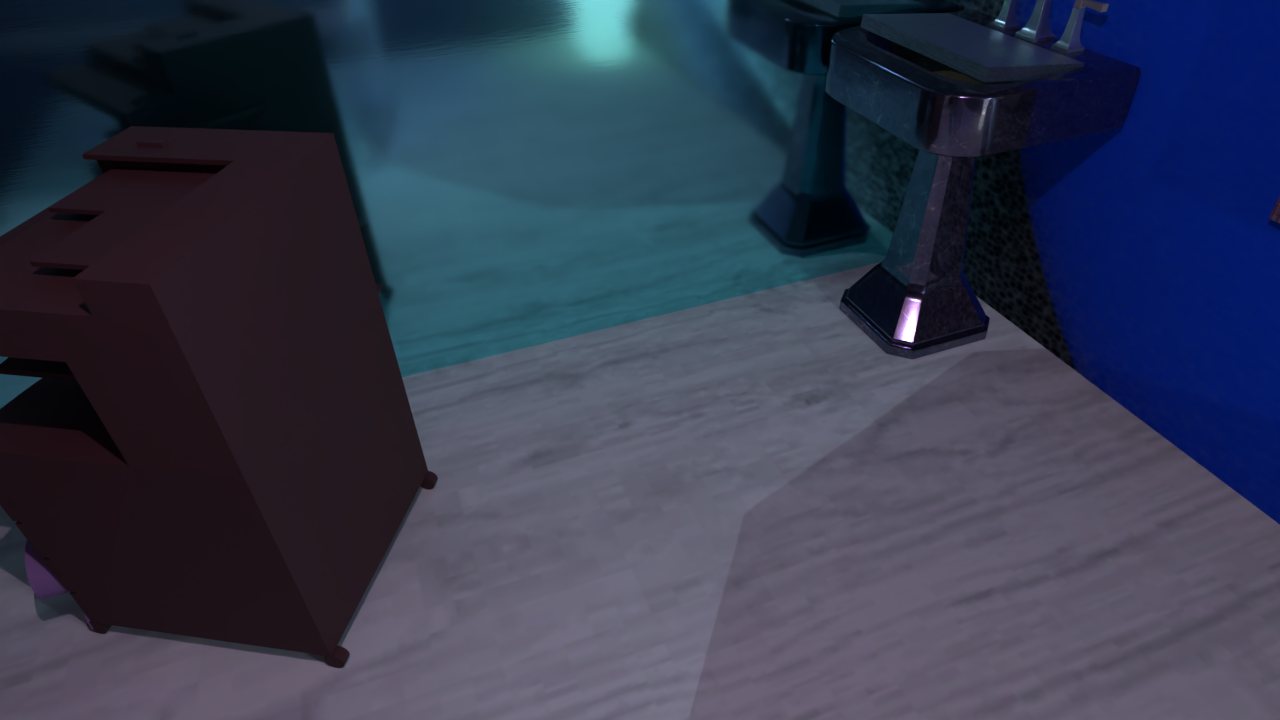} &
    \formattedgraphics{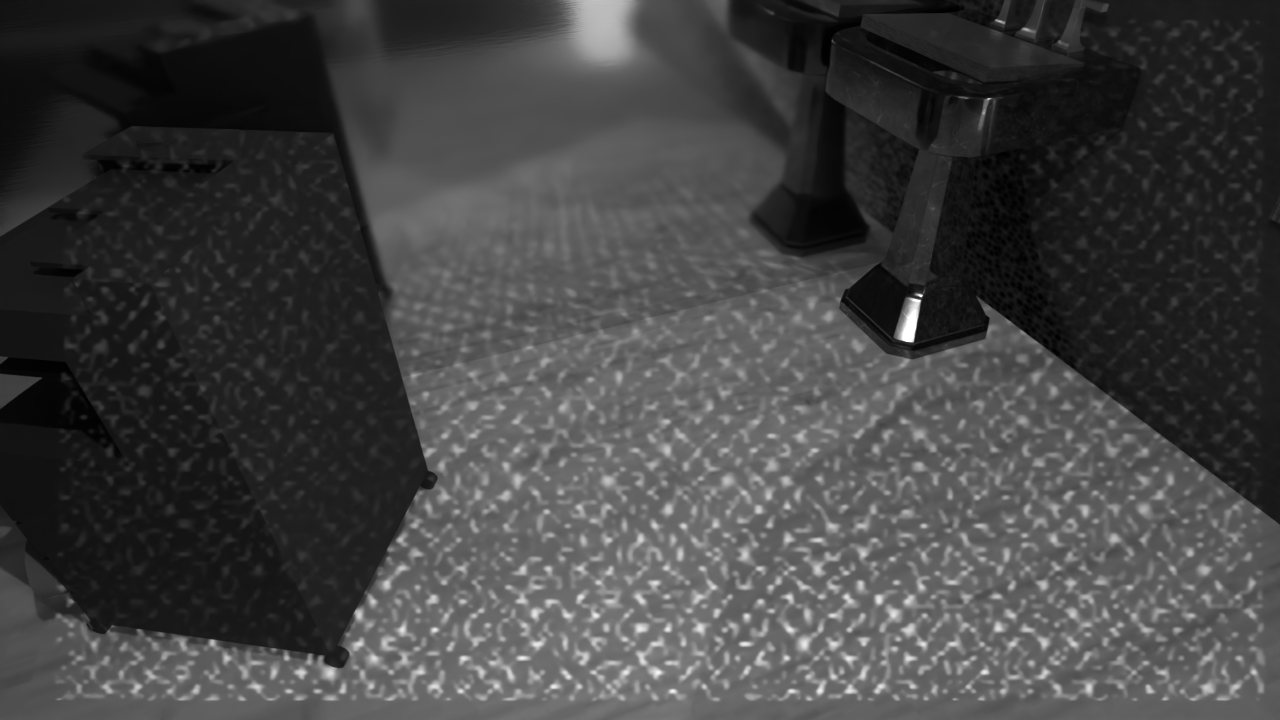} &
    \formattedgraphics{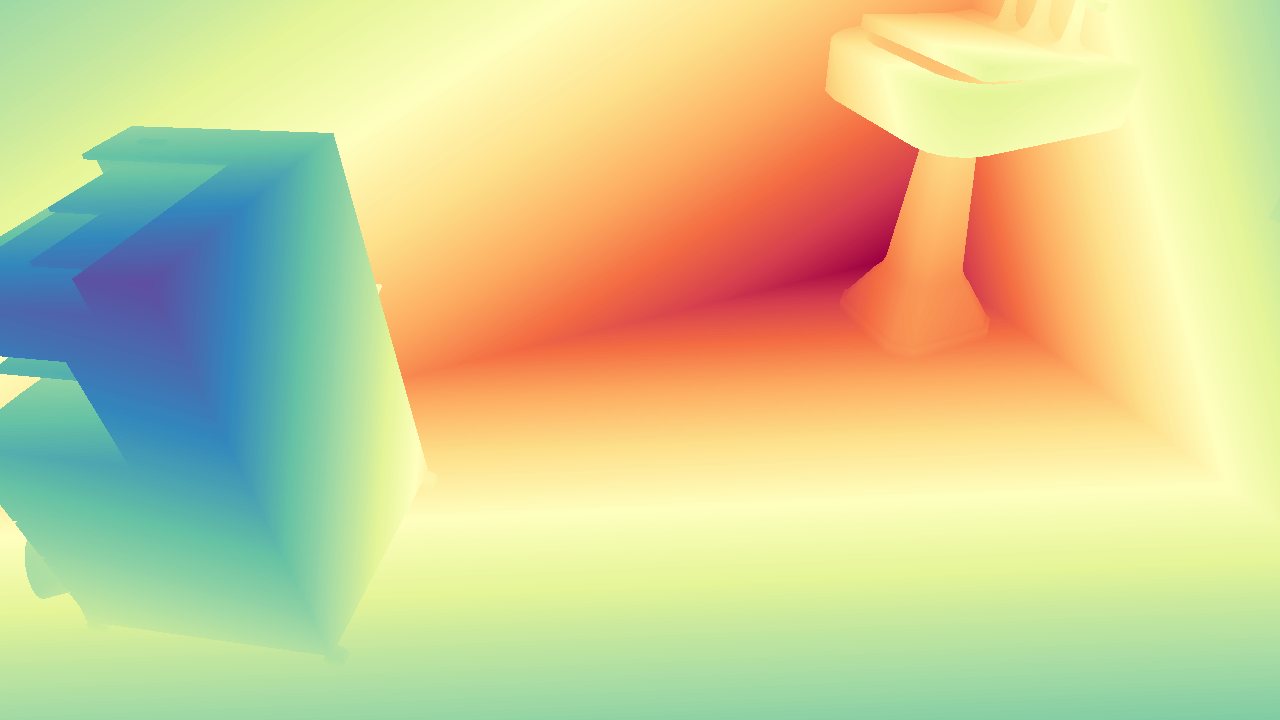} &
    \formattedgraphics{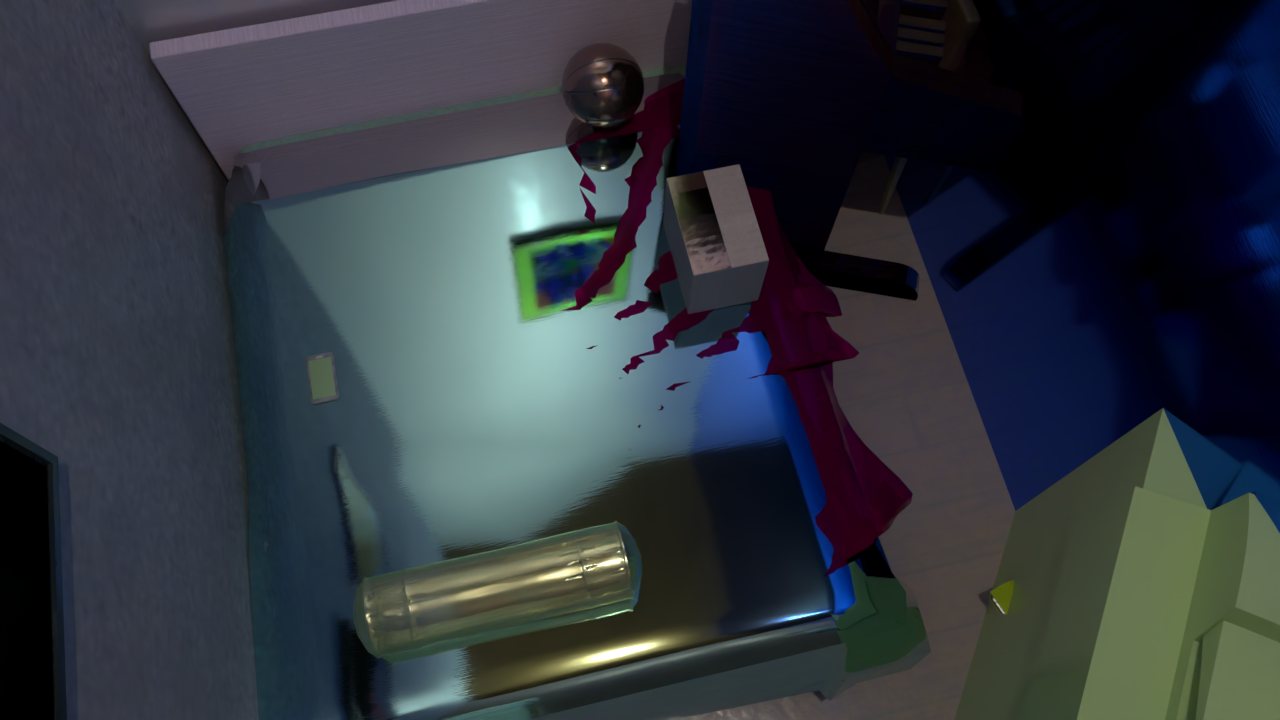} &
    \formattedgraphics{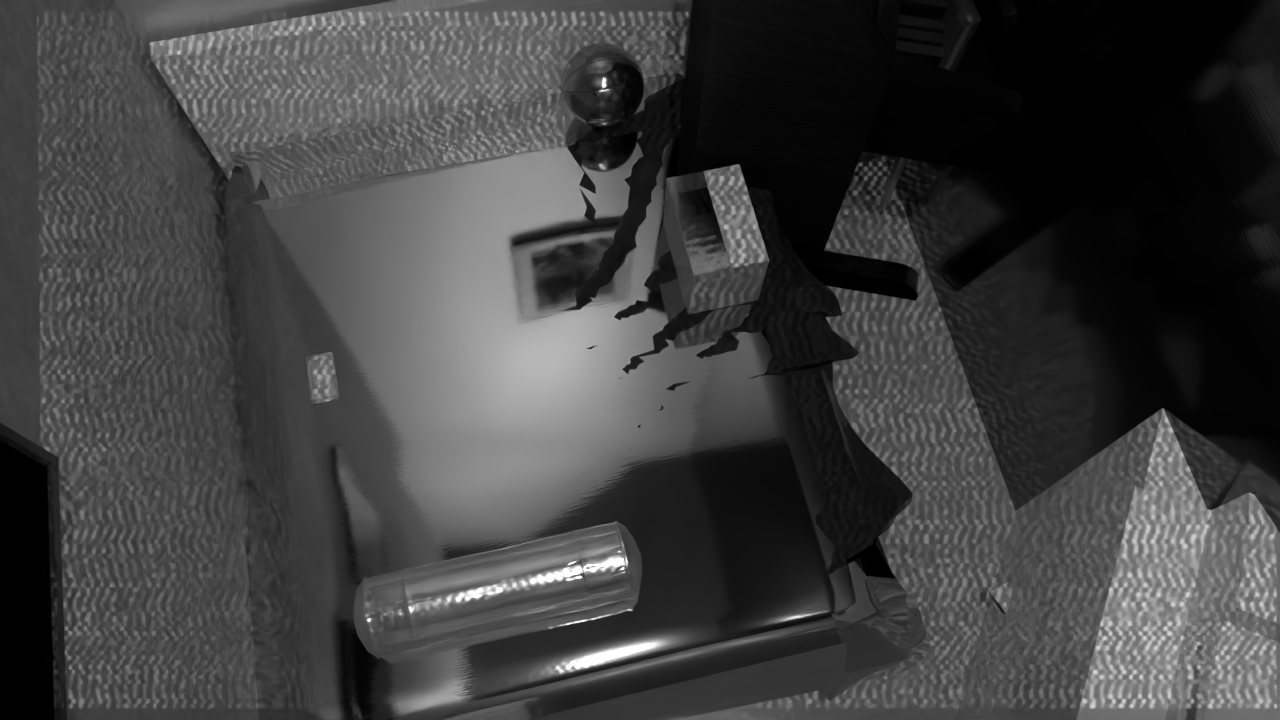} &
    \formattedgraphics{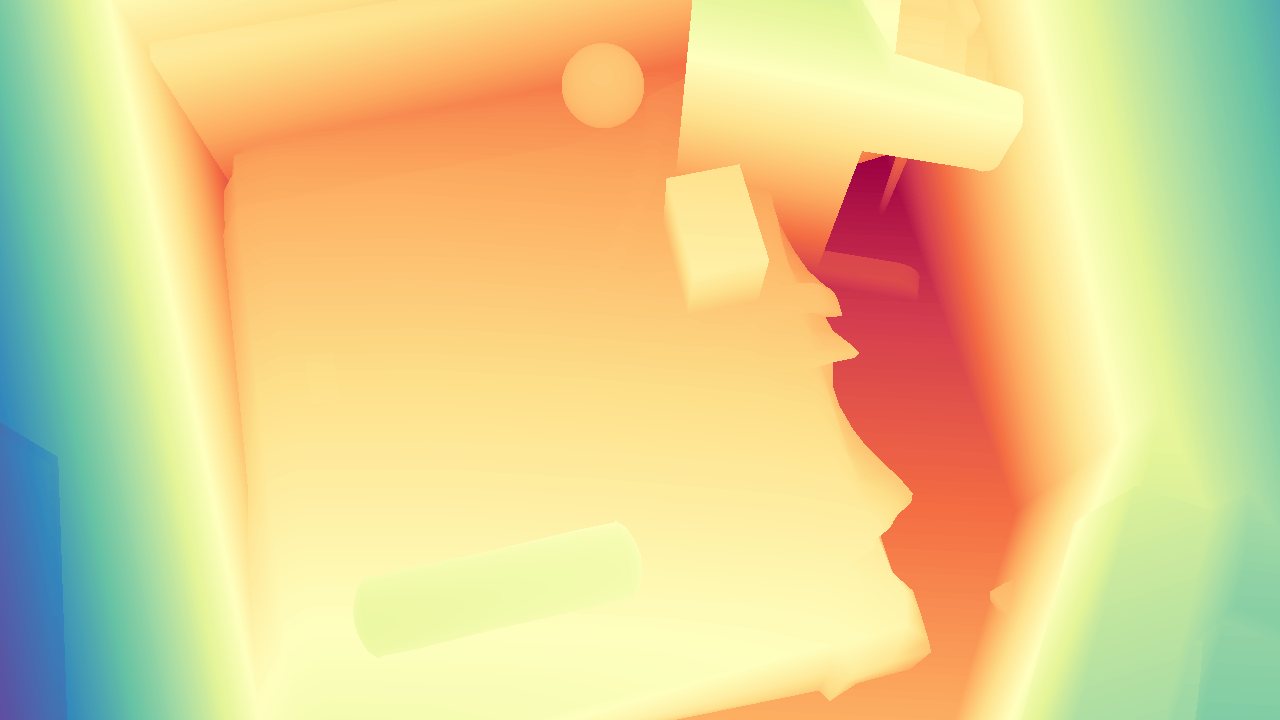} \\[1pt]
    \formattedgraphics{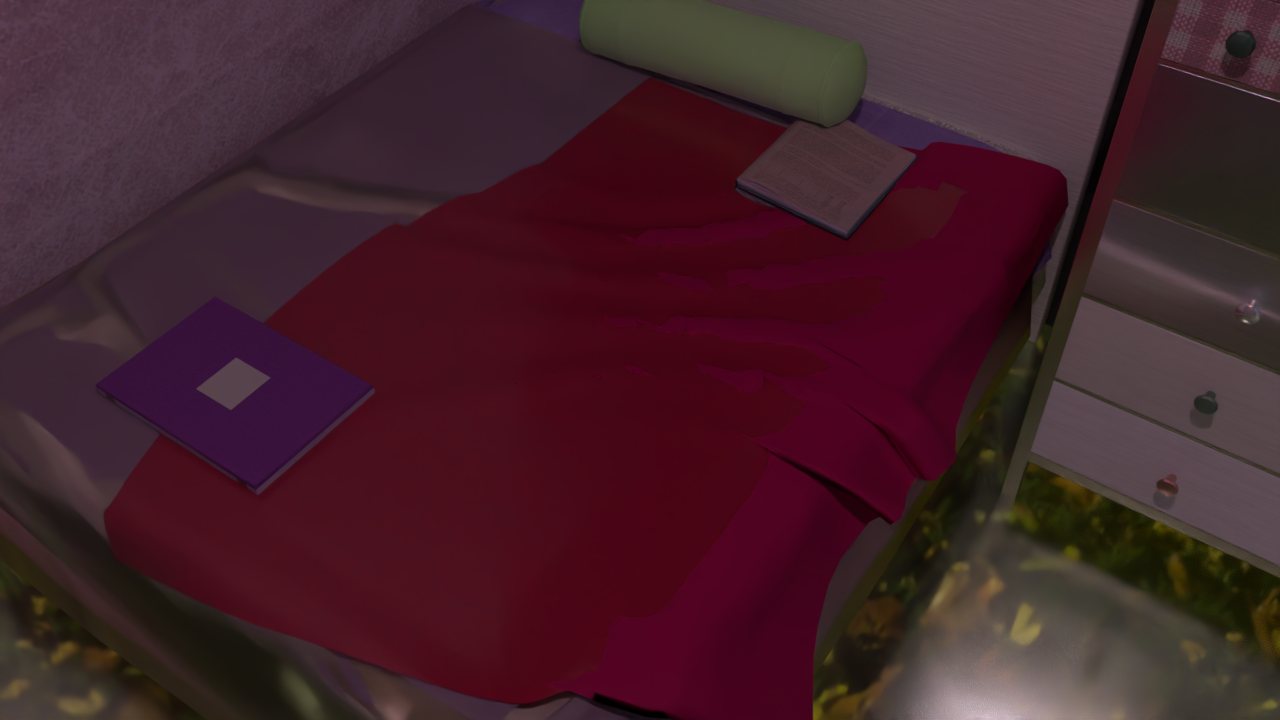} &
    \formattedgraphics{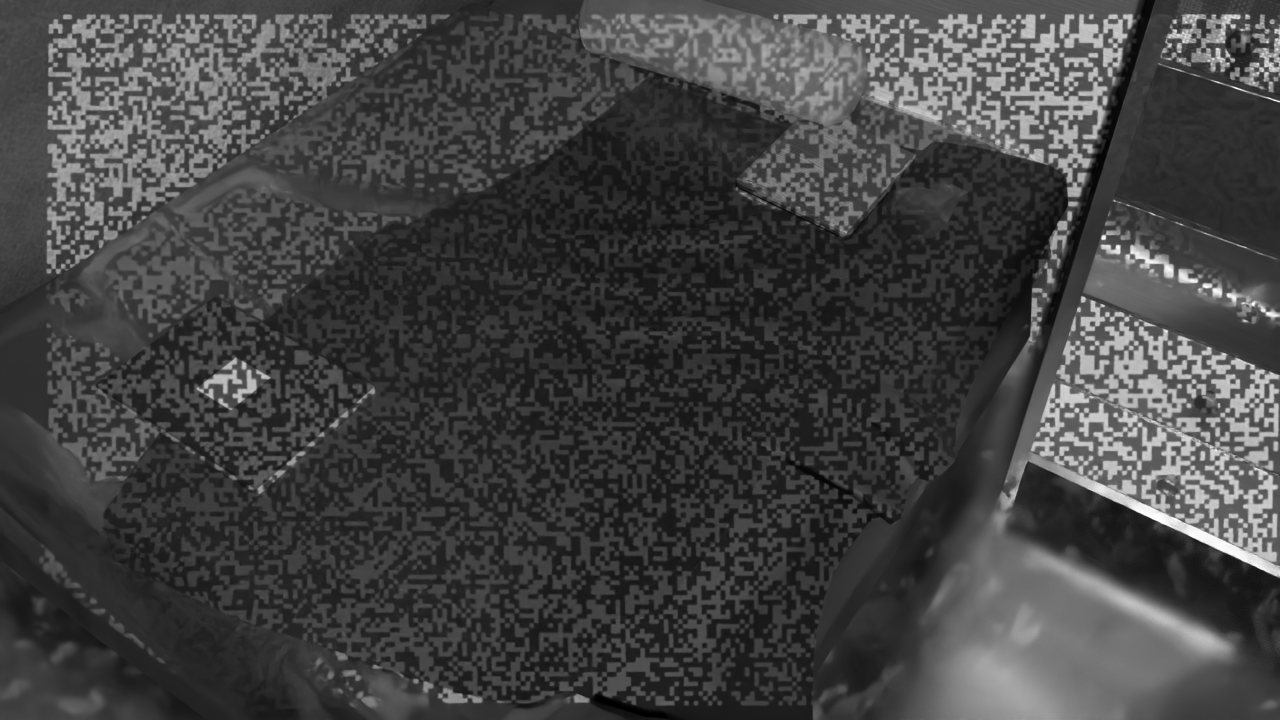} &
    \formattedgraphics{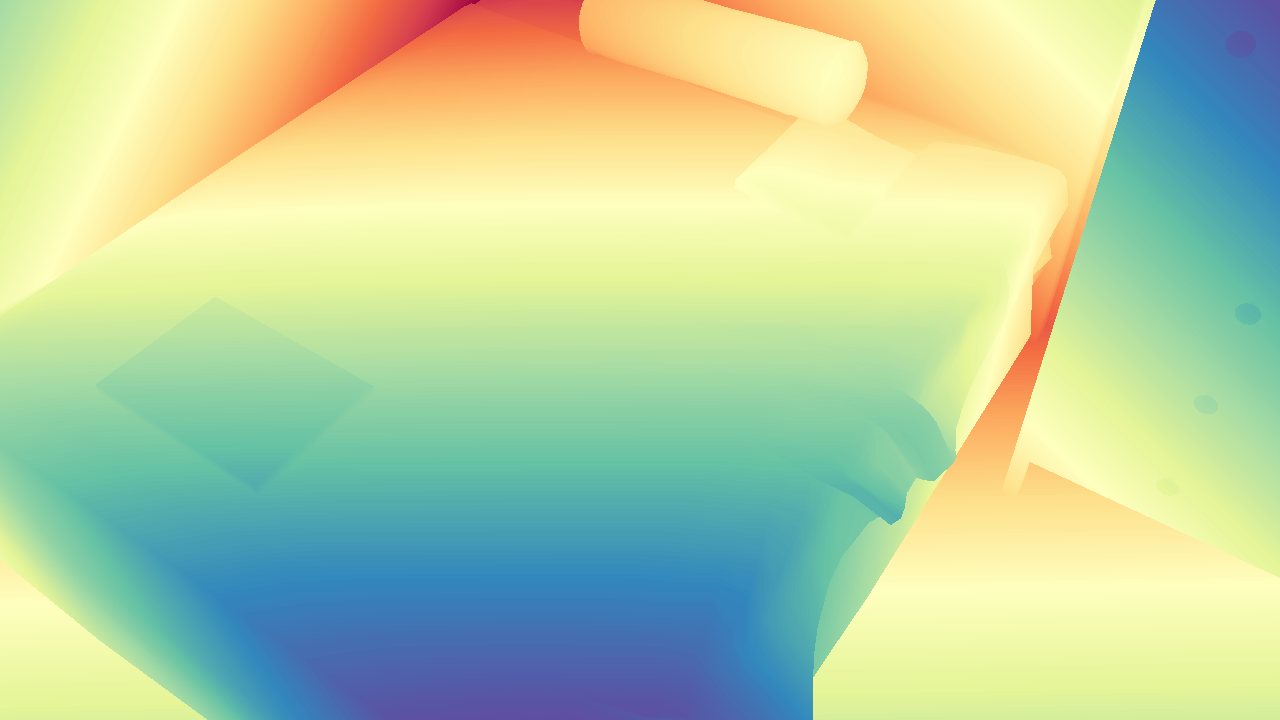} &
    \formattedgraphics{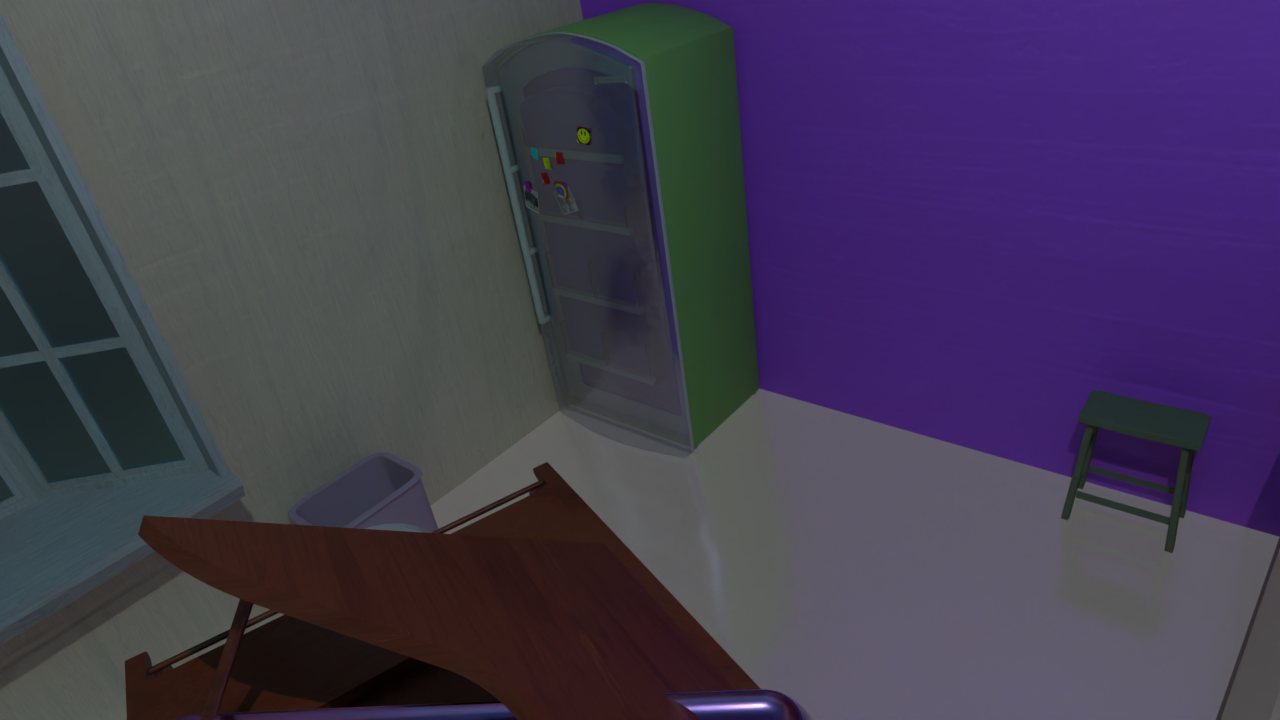} &
    \formattedgraphics{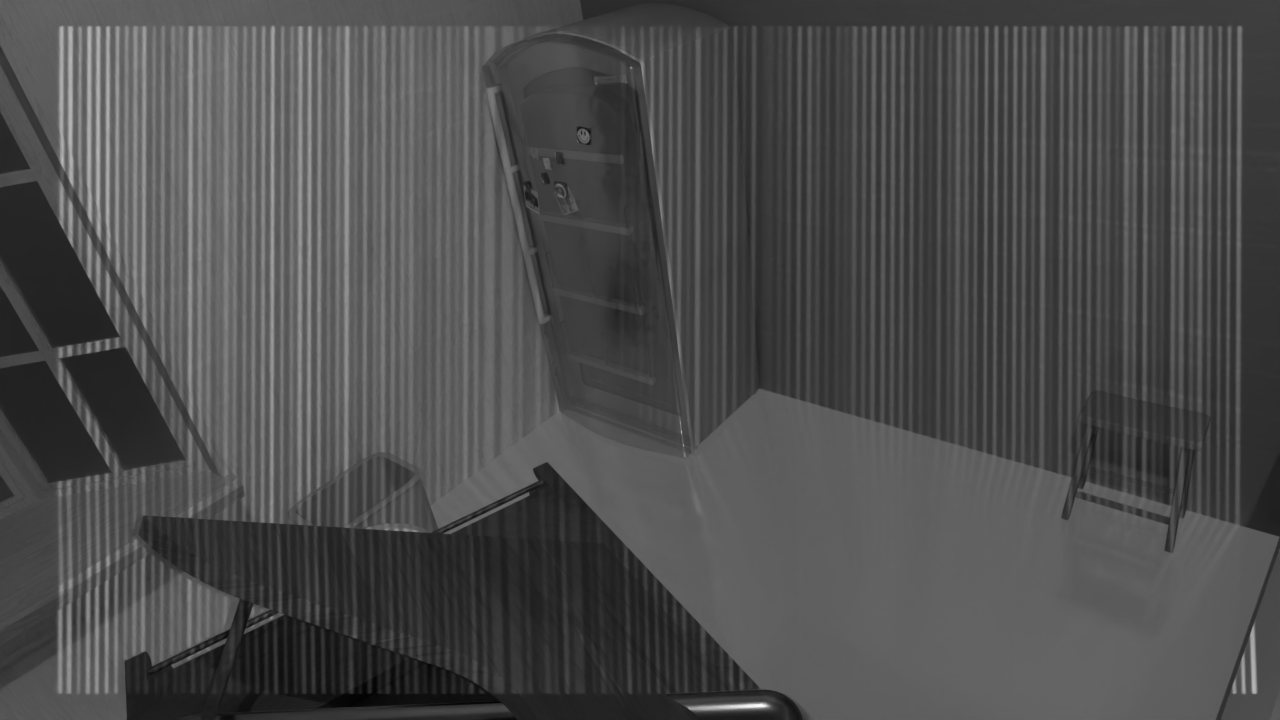} &
    \formattedgraphics{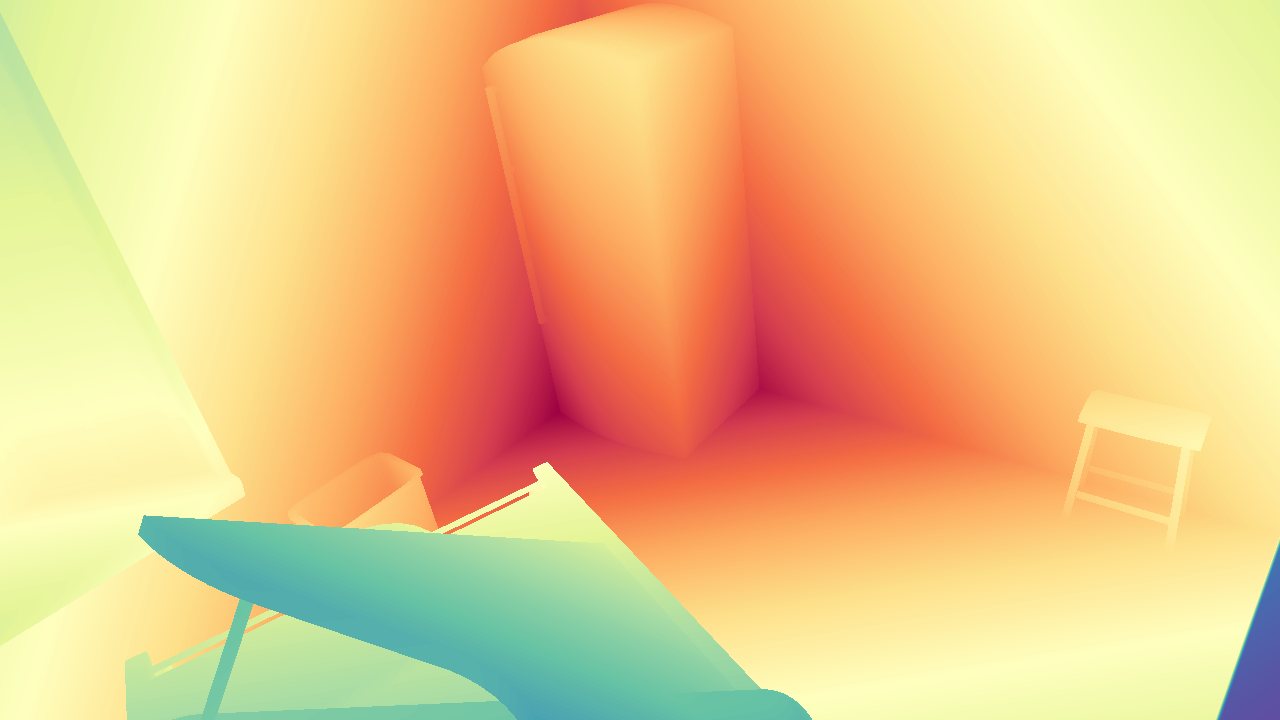} \\[1pt]
    \formattedgraphics{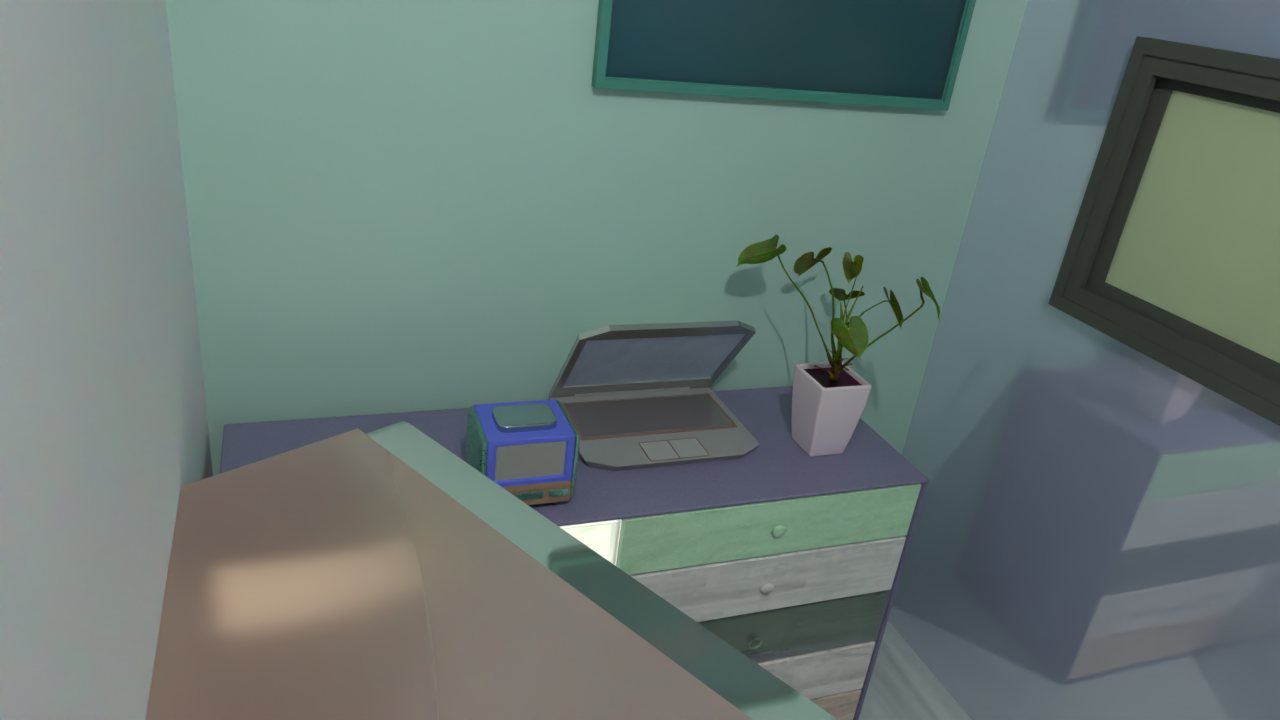} &
    \formattedgraphics{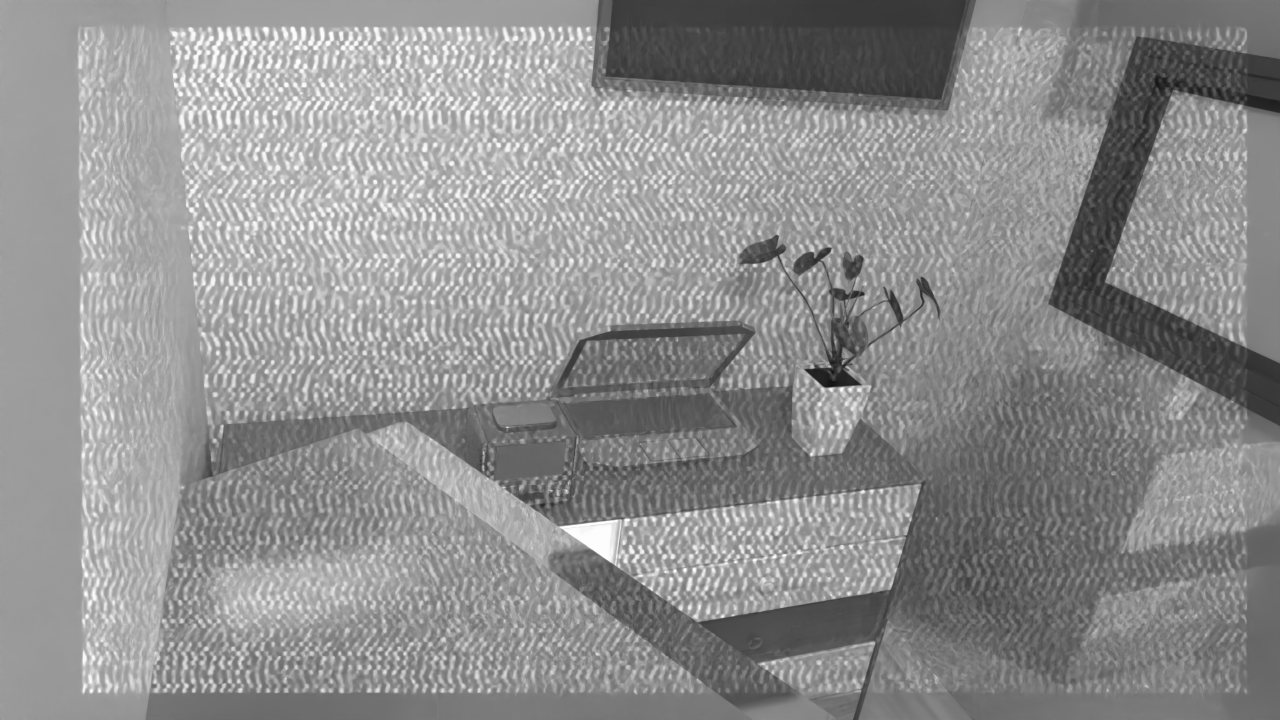} &
    \formattedgraphics{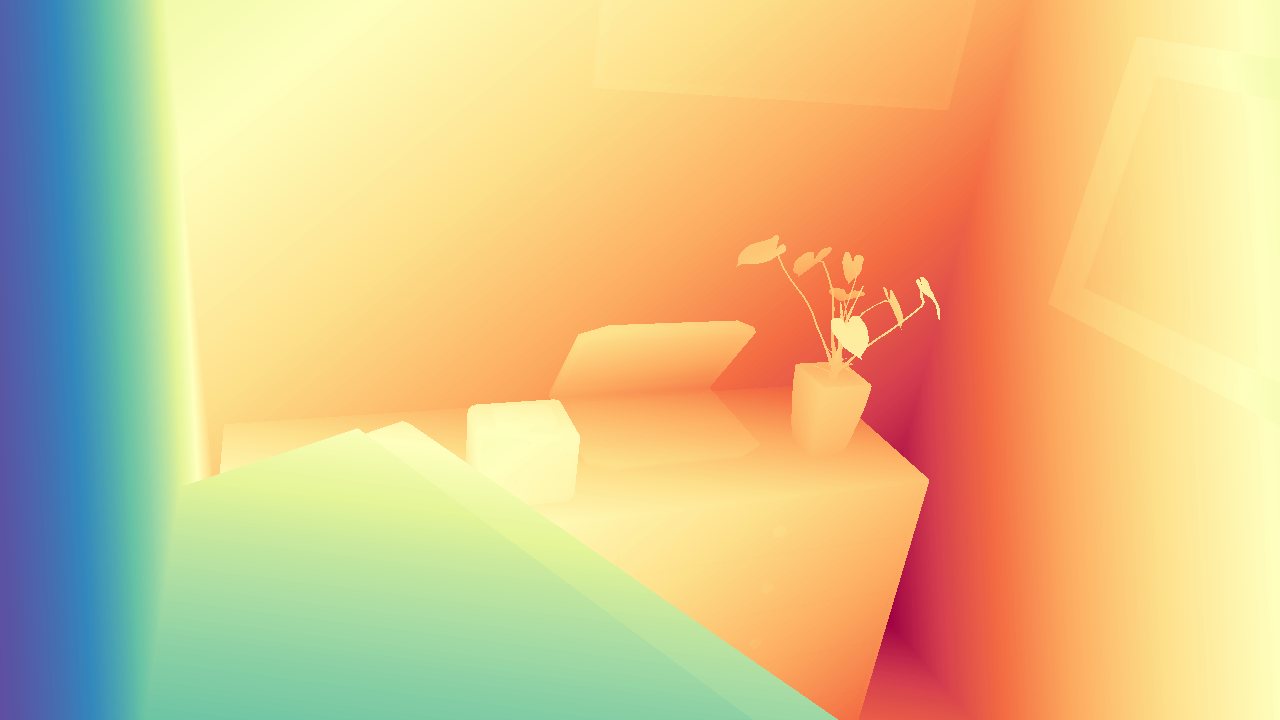} &
    \formattedgraphics{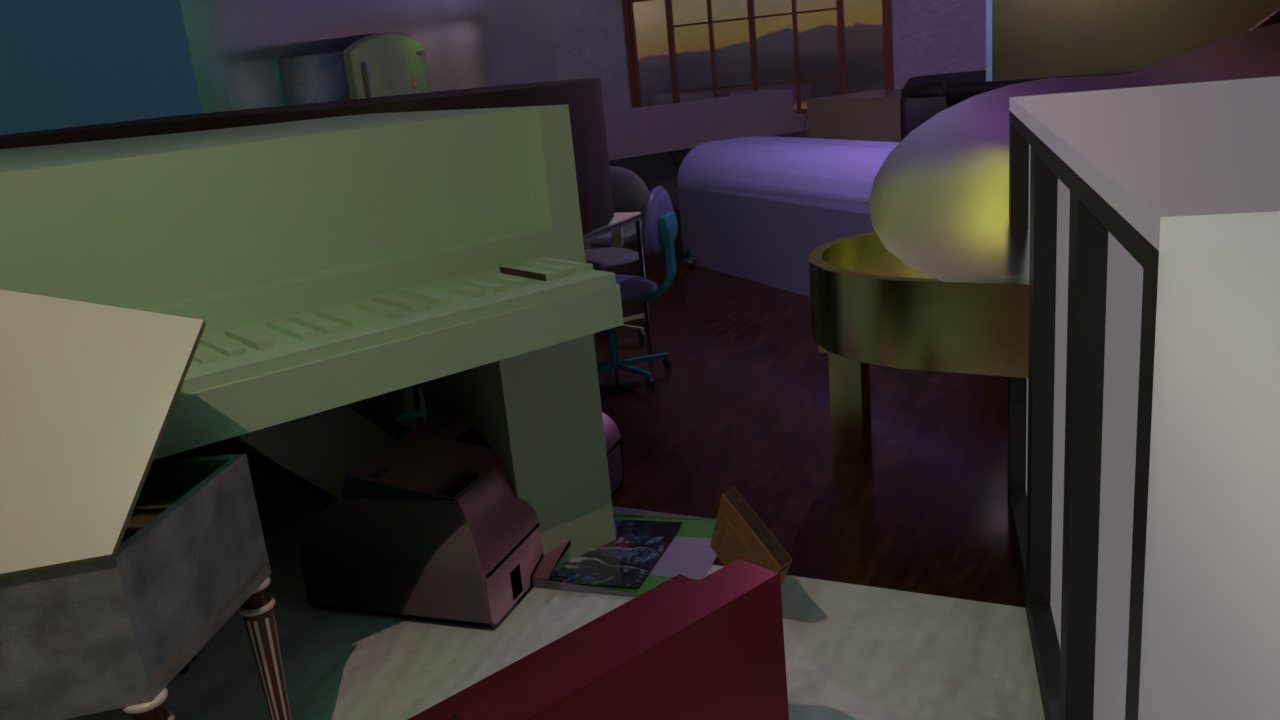} &
    \formattedgraphics{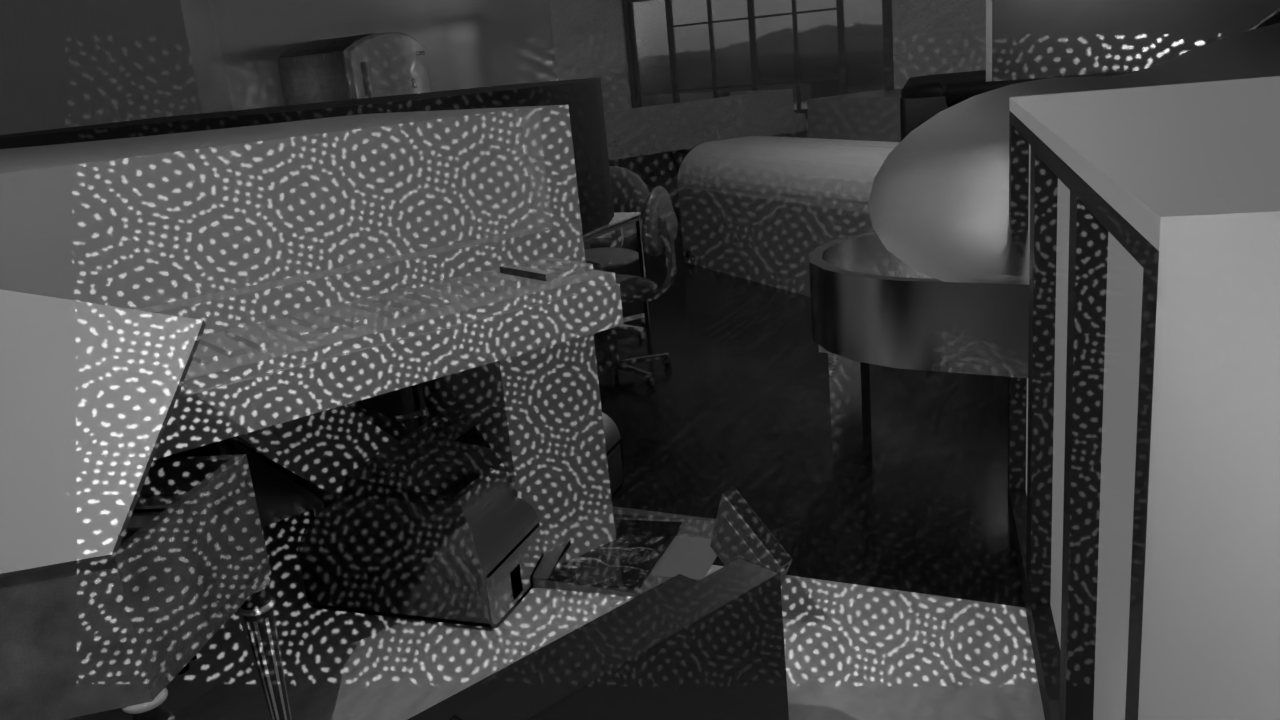} &
    \formattedgraphics{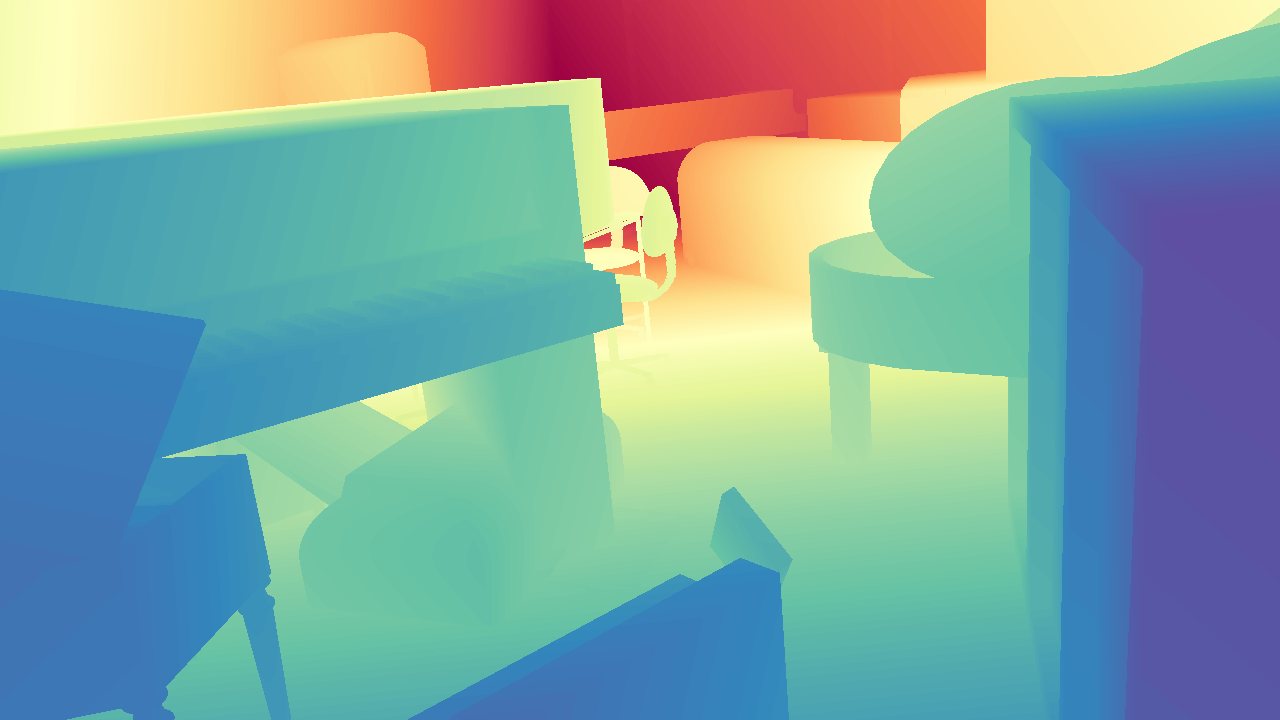} \\[1pt]
    \formattedgraphics{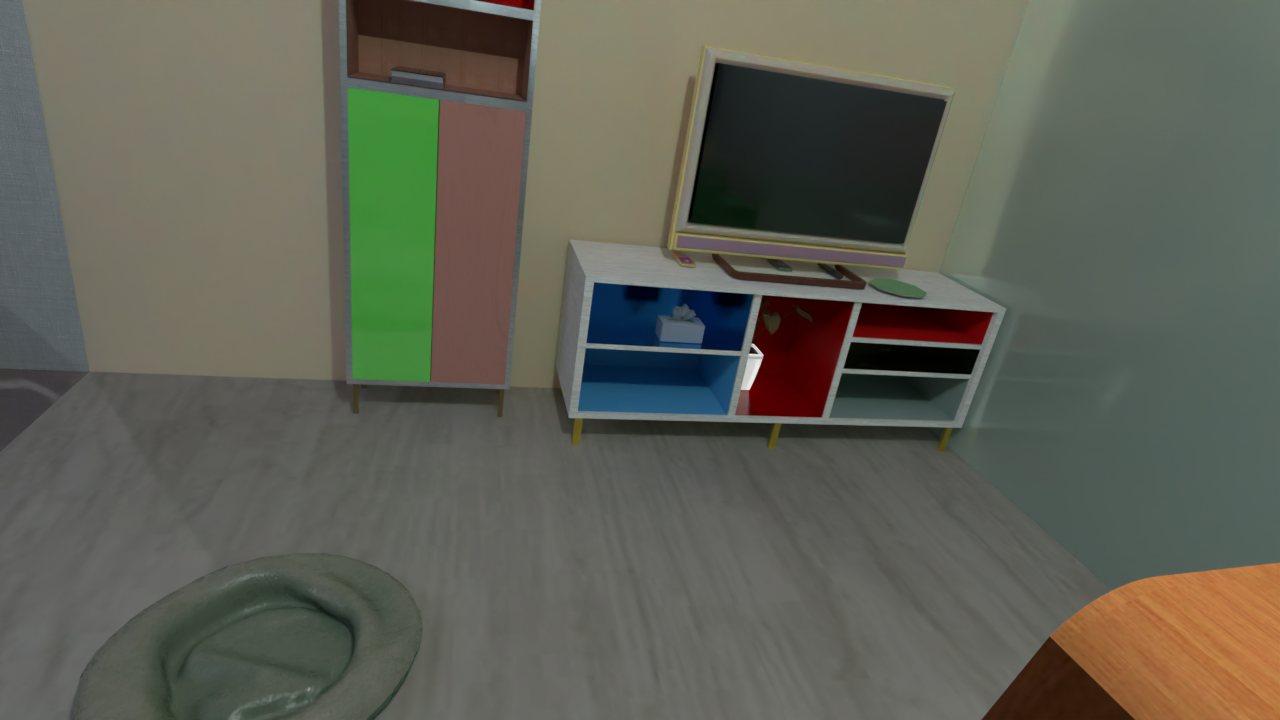} &
    \formattedgraphics{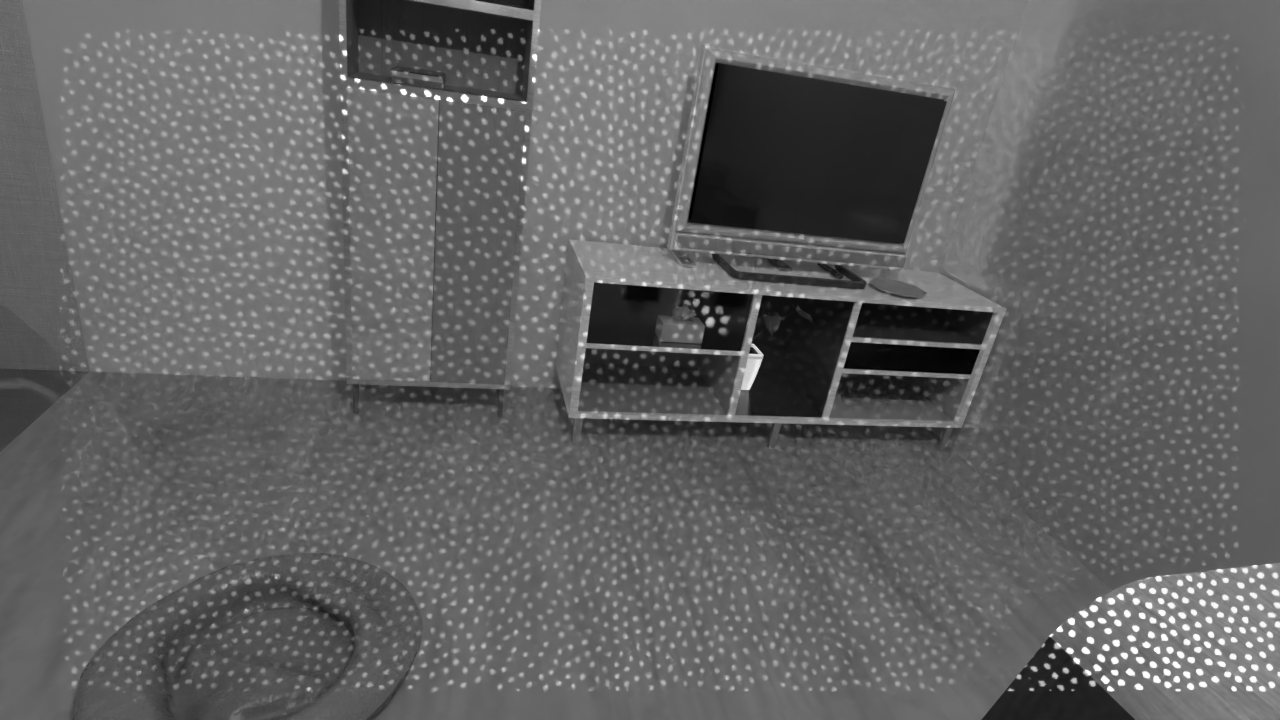} &
    \formattedgraphics{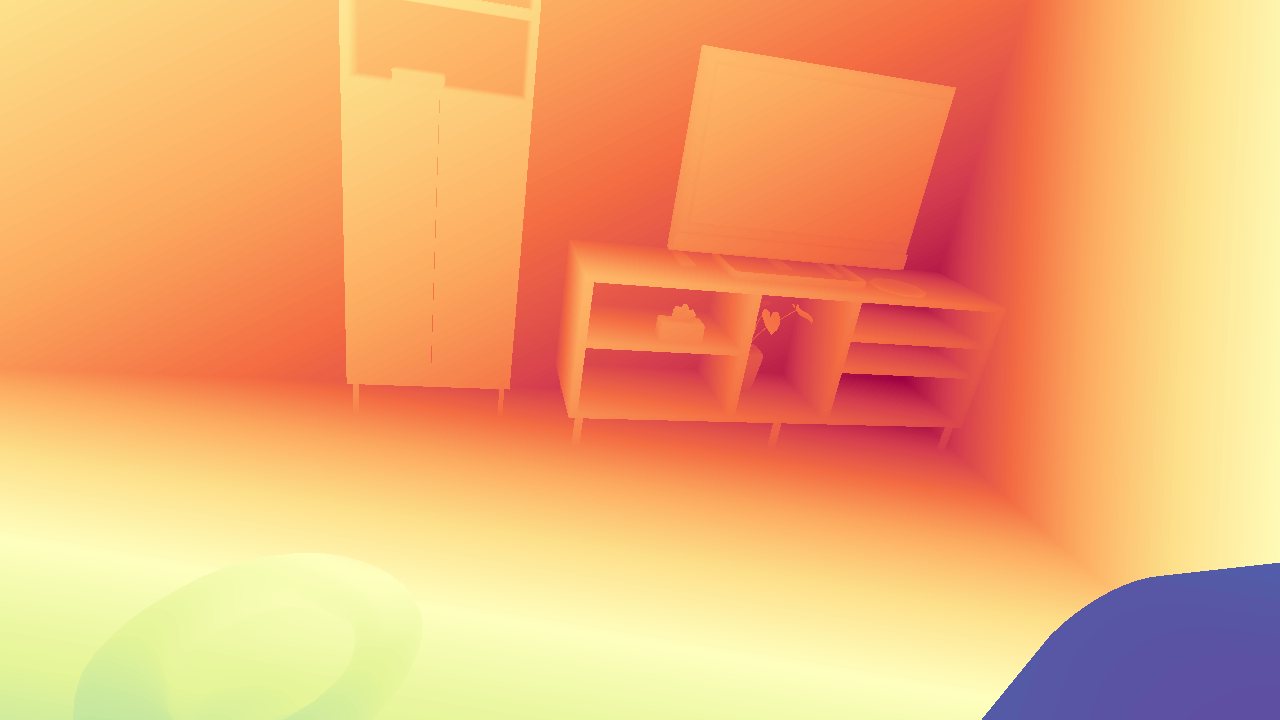} &
    \formattedgraphics{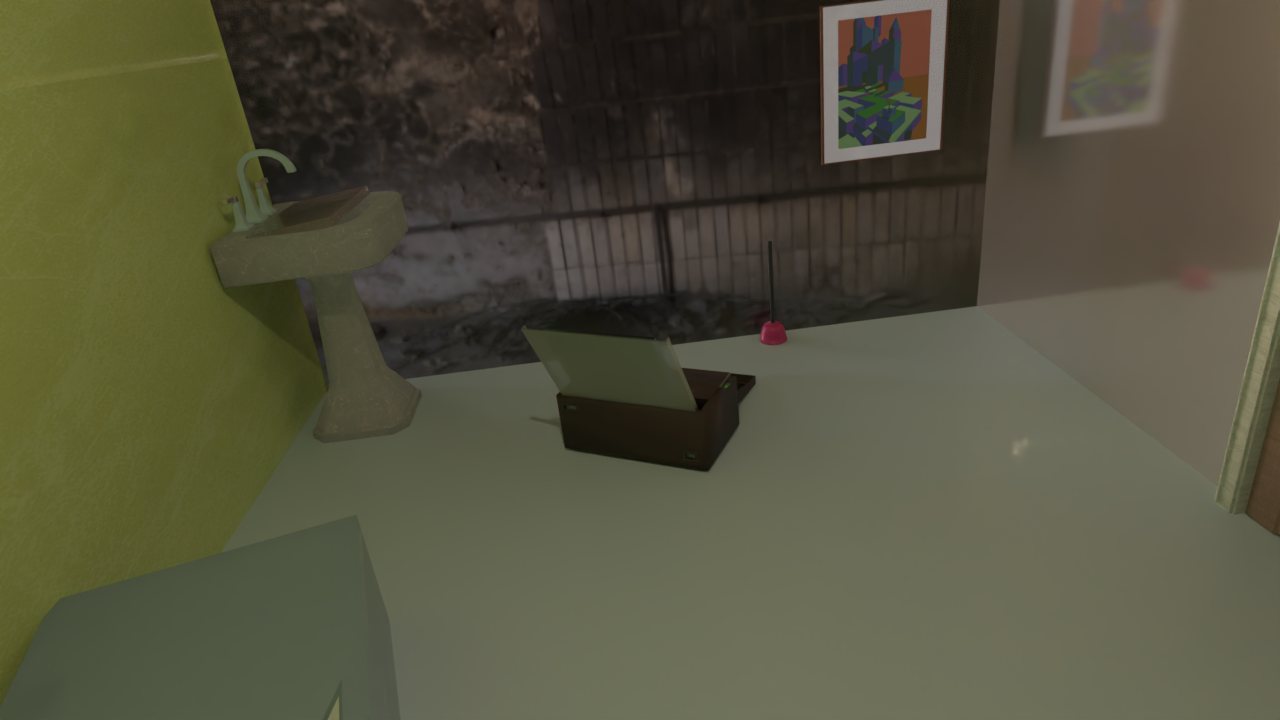} &
    \formattedgraphics{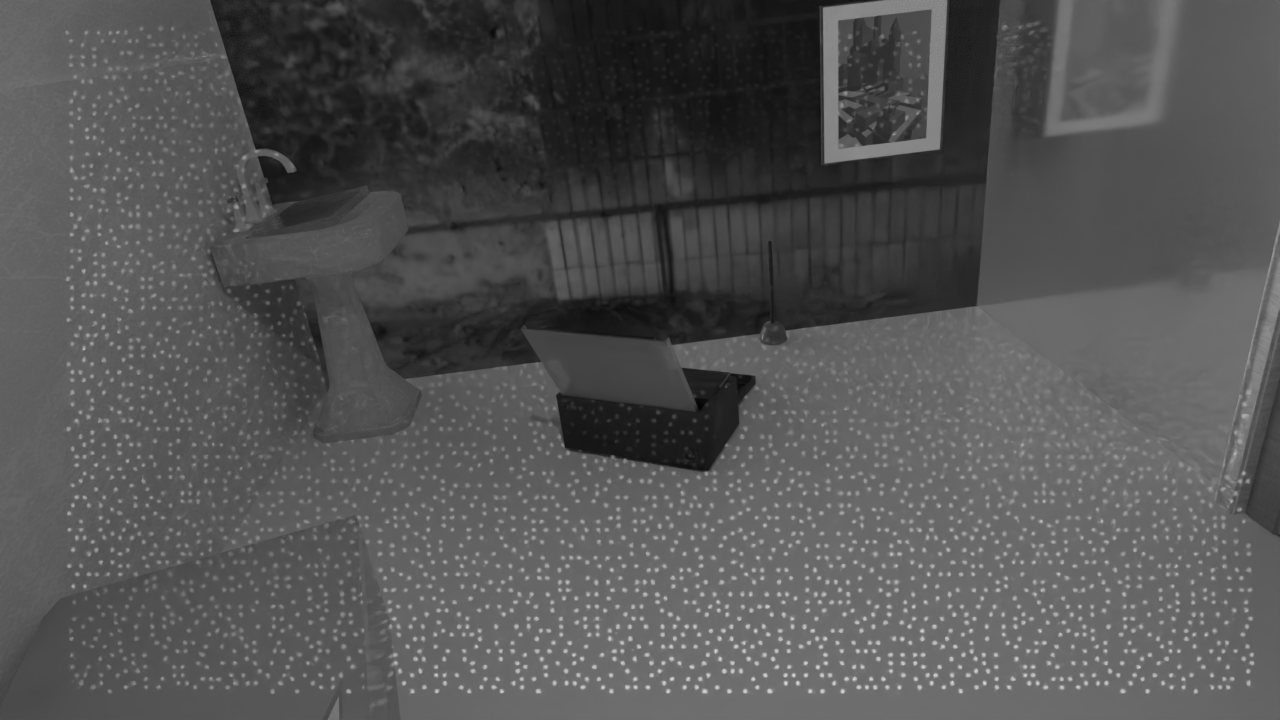} &
    \formattedgraphics{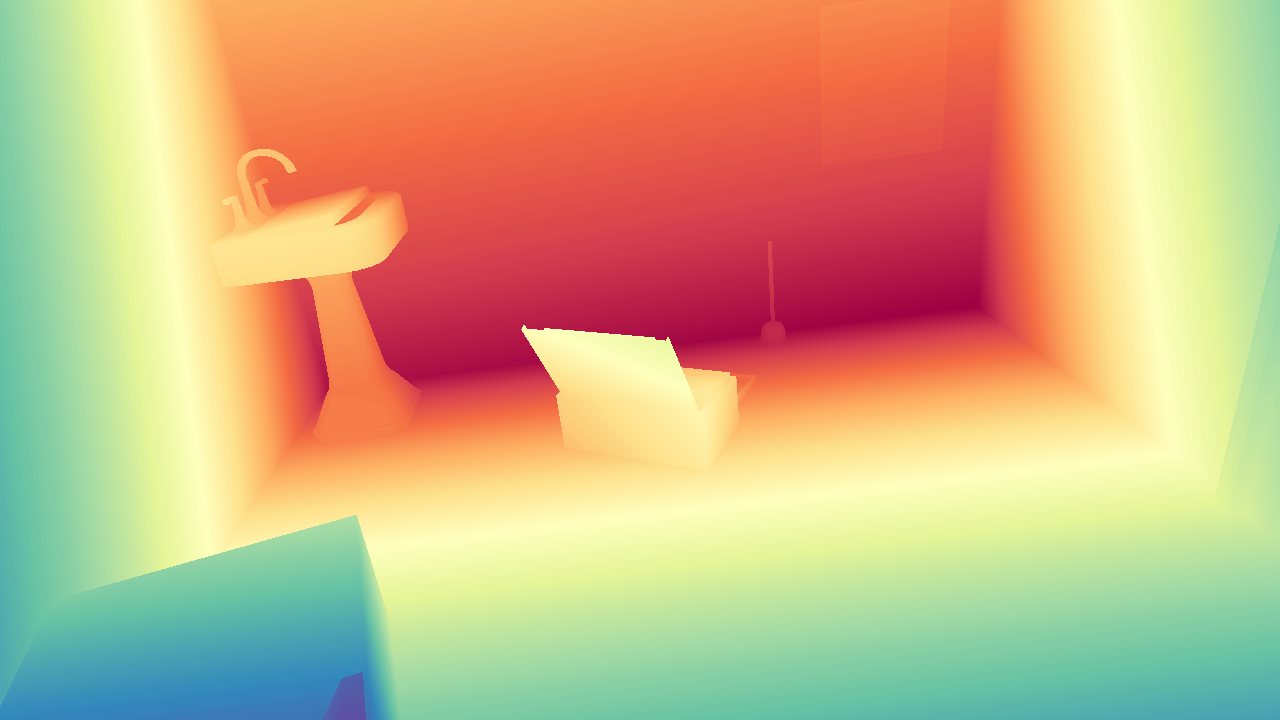} \\[1pt]
    \formattedgraphics{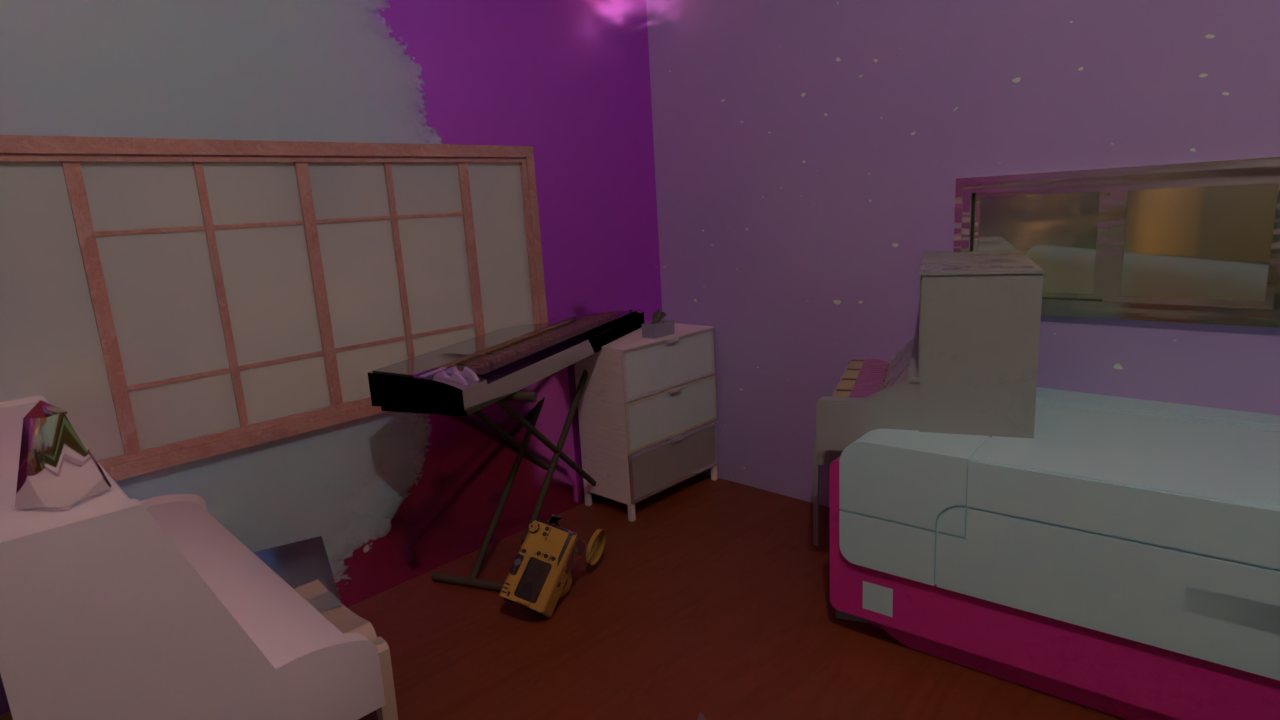} &
    \formattedgraphics{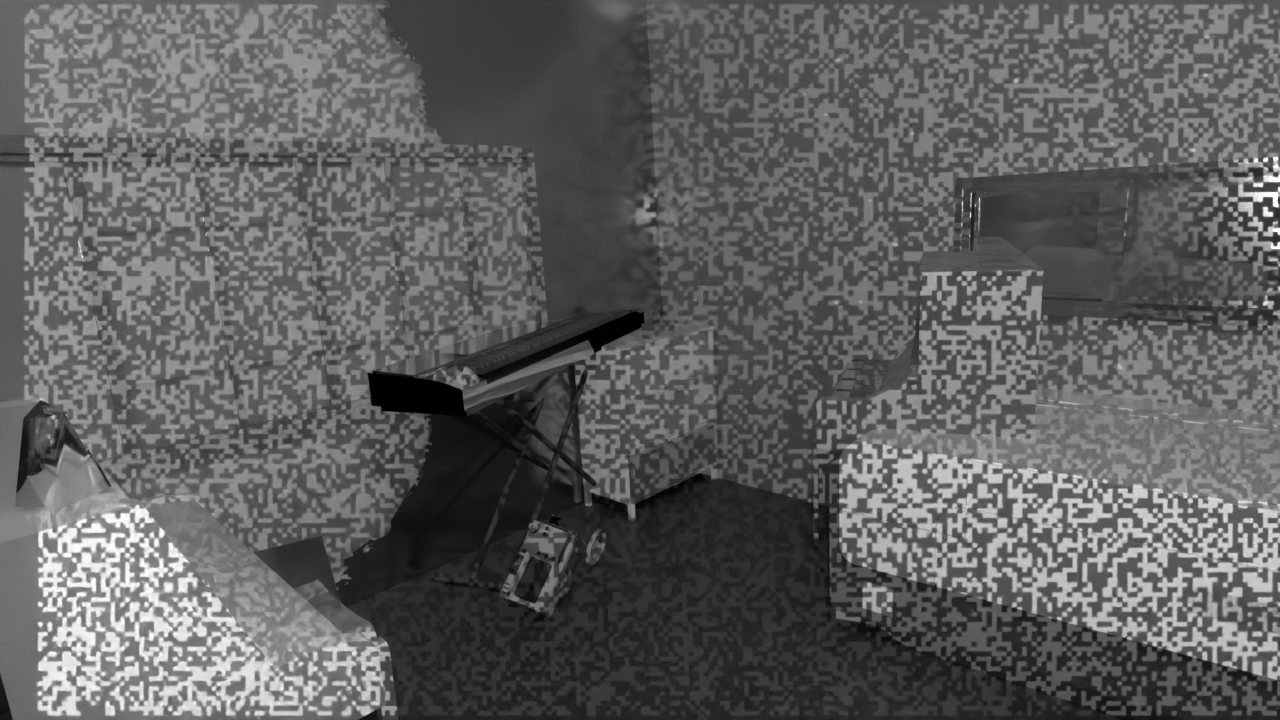} &
    \formattedgraphics{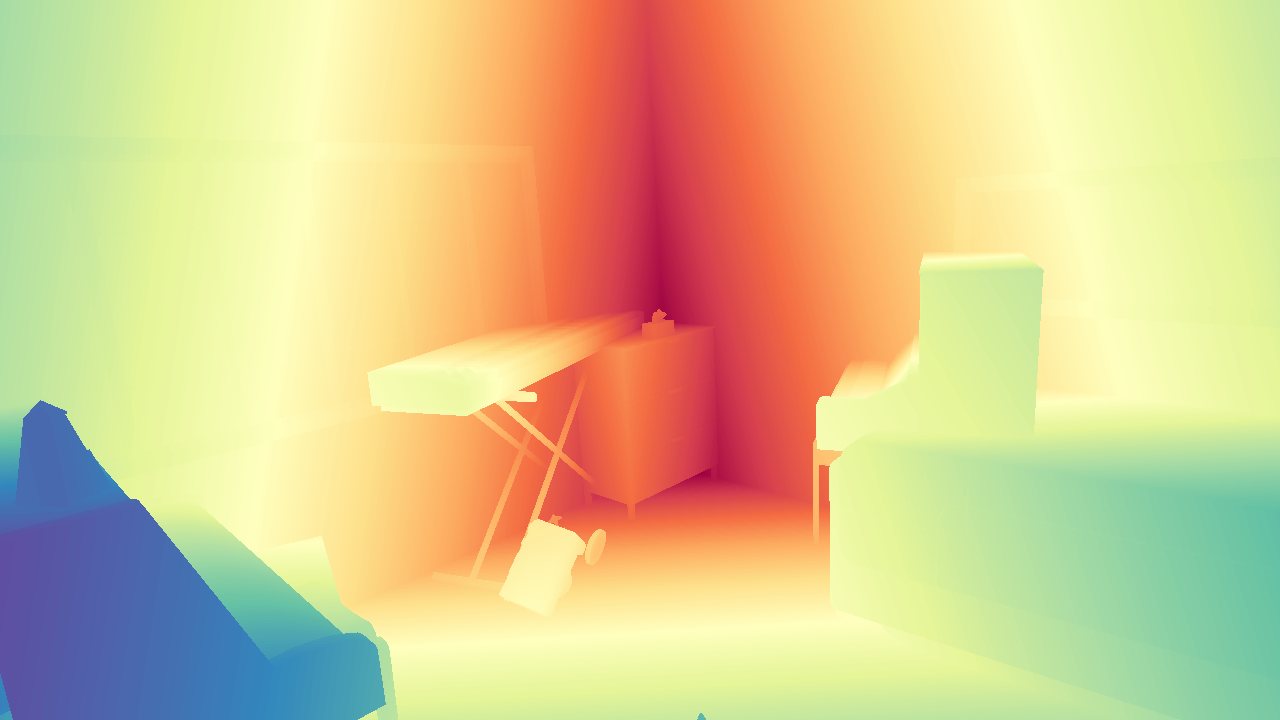} &
    \formattedgraphics{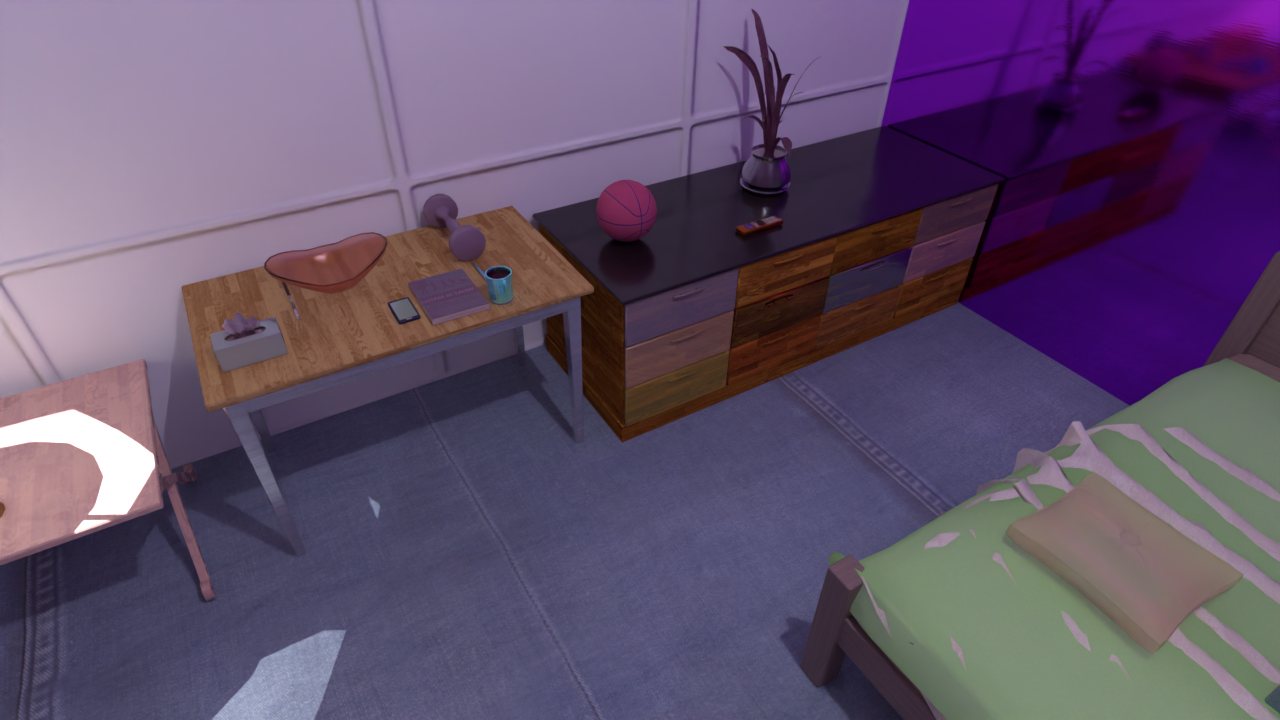} &
    \formattedgraphics{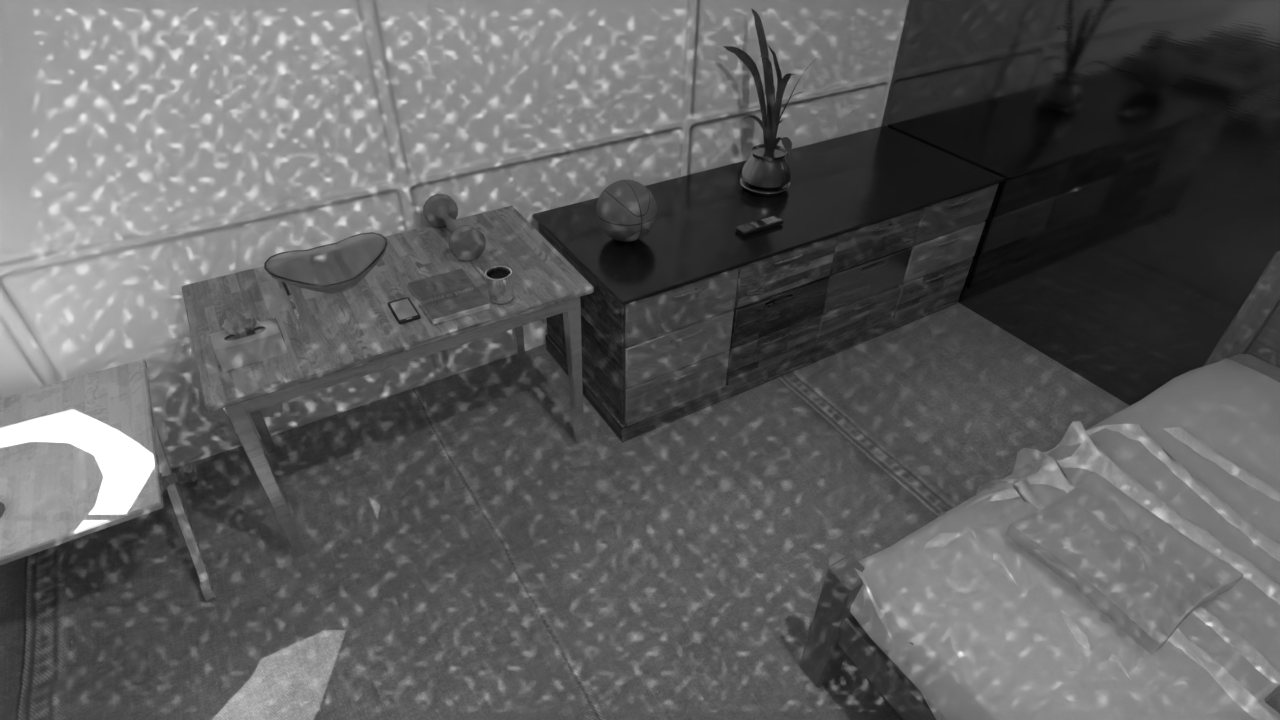} &
    \formattedgraphics{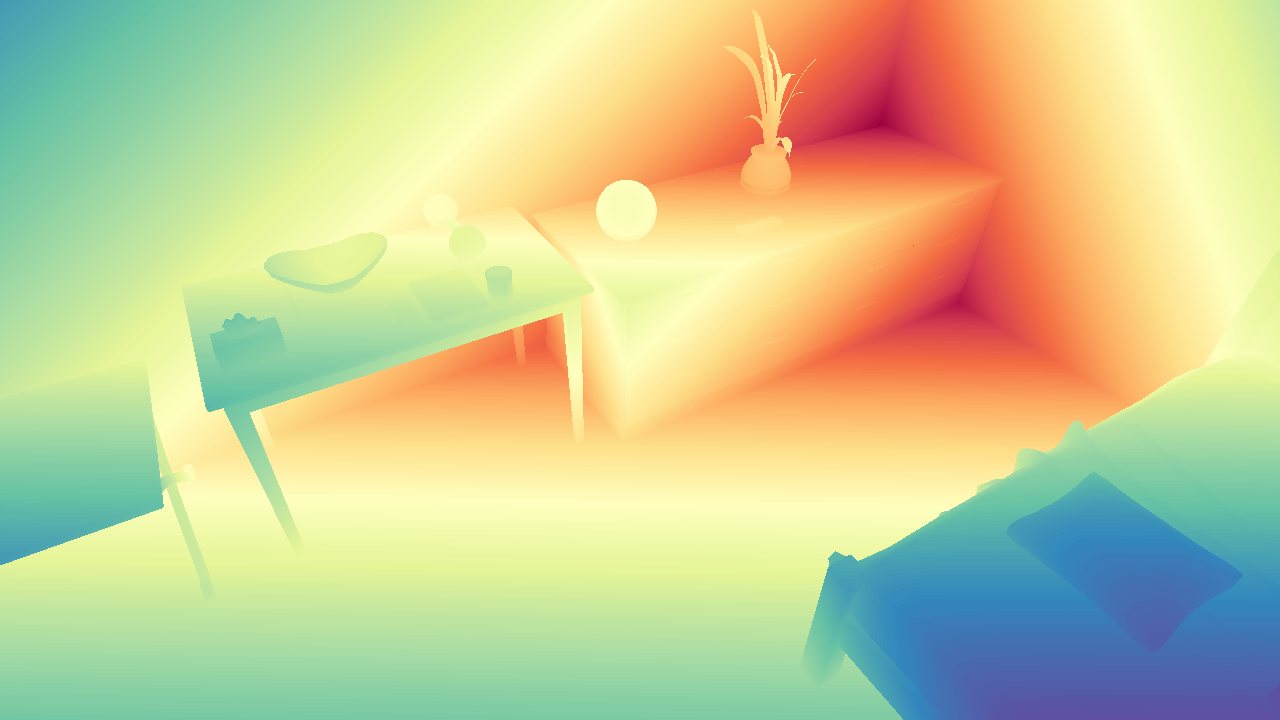} \\[1pt]

  \end{tabular}
  \caption{\textbf{Dataset Examples.} Each sample consists of the left and right RGB images, the pseudo-IR images, and a ground-truth depth map. The randomized strategy offers a diverse dataset that mirrors real-world structured light scanning scenarios, enabling robust training and evaluation of our method.}
  \label{fig:dataset-part1}
\end{figure*}

\renewcommand{\arraystretch}{1}

\bibliographystyle{ACM-Reference-Format}
\bibliography{references_appendix}